\documentclass[twocolumn]{svjour3}          

\smartqed           
\journalname{International Journal of Computer Vision}

\usepackage{graphicx}
\usepackage{latexsym}
\usepackage{times}
\usepackage{epsfig}
\usepackage{graphicx}
\usepackage{amsmath}
\usepackage{amssymb}
\usepackage{indentfirst}

\usepackage{bbold}
\usepackage{dsfont}

\usepackage{enumerate}
\usepackage{enumitem}
\usepackage{xspace}
\usepackage{xcolor}

\usepackage[toc,page]{appendix}

\usepackage{amsfonts}

\usepackage{caption}
\usepackage{subcaption}
\usepackage{bm}
\usepackage{isomath}

\usepackage[numbers]{natbib}

\usepackage[pagebackref=true,breaklinks=true,colorlinks,bookmarks=false,urlcolor={green!70!black},citecolor={green!70!black}]{hyperref}
\usepackage{footmisc}
\usepackage{stmaryrd}

\usepackage{fixltx2e}

\usepackage{dblfloatfix}
\usepackage{pbox}
\usepackage{capt-of}

\usepackage[normalem]{ulem}
\usepackage{multirow}

\usepackage{colortbl}

\usepackage{array}

\usepackage{colortbl}
\usepackage{tabularx}
\usepackage{arydshln}
\usepackage{xspace}

\usepackage{bbm}
\usepackage{algpseudocode}
\usepackage{algorithm}

\usepackage{booktabs}

\newcommand{\gray}[1]{\textcolor{gray}{#1}}

\usepackage{multibib}
\newcites{latex}{References}

\newcommand{\comment}[1]{}

\makeatletter
\@namedef{ver@everyshi.sty}{}
\makeatother

\definecolor{pink}{HTML}{db5a6b}
\definecolor{lblue}{HTML}{2e4e7e}
\definecolor{tiffany}{HTML}{1bd1a5}

\makeatletter
\DeclareRobustCommand\onedot{\futurelet\@let@token\bmv@onedotaux}
\def\bmv@onedotaux{\ifx\@let@token.\else.\null\fi\xspace}
\def\eg{\emph{e.g}\onedot} 
\def\ie{\emph{i.e}\onedot} 
 
\def\etc{\emph{etc}\onedot} \def\vs{\emph{vs}\onedot}
\def\wrt{w.r.t\onedot} 
\def\aka{a.k.a\onedot} 
\def\etal{\emph{et al}\onedot}
\makeatother

\definecolor{DarkGreen}{rgb}{0.0,0.5,0}

\usepackage{pifont}

\renewcommand\vec[1]{\ensuremath\boldsymbol{#1}}
\renewcommand\cdots{...}

\newcommand{\mZ}{\mathbf{Z}}

\newcommand{\vy}{\mathbf{y}}
\newcommand{\malpha}{\boldsymbol{A}}
\newcommand{\valpha}{\bm{\alpha}}

\newcommand{\tD}{\vec{\mathcal{D}}}
\newcommand{\tX}{\vec{\mathcal{X}}}

\newcommand{\mX}{\mathbf{X}}
\newcommand{\mA}{\mathbf{A}}

\newcommand{\mbrp}[1]{\mathbb{R}_{+}^{#1}}
\newcommand{\mbr}[1]{\mathbb{R}^{#1}}

\newcommand{\tAnb}{\mathcal{A}}

\newcommand{\idx}[1]{\mathcal{I}_{#1}}

\newcommand{\vpsi}{\boldsymbol{\psi}}

\newcommand{\mPsi}{\vec{\Psi}}

\DeclareMathOperator*{\myinf}{Inf}
\DeclareMathOperator*{\argmin}{arg\,min}

\DeclareMathOperator*{\softming}{SoftMin_{\text{$\gamma$}}}
\DeclareMathOperator*{\topminb}{TopMin_\text{$\beta$}}
\DeclareMathOperator*{\topmaxbb}{TopMax_{NZ\text{$\beta$}}}

\def\eg{\emph{e.g.}}
\def\wrt{\emph{w.r.t.}}

\def\ie{\emph{i.e.}}

\def\etc{\emph{etc.}} 
\def\vs{\emph{vs.}}

\def\etal{\emph{et al.}}

\newcommand{\mPhi}{\boldsymbol{\Phi}}

\newcommand{\mOmega}{\boldsymbol{\Omega}}

\newcommand{\mM}{\boldsymbol{M}}

\newcommand{\mD}{\boldsymbol{D}}
\newcommand{\vd}{\boldsymbol{d}}

\newcommand{\stkout}[1]{{\ifmmode\text{\sout{\ensuremath{#1}}}\else\sout{#1}\fi}}

\newcommand{\vm}{\mathbf{m}}

\begin{document}
\title{
Meet JEANIE: a Similarity Measure for 3D Skeleton Sequences via Temporal-Viewpoint Alignment 
}


\author{Lei Wang$^*$        \and Jun Liu    \and Liang Zheng \and Tom Gedeon \and
        Piotr Koniusz$^*$
}

\institute{$\cdot\;\!$L. Wang is a Research Fellow at the School of Computing, 
the Australian National University (ANU). 
           \email{lei.w@anu.edu.au}.\\         
           \and
           J. Liu is an Assistant Professor at the 
           Singapore University of Technology and Design.
           \\         
           \and
           L. Zheng is an Associate Professor in the School of Computing, ANU.\\
            \and
        T. Gedeon is an Artificial Intelligence Chair for Human-Centric Advancements at the Curtin University, Australia.
        \\ \and
           P. Koniusz ($^*$: equal contribution, PK: corresponding author) is a Principal Research Scientist at Data61{\color{red}\ding{170}}CSIRO and an Honorary Associate Professor at the ANU. 
           \email{piotr.koniusz@data61.csiro.au}. $\qquad$Code is available: \href{https://github.com/LeiWangR/JEANIE}{https://github.com/LeiWangR/JEANIE}.
}

\date{Received: 08.30.2023 / Accepted: 03.25.2024}

\maketitle

\begin{sloppypar}
\begin{abstract}
Video sequences exhibit significant nuisance  variations (undesired effects)  of  speed of actions,  temporal locations, and subjects' poses, leading to temporal-viewpoint misalignment when comparing two sets of frames or evaluating the similarity of two sequences. Thus, we propose Joint tEmporal and cAmera viewpoiNt alIgnmEnt (JEANIE) for sequence pairs. In particular, we focus on 3D skeleton sequences whose camera and subjects' poses  can be easily manipulated in 3D. We evaluate JEANIE on skeletal Few-shot Action Recognition (FSAR), where  matching well temporal blocks (temporal chunks that make up a sequence) of support-query sequence pairs (by factoring out nuisance variations) is essential due to limited samples of novel classes. Given a query sequence, we create its several views by simulating several camera locations. For a support sequence, we match it with view-simulated query sequences, as in the popular Dynamic Time Warping (DTW). 
Specifically, each support temporal block can be matched to the query temporal block with the same or adjacent (next) temporal index, and adjacent camera views to achieve joint local temporal-viewpoint warping. JEANIE selects the smallest distance among   matching paths with different temporal-viewpoint warping patterns, an advantage over DTW which only performs temporal alignment.
We also propose an unsupervised FSAR akin to clustering of sequences with JEANIE as a distance measure. JEANIE achieves state-of-the-art results on NTU-60, NTU-120, Kinetics-skeleton and UWA3D Multiview Activity II on supervised and unsupervised FSAR, and their meta-learning inspired fusion.
\end{abstract}
\end{sloppypar}

\section{Introduction}
\label{sec:intro}

\begin{sloppypar}
Action recognition 
is  a key topic in computer vision, with  applications in video surveillance \cite{lei_thesis_2017,lei_icip_2019,wang2024taylor}, human-computer interaction, sport analysis and robotics. Many pipelines \cite{Tran_2015_ICCV,Feichtenhofer_2016_CVPR,Feichtenhofer_2017_CVPR,Carreira_2017_CVPR,high-order-first,democratic,lei_tip_2019,hosvd,wang2023flow,jeanie,wang2023robust,Rahman_2023_CVPR,Zhang_2024_WACV,wang2024taylor,li2023stprivacy} perform (action) classification given a large amount of labeled training data. However, manually labeling videos for 3D skeleton sequences is laborious, and such pipelines need to be retrained or finetuned for new class concepts. Popular  action recognition networks such as the two-stream neural network \cite{Feichtenhofer_2016_CVPR,Feichtenhofer_2017_CVPR,Wang_2017_CVPR} and 3D   Convolutional Neural Network (3D CNN) \cite{Tran_2015_ICCV,Carreira_2017_CVPR}   aggregate  frame-wise and temporal block representations, respectively. However, such networks are  trained on large-scale datasets such as Kinetics \cite{Carreira_2017_CVPR,lei_iccv_2019,lei_mm_21,koniusz2021high} under a fixed set of training classes. 
\end{sloppypar}

Thus, there exists a growing interest in devising effective  Few-shot Learning (FSL) models for action recognition, termed Few-shot Action Recognition (FSAR), that rapidly adapt to  novel classes given few training samples~\cite{mishra2018generative,xu2018dense,guo2018neural,dwivedi2019protogan,hongguang2020eccv,kaidi2020cvpr,udtw_eccv22}. FSAR  models are scarce due to the volumetric nature of videos and large intra-class variations. 

\begin{figure*}[t]
\begin{minipage}{0.75\linewidth}
\centering
\centering\includegraphics[width=\linewidth]{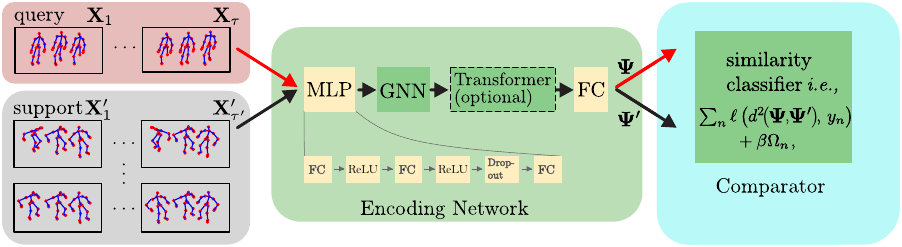}
\caption{Skeletal FSAR (simplified overview) takes episodes of query and support sequences, splits them into temporal blocks ($\mathbf{X}_1,\cdots,\mathbf{X}_\tau$ and $\mathbf{X}'_1,\cdots,\mathbf{X}'_\tau$), passes them to the Encoding Network to obtain features $\mathbf{\Psi}=[\boldsymbol{\psi}_1,\cdots,\boldsymbol{\psi}_\tau]$ and $\mathbf{\Psi}'=[\boldsymbol{\psi}'_1,\cdots,\boldsymbol{\psi}'_{\tau'}]$, and the Comparator which typically uses some distance measure $d(\cdot,\cdot)$, regularization $\Omega$ and the similarity classifier $\ell(\cdot,\cdot)$. }\label{fig:pipe_basic_A}
\end{minipage}
\hspace{0.2cm}
\begin{minipage}{0.23\linewidth}
\centering
\centering\includegraphics[width=0.45\textwidth]{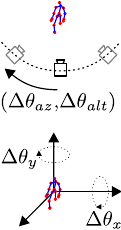}
\caption{One may use ({\em top}) stereo projections to simulate different camera views or simply use ({\em bottom}) Euler angles to rotate 3D scene.}\label{fig:pipe_angles}
\end{minipage}
%
\end{figure*}

In contrast, FSL for image recognition has been widely studied  \cite{miller_one_example,Li9596,NIPS2004_2576,BartU05,fei2006one,lake_oneshot} including contemporary CNN-based FSL methods  
\cite{meta25,f4Matching,f1,f5Model-Agnostic,f8Relation,sosn} which use meta-learning, prototype-based learning or feature representation learning. Just in 2020--2024, many FSL methods \cite{guo2020broader,nikita2020eccv,shuo2020eccv,moshe2020eccv,qinxuan2021wacv,nanyi2020accv,jiechao2020accv,kai2020cvpr,thomas2020cvpr,kaidi2020cvpr,luming2020cvpr,maxexp,mlso,hao_fsl,keypoint_fsl,hao_fsl2,Kang_2023_CVPR,Shi_2024_WACV,Zhang_2021_CVPR,shan_fsl,Zhang_2022_CVPR,Zhang_eccv_2022,keypoint_fsl2} have been dedicated to image classification or detection. 
In contrast, in this paper, we aim at advancing few-shot action recognition of articulated set of connected 3D body joints, simply put, skeletal FSAR.

With the exception of very recent models  \cite{liu_fsl_cvpr_2017,Liu_2019_NTURGBD120,2021dml,memmesheimer2021skeletondml,ma2022learning,udtw_eccv22,zhu2023adaptive}, FSAR approaches that learn from skeleton-based 3D body joints are  scarce. The above situation prevails despite action recognition from articulated sets of connected body joints, expressed as 3D coordinates, does offer a number of advantages over videos such as (i) the lack of the background clutter, (ii) the volume of data being several orders of magnitude smaller, and (iii) the 3D geometric manipulations of skeletal sequences being algorithm-friendly.

\begin{figure*}[t]
\includegraphics[width=\linewidth]{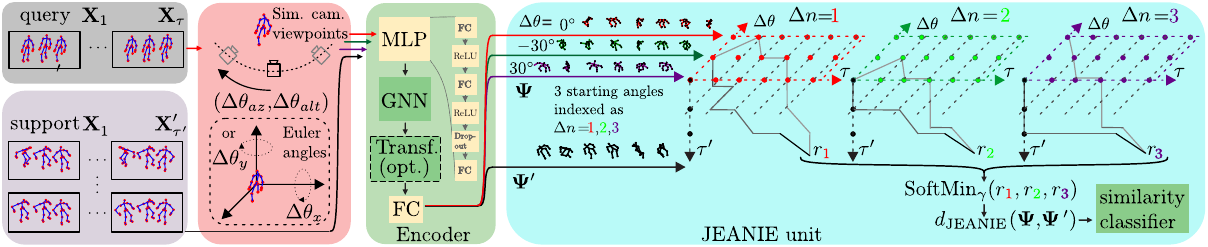}
\caption{Our 3D skeleton-based FSAR with JEANIE. Frames from a query sequence and a support sequence are split into short-term temporal blocks $\mX_1,\cdots,\mX_{\tau}$ and $\mX'_1,\cdots,\mX'_{\tau'}$ of length $M$ given stride $S$. Subsequently, we generate (i) multiple rotations by $(\Delta\theta_x,\Delta\theta_y)$ of each query skeleton by either Euler angles (baseline approach) or (ii) simulated camera views (gray cameras) by camera shifts  $(\Delta\theta_{az},\Delta\theta_{alt})$  \wrt ~the assumed average camera location (black camera). We pass all skeletons via Encoding Network (with an optional transformer) to obtain feature tensors $\mPsi$ and $\mPsi'$, which are directed to JEANIE. We note that the temporal-viewpoint alignment takes place in 4D space (we show a 3D case with three views: $-30^\circ, 0^\circ, 30^\circ$). Temporally-wise, JEANIE starts from the same $t\!=\!(1,1)$ and finishes at $t\!=\!(\tau,\tau')$ (as in DTW). Viewpoint-wise, JEANIE starts from every possible camera shift $\Delta\theta\in\{-30^\circ, 0^\circ, 30^\circ\}$ (we do not know the true correct pose) and finishes at one of possible camera shifts. At each step, the path may move by no more than $(\pm\!\Delta\theta_{az},\pm\!\Delta\theta_{alt})$ to prevent erroneous alignments. Finally, SoftMin picks up the smallest distance.}
\label{fig:pipe}
\end{figure*}

Video sequences may be captured under varying camera poses where subjects may follow different trajectories resulting in subjects' pose variations. Variations of action speed, location, and motion dynamics are also common. Yet, FSAR has to learn and infer similarity between support-query sequence pairs under the limited number of samples of novel classes. Thus, a good  measure of similarity between  support-query sequence pairs has to factor out the above variations. To this end, we propose a  FSAR  model that learns on   skeleton-based 3D body joints via Joint tEmporal and cAmera  viewpoiNt alIgnmEnt (JEANIE). We focus on 3D skeleton sequences as camera/subject's pose can be easily altered in 3D by the use of projective camera geometry. 

\begin{figure*}[t]
\begin{minipage}{0.3\linewidth}
\centering
\includegraphics[trim=0 0 0 0, clip=true,width=0.80\linewidth]{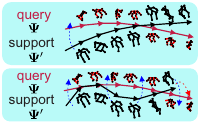}
\caption{\label{fig:sq-match}({\em top}) In viewpoint-invariant learning, the distance between query features $\mPsi$ and support features $\mPsi'$ has to be computed. The blue arrow indicates that trajectories of both actions need alignment. ({\em bottom}) In real life, subject's 3D body joints deviate from one ideal trajectory, and so advanced viewpoint alignment strategy is needed.}
\end{minipage}
$\;\;\;$
\begin{minipage}{0.36\linewidth}
\centering
\includegraphics[trim=0cm 0cm 0cm 0cm, clip=true,width=0.97\linewidth]{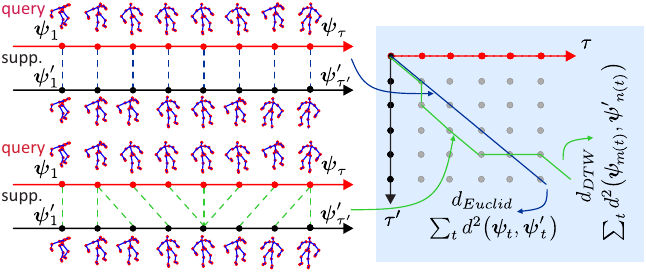}
\caption{\label{fig:dtw_vs_euc}Euclidean dist. \vs~DTW. ({\em top}) Feature vectors $\boldsymbol{\psi}_t$ and $\boldsymbol{\psi}'_{t}$ of query and support frames (or temp. blocks) are matched along time $t$: $d_{Euclid}(\mathbf{\Psi},\mathbf{\Psi}')\!=\!\sum_t d^2(\boldsymbol{\psi}_t, \boldsymbol{\psi}'_{t})$. ({\em bottom}) For DTW, a path with minimum aggregated distance is selected as $d_{DTW}(\mathbf{\Psi},\mathbf{\Psi}')\!=\!\sum_t d^2(\boldsymbol{\psi}_{m(t)}, \boldsymbol{\psi}'_{n(t)})$, and $m(t)$ and $n(t)$  parameterize query and support indexes. One is permitted steps $\downarrow$, $\searrow$, $\rightarrow$  in the graph. We expect $d_{DTW}\leq d_{Euclid}$.
}
\end{minipage}
$\;\;\;$
\begin{minipage}{0.3\linewidth}
\centering
\includegraphics[trim=0cm 0cm 0cm 0cm, clip=true,width=0.80\linewidth]{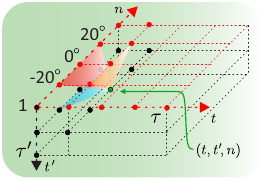}
\caption{\label{fig:jeanie_temp-shift}JEANIE (1-max shift). 
We loop over all points. At $(t,t',n)$ (green point) we add its base distance to the minimum of accumulated distances at $(t,t'\!\!-\!1,n\!-\!1)$, $(t,t'\!\!-\!1,n)$, $(t,t'\!\!-\!1,n\!+\!1)$ (orange plane), $(t\!-\!1,t'\!\!-\!1,n\!-\!1)$, $(t\!-\!1,t'\!\!-\!1,n)$, $(t\!-\!1,t'\!\!-\!1,n\!+\!1)$ (red plane) and  $(t\!-\!1,t'\!,n\!-\!1)$, $(t\!-\!1,t'\!,n)$, $(t\!-\!1,t'\!,n\!+\!1)$ (blue plane). 
}
\end{minipage}
\end{figure*}

JEANIE achieves good matching of queries with support sequences by simultaneously modeling the optimal (i) temporal and (ii) viewpoint alignments. To this end, we build on  soft-DTW \cite{marco2017icml}, a differentiable variant of Dynamic Time Warping (DTW) \cite{marco2011icml} (Fig. \ref{fig:dtw_vs_euc} is an overview how DTW differs from the Euclidean distance). Given a query sequence, we create its several views by simulating several camera locations. For a support sequence, we can match it with view-simulated query
sequences as in DTW. Specifically, with the goal of computing optimal distance,
each support temporal block\footnote{In fact, we bundle several neighboring frames into a temporal block, and perform alignment between support-query sequence pairs by temporally aligning temporal blocks rather than individual frames.} can be matched to the query temporal block with
the same temporal block index or neighbouring temporal block index to perform
a local time warping step. However, we simultaneously also let
each support temporal block match across adjacent camera views of
the query temporal block to achieve camera viewpoint warping. Multiple alignment patterns of query and support blocks result in multiple paths across temporal and viewpoint modes. Thus, each path represents a matching plan describing between which support-query block pairs the feature distances are evaluated and aggregated. By the use of soft-minimum, the path with the minimum aggregated distance is selected as the output of JEANIE.  Thus, while DTW provides optimal temporal alignment of support-query sequence pairs, JEANIE  simultaneously provides the optimal joint temporal-viewpoint alignment. 

To facilitate the viewpoint alignment in JEANIE, we use easy 3D geometric operations.
Specifically, we obtain skeletons under several viewpoints by  rotating  skeletons (zero-centered by hip) via Euler angles \cite{eulera}, 
or generating skeleton locations given simulated camera positions, according to the algebra of stereo projections \cite{sterproj}.

We note that  view-adaptive models 
for action recognition do exist. View Adaptive Recurrent Neural Network \cite{Zhang_2017_ICCV,8630687} is a classification model equipped with a view-adaptive subnetwork that contains the rotation/translation switches within its RNN backbone and the main LSTM-based network. Temporal Segment Network \cite{wang_2019_tpami} models long-range temporal structures with a new segment-based sampling and aggregation module. However, such   pipelines  require a large number of training samples with varying viewpoints and temporal shifts  to learn a robust model. Their limitations become evident when a network trained under a  fixed set of action classes has to be adapted to samples of novel classes. 
Our JEANIE does not suffer from such a limitation. 

Figure \ref{fig:pipe_basic_A} is a simplified overview of our pipeline which can serve as a template for baseline FSAR. It shows that our pipeline consists of an MLP  which takes neighboring frames forming a temporal block. Each sequence consists of several such temporal blocks.  However, as in Figure \ref{fig:pipe_angles}, we sample desired Euler rotations or simulated camera viewpoints, generate multiple skeleton views, and pass them to the MLP to get block-wise feature maps fed into a Graph Neural Network (GNN) \cite{kipf2017semi,uai_ke,felix2019icml,johannes2019iclr,wang2019linkage,hao2021iclr,zhang2023spectral,zhang2023mitigating}. 
We mainly use a linear  S$^2$GC~\cite{hao2021iclr,coles_hao,zhu2022generalized,wang2023adaptive}, with an optional transformer~\cite{dosovitskiy2020image}, and an FC layer to obtain  block feature vectors passed to  JEANIE whose output distance measurements flow into our  similarity classifier.
Figure \ref{fig:pipe} is a detailed overview of our supervised FSAR pipeline. 

Note that JEANIE can be thought of as a kernel in Reproducing Kernel Hilbert Spaces (RKHS) \cite{Smola03kernelsand} based on Optimal Transport \cite{ot} with a specific  temporal-viewpoint transportation plan. As kernels capture the similarity of sample pairs instead of modeling class labels, they are a natural choice for FSL and FSAR problems. 

%
In this paper, we extend our supervised FSAR model \cite{Wang_2022_ACCV} by introducing an unsupervised FSAR model, and a fusion of both supervised and unsupervised models. Our rationale for  an unsupervised FSAR extension is to demonstrate that the invariance properties of JEANIE (dealing with temporal and viewpoint variations) help naturally match sequences of the same class without the use of additional knowledge (class labels). Such a setting demonstrates that JEANIE is able to limit intra-class variations (temporal and viewpoint variations) facilitating unsupervised matching of sequences.

For unsupervised FSAR, JEANIE is used as a distance measure in the feature reconstruction term of dictionary learning and feature coding steps. Features of the temporal blocks are projected into such a dictionary space and the projection codes representing sequences are used for similarity measure between support-query sequences. This idea is similar to clustering training sequences into k-means clusters \cite{Csurka04visualcategorization} to form a dictionary. Then the assignments of test query sequences to such a dictionary can reveal their class labels based on labeled test support sequence falling into the same cluster. However, even with JEANIE used as a distance measure, one-hot assignments resulting from k-means are suboptimal. Thus, we investigate more recent soft assignment \cite{bilmes1998gentle, 10.1007/978-3-540-88690-7_52,6116129,6126534} and sparse coding approaches \cite{10.5555/2976456.2976557, 5206757}. 

Finally, we also introduce  a simple fusion of supervised and unsupervised FSAR 
by  alignment  of supervised and unsupervised FSAR features or by MAML-inspired \cite{f5Model-Agnostic} fusion of unsupervised and supervised FSAR losses in the so-called inner and outer loop, respectively.

\vspace{0.2cm}

\begin{sloppypar}

Below are our contributions:
\renewcommand{\labelenumi}{\roman{enumi}.}
\begin{enumerate}[leftmargin=0.6cm]
\item We propose JEANIE that performs the joint alignment of temporal blocks and simulated camera viewpoints of 3D skeletons between support-query sequences to select the optimal alignment path which realizes joint temporal (time) and viewpoint warping. We evaluate JEANIE on skeletal few-shot action recognition, where matching correctly support and query sequence pairs (by factoring out nuisance variations) is essential due to limited samples representing novel classes.
\item To simulate different camera locations for 
3D  skeleton  sequences, we consider rotating them  (1) by Euler angles within a specified range along  axes, or (2) towards the  simulated camera locations based on the  algebra of stereo projection. %
\item  We propose unsupervised FSAR where JEANIE is used as a distance measure in the feature reconstruction term of dictionary learning and coding steps (we investigate several such coders). We use projection
codes to represent sequences. Moreover, we also introduce an effective fusion of both supervised and unsupervised FSAR models by unsupervised and supervised feature alignment term or MAML-inspired fusion of unsupervised and supervised FSAR losses. %
\item As minor contributions, we investigate different GNN backbones (combined with an optional transformer), as well as the optimal temporal size and stride for temporal blocks encoded by a simple 3-layer MLP unit before forwarding them to GNN. We also propose a simple  similarity-based loss encouraging the alignment of within-class sequences and preventing the alignment of between-class sequences.
\end{enumerate}

\end{sloppypar}

We achieve state-of-the-art results  on  few-shot action recognition on 
 large-scale NTU-60 \cite{Shahroudy_2016_NTURGBD},   NTU-120 \cite{Liu_2019_NTURGBD120}, Kine- tics-skeleton~\cite{stgcn2018aaai}, and  UWA3D Multiview Activity II \cite{Rahmani2016}.

\section{Related Works}
\label{sec:rel}
Below, we describe 3D skeleton-based  AR,   FSAR approaches, 
and Graph Neural Networks.

\vspace{0.1cm}
\noindent\textbf{Action recognition (3D skeletons).} 
  3D skeleton-based  action recognition pipelines often  use GCNs \cite{kipf2017semi}, \eg,  spatio-temporal GCN (ST-GCN) \cite{stgcn2018aaai}, 
    Attention enhanced Graph Convolutional LSTM network (AGC-LSTM)~\cite{Si_2019_CVPR},  Actional-Structural GCN (AS-GCN)~\cite{Li_2019_CVPR}, Dynamic Directed GCN (DDGCN)~\cite{10.1007/978-3-030-58565-5_45}, Decoupling GCN with DropGraph module~\cite{10.1007/978-3-030-58586-0_32}, Shift-GCN~\cite{Cheng_2020_CVPR}, Semantics-Guided Neural Networks (SGN)~\cite{Zhang_2020_CVPR2}, AdaSGN~\cite{Shi_2021_ICCV}, Context Aware GCN (CA-GCN)~\cite{Zhang_2020_CVPR1}, Channel-wise Topology Refinement Graph Convolution Network (CTR-GCN)~\cite{chen2021channel}, Efficient GCN \cite{9729609} and Disentangling and Unifying Graph Convolutions \cite{Liu_2020_CVPR}.
  As ST-GCN applies convolution along links between body joints, structurally distant joints, which may cover key patterns of actions, are largely ignored. While GCN can be applied to a fully-connected graph to capture complex interactions of body joints,  groups of nodes across space/time can be captured with tensors \cite{tensor_ar_old,hosvd}, semi-dynamic hypergraph neural networks \cite{ijcai2020-109}, hypergraph GNN \cite{9329123}, angular features \cite{qin_tnnls_22}, Higher-order Transformer (HoT) \cite{kim2021transformers} and Multi-order Multi-mode Transformer (3Mformer) \cite{lei_3former}. PoF2I \cite{8691567} transforms pose features into pixels. Recently,  Koopman pooling~\cite{Wang_2023_CVPR2}, an auxiliary feature refinement head~\cite{Zhou_2023_CVPR}, a Spatial-Temporal Mesh Transformer (STMT)~\cite{Zhu_2023_CVPR}, Strengthening Skeletal Recognizers \cite{qin2022strengthening}, and a Skeleton Cloud Colorization \cite{yang2023self} have been proposed for 3D skeleton-based AR.  

  However, such models  
rely on large-scale datasets to train large numbers of parameters, and cannot be  adapted with ease to novel class concepts whereas FSAR can.

\vspace{0.1cm}
\noindent\textbf{FSAR (videos).} 
Approaches \cite{mishra2018generative,guo2018neural,xu2018dense} use a generative model, graph matching on 3D coordinates and dilated networks, 
respectively. Approach \cite{Zhu_2018_ECCV} uses a  compound memory network. 
ProtoGAN \cite{dwivedi2019protogan}  generates action prototypes. Recent FSAR model \cite{hongguang2020eccv} uses permutation-invariant attention and second-order aggregation of temporal video blocks, whereas approach   \cite{kaidi2020cvpr} proposes a modified temporal alignment for query-support pairs via DTW. Recent video FSAR models include a mixed-supervised hierarchical contrastive learning (HCL)~\cite{zheng2022few}, Compound Prototype Matching~\cite{huang2022compound}, Spatio-temporal Relation Modeling~\cite{Thatipelli_2022_CVPR},  motion-augmented long-short contrastive learning (MoLo)~\cite{Wang_2023_CVPR} and 
Active Multimodal Few-shot Action Recognition (AMFAR) framework~\cite{Wanyan_2023_CVPR}. 

\vspace{0.1cm}
\noindent\textbf{FSAR (3D skeletons).} 
%
%
Few FSAR models use 3D skeletons 
\cite{liu_fsl_cvpr_2017,Liu_2019_NTURGBD120,2021dml,memmesheimer2021skeletondml,yang2023one}. Global Context-Aware Attention LSTM \cite{liu_fsl_cvpr_2017} focuses on informative joints. 
Action-Part Semantic Relevance-aware (APSR) model \cite{Liu_2019_NTURGBD120}  uses    semantic relevance among each body part and  action class at the  distributed  word  embedding  level. 
Signal Level Deep Metric Learning (DML) \cite{2021dml}  and Skeleton-DML \cite{memmesheimer2021skeletondml} 
encode  signals  as  images,  extract  CNN features and use multi-similarity miner loss. New skeletal FSAR includes Disentangled and Adaptive Spatial-Temporal Matching (DASTM) \cite{ma2022learning}, Adaptive Local-Component-Aware Graph Convolutional Network (ALCA-GCN)~\cite{zhu2023adaptive} and  uncertainty-DTW \cite{udtw_eccv22}. 

In contrast, we use temporal blocks of skeleton sequences encoded by GNNs under multiple simulated camera viewpoints to jointly apply temporal and viewpoint  alignment of query-support sequences to factor out nuisance variability.

\begin{sloppypar}
\vspace{0.1cm}
\noindent\textbf{Graph Neural Networks.} 
 GNNs modified to act on the specific structure of 3D skeletal data are very popular in action recognition, as detailed in ``Action recognition (3D skeletons)'' at the beginning of Section \ref{sec:rel}. In this paper, we leverage standard GNNs due to their good ability to represent graph-structured data. GCN \cite{kipf2017semi} applies graph convolution in the spectral domain, and enjoys the depth-efficiency when stacking multiple layers due to non-linearities. However, depth-efficiency extends the runtime due to backpropagation through consecutive layers. In contrast, a very recent family of so-called spectral filters do not require depth-efficiency but apply filters based on heat diffusion on graph adjacency matrix. As a result, these are fast linear models as learnable weights act on filtered node representations. Unlike general GNNs, SGC \cite{felix2019icml}, APPNP \cite{johannes2019iclr} and S$^2$GC~\cite{hao2021iclr} are  three such linear models which we investigate for the backbone, followed by an optional transformer, and an FC layer. 
\end{sloppypar}

\vspace{0.1cm}
\noindent\textbf{Transformers in action recognition.} Transformers have become popular in action recognition \cite{PLIZZARI2021103219,9728206,10.1145/3474085.3475473,8954447,DBLP:journals/corr/abs-2012-06399}. Vision Transformer (ViT)~\cite{dosovitskiy2020image} is the first  transformer model  for image classification but transformers find application even in recent pre-training models \cite{ropim-pretraining}. The success of transformers relies on their ability to establish exhaustive attention among visual tokens.  
Recent transformer-based AR models include Uncertainty-Guided Probabilistic Transformer (UGPT) \cite{Guo_2022_CVPR}, Recurrent Vision Transformer (RViT)~\cite{Yang_2022_CVPR}, Spatio-TemporAl cRoss (STAR)-transformer~\cite{Ahn_2023_WACV}, DirecFormer \cite{Truong_2022_CVPR}, Spatial-Temporal Mesh Transformer (STMT)~\cite{Zhu_2023_CVPR}, Semi-Supervised Video Transformer (SVFormer)~\cite{Xing_2023_CVPR} and Multi-order
Multi-mode Transformer (3Mformer)~\cite{Wang_2023_CVPR}.

In this work, we 
apply a simple optional transformer block with few layers following GNN to capture better block-level dependencies of 3D human body joints. 

\vspace{0.1cm}
\noindent\textbf{Multi-view  action recognition.} 
Multi-modal sensors enable multi-view action recognition \cite{lei_tip_2019,Zhang_2017_ICCV}. 
A Generative Multi-View Action Recognition framework  \cite{Wang_2019_ICCV}  integrates 
RGB and depth data by View Correlation Discovery Network while Synthetic Humans \cite{varol2021synthetic} generates synthetic training data to improve generalization to unseen viewpoints. Some works use multiple views of the  subject  \cite{Shahroudy_2016_NTURGBD,Liu_2019_NTURGBD120,8630687,Wang_2019_ICCV} to overcome the viewpoint variations for action recognition. 
Recently, a supervised contrastive learning framework \cite{Shah_2023_WACV} for multi-view was introduced.

In contrast, our  JEANIE performs jointly the  temporal and simulated viewpoint alignment in an end-to-end FSAR setting. This is a novel paradigm  based on improving the notion of similarity between sequences of support-query pair rather than learning class concepts. 

\section{Approach}

\label{sec:appr}

\begin{figure*}[t]
\centering
%

\begin{subfigure}[t]{0.32\linewidth}
\centering\includegraphics[trim=1.8cm 2.25cm 0.5cm 3.5cm, clip=true, width=\linewidth]{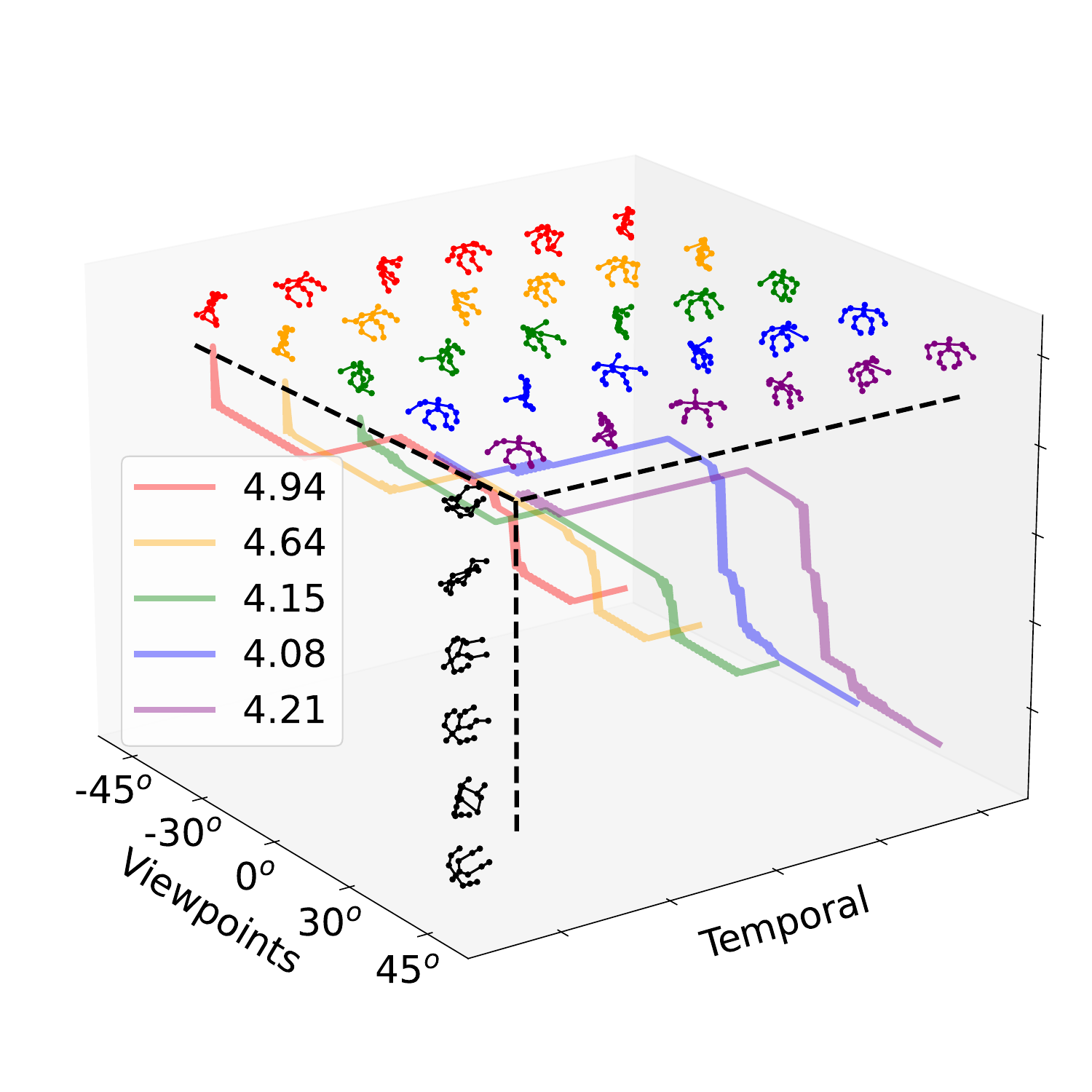}
\caption{soft-DTW (applied per view)}\label{fig:sdtw}
\end{subfigure}
\begin{subfigure}[t]{0.32\linewidth}
\centering\includegraphics[trim=1.8cm 2.25cm 0.5cm 3.5cm, clip=true, width=\linewidth]{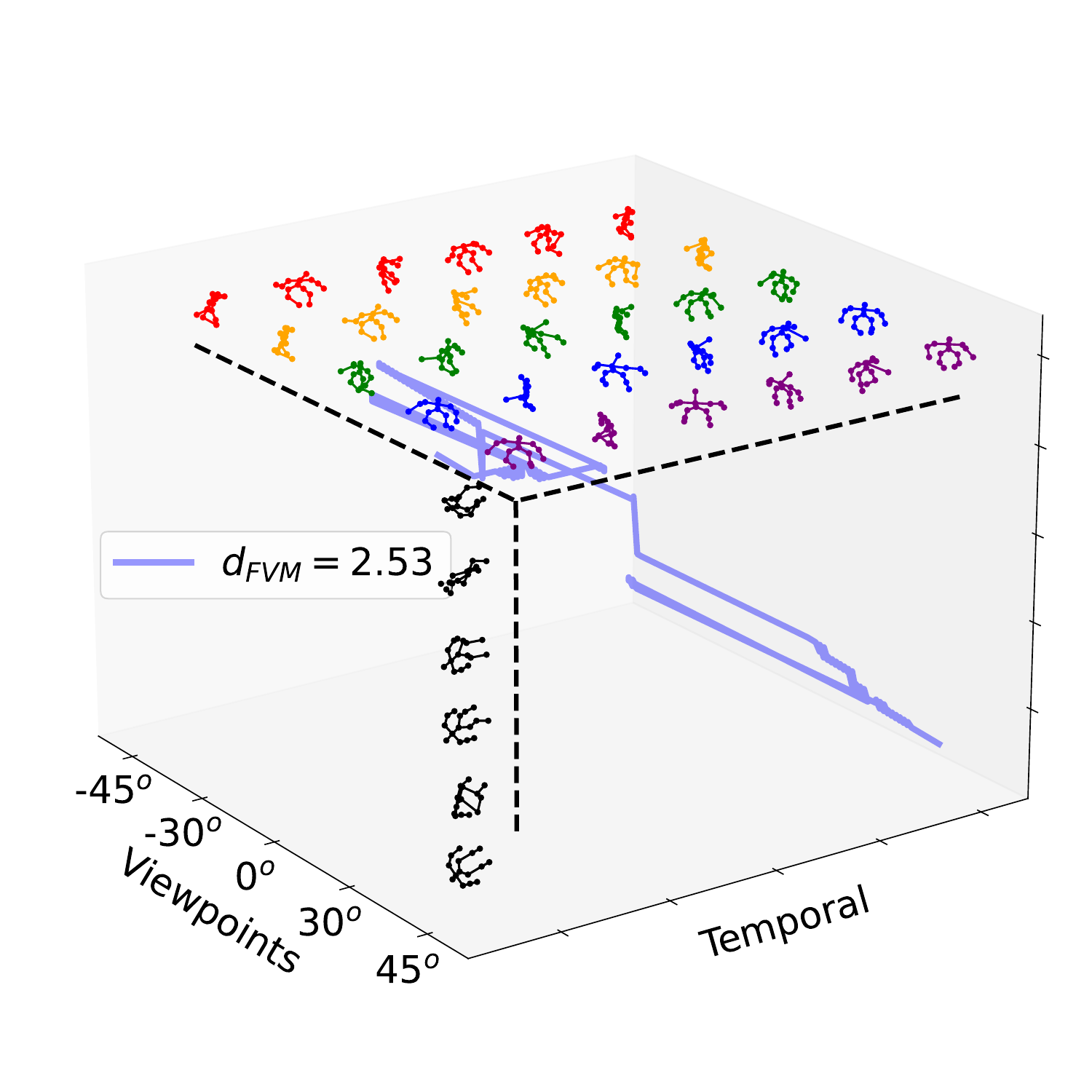}
\caption{FVM}\label{fig:fvm}
\end{subfigure}
\begin{subfigure}[t]{0.32\linewidth}
\centering\includegraphics[trim=1.8cm 2.25cm 0.5cm 3.5cm, clip=true, width=\linewidth]{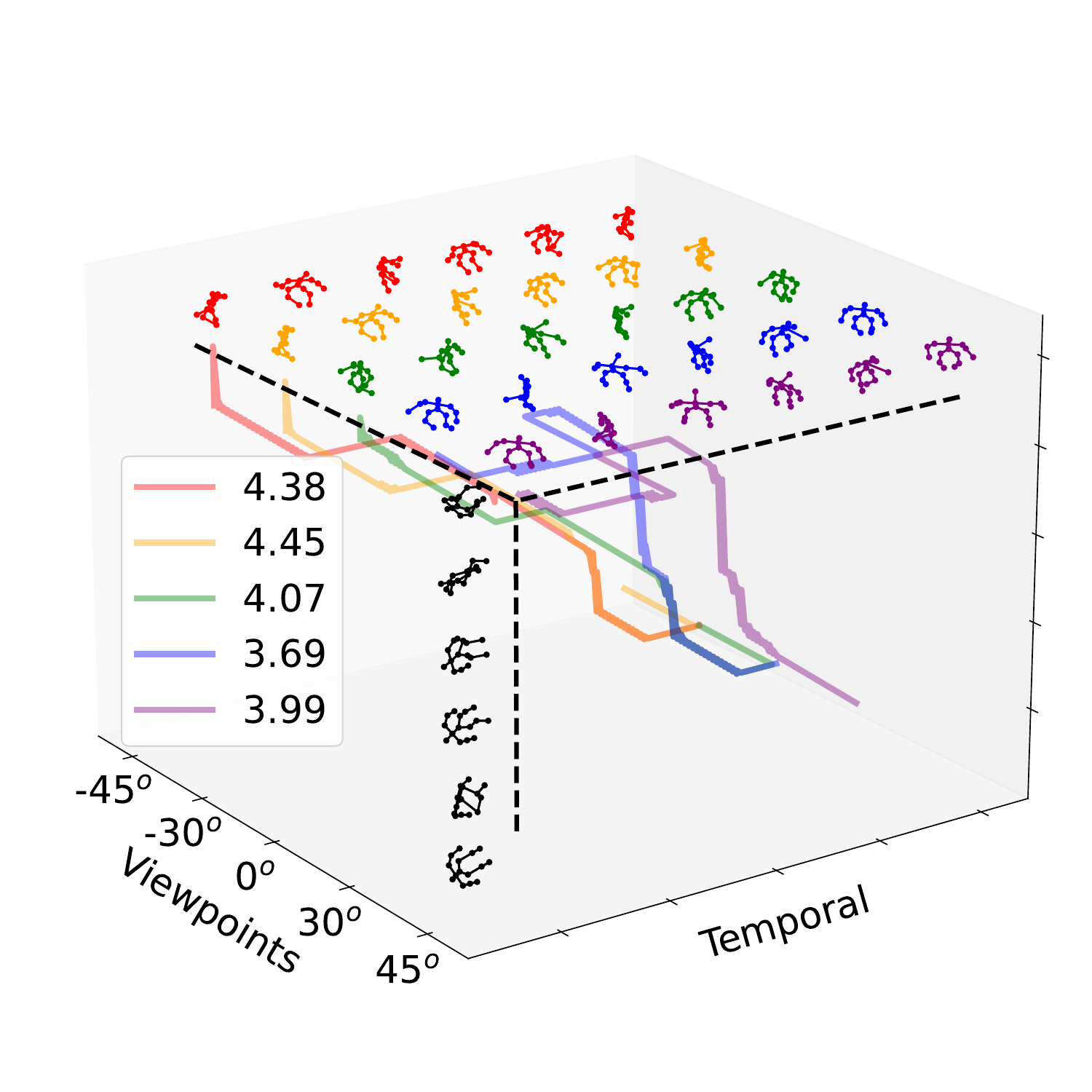}
\caption{JEANIE (1-max shift)}\label{fig:jeanie-v2}
\end{subfigure}
\caption{A comparison of paths in 3D for soft-DTW, Free Viewpoint Matching (FVM) and our JEANIE.
For a given query skeleton sequence (green color), we choose viewing angles between $-45^\circ$ and $45^\circ$ for the camera viewpoint simulation. The support skeleton sequence is shown in black color.
(a) soft-DTW finds each individual alignment per viewpoint fixed throughout alignment:  $d_\text{shortest}\!=\!4.08$. Notice that each path ``stays'' within the same view--it does not cross into other views. (b) FVM is a greedy matching algorithm that in each time step  seeks the best alignment pose from all viewpoints which leads to unrealistic zigzag path (person cannot jump from front to back view suddenly): $d_\text{FVM}\!=\!2.53$. (c) Our JEANIE (1-max shift) is able to find smooth joint viewpoint-temporal alignment between support and query sequences. We show each optimal path for each possible starting position: $d_\text{JEANIE}\!=\!3.69$. While $d_\text{FVM}\!=\!2.53$ for FVM is overoptimistic, $d_\text{shortest}\!=\!4.08$ for fixed-view matching is too pessimistic, whereas JEANIE strikes the right matching balance with $d_\text{JEANIE}\!=\!3.69$.}
\label{fig:jeanie_fvm_plots}
\end{figure*}

To learn similarity and dissimilarity between pairs of sequences of 3D body joints representing query and support samples from episodes, our goal is to find a joint viewpoint-temporal alignment of query and support, and minimize or maximize the matching distance $d_\text{JEANIE}$ (end-to-end setting) for same or different support-query labels, respectively. 
Fig.~\ref{fig:sq-match} (top) shows that sometimes matching of query and support may be as easy as rotating one trajectory onto another, in order to achieve viewpoint invariance. 
A viewpoint invariant distance \cite{inv_kern} can be defined as:
\begin{equation}
d_{\text{inv}}(\mPsi, \mPsi')\!=\!\myinf\limits_{\gamma,\gamma'\in T} d\big(\gamma(\mPsi), \gamma'(\mPsi')\big),
\label{eq:inv}
\end{equation}
where $T$ is a set of transformations required to achieve a viewpoint invariance, $d(\cdot,\cdot)$ is some base distance, \eg, the Euclidean distance, and $\mPsi$ and $\mPsi'$ are features describing query and support pair of sequences. Typically, $T$ may include 3D rotations to rotate one trajectory onto the other. However, a global viewpoint alignment of two sequences is suboptimal. Trajectories are unlikely to be straight 2D lines in the 3D space so one may not be able to rotate the query trajectory to align with the support trajectory. Fig.~\ref{fig:sq-match} (bottom) shows that the subjects' poses locally follow complicated non-linear paths.

Thus, we propose JEANIE that aligns and warps query / support sequences based on the feature similarity. One can think of JEANIE as performing Eq. \eqref{eq:inv} with $T$ containing all possible combinations of local time-warping augmentations of sequences and camera pose augmentations for each frame (or temporal block). 
JEANIE unit  in Fig. \ref{fig:pipe} realizes such a strategy. Figure \ref{fig:jeanie_temp-shift} (discussed later in the text) shows one step of the temporal-viewpoint computations of JEANIE in search for optimal temporal-viewpoint alignment path between query and support sequences. 
Soft-minimum across all such possible alignment paths can be equivalently written as an  infimum over a set of specific transformations in Eq. \eqref{eq:inv}. 

Below, we detail our pipeline, and explain the proposed JEANIE, Encoding Network (EN), feature coding and dictionary learning, and our loss function. Firstly, we present our notations.

\noindent\textbf{Notations.}  $\idx{K}$ stands for the index set $\{1,2,\cdots,K\}$. Concatenation of $\alpha_i$ is denoted by $[\alpha_i]_{i\in\idx{I}}$, whereas $\mX_{:,i}$ means we extract/access column $i$ of matrix $\mD$. Calligraphic mathcal fonts denote tensors (\eg, $\tD$), capitalized bold symbols are matrices (\eg, $\mD$), lowercase bold symbols  are vectors (\eg, $\vpsi$), and regular fonts denote scalars.

\vspace{0.1cm}
\noindent{\bf Prerequisites.} Below we refer to prerequisites used in the subsequent chapters. 
Appendix \ref{supp:prereq_euler} explains how Euler angles and stereo projections are used in simulating different skeleton viewpoints.
 Appendix \ref{sec:gnn} explains  several GNN approaches that we use in our Encoding Network.
 Appendix \ref{sec:feat_code} explains several feature coding and dictionary learning strategies which we use for unsupervised FSAR.

\begin{sloppypar}

\subsection{Encoding Network (EN)}  
\label{sec:en}

We start by generating $K\!\times\!K'$ Euler rotations or $K\!\times\!K'$ simulated camera views (moved gradually from the estimated camera location) of query skeletons. 
Our EN contains a simple 3-layer MLP unit (FC, ReLU, FC, ReLU, Dropout, FC), GNN, optional  Transformer~\cite{dosovitskiy2020image} and FC. 
The MLP unit takes $M$ neighboring frames, each with $J$ 3D skeleton body joints, forming one temporal block $\mX\!\in\!\mbr{3\times J\times M}$, where $3$ indicates 3D Cartesian coordinates. In total, depending on stride $S$, we obtain some $\tau$ temporal blocks which capture the short temporal dependency, whereas the long temporal dependency is modeled with our JEANIE. Each temporal block is encoded by the MLP into a $d\!\times\!J$ dimensional feature map:
\begin{equation}
    \widehat{\mX}\!=\!\left(\text{MLP}(\mX; \mathcal{F}_{MLP})\right)^T\!\in\!\mbr{J\times d}.
    \label{eq:mlp_input}
\end{equation}
We obtain  $K\!\times\!K'\!\times\!\tau$ query and $\tau'$ support feature maps, each of size $J\times d$. Each maps is forwarded to a GNN. For  S$^2$GC \cite{hao2021iclr} (default GNN in our work) with $L$ layers, we have:
%
\begin{equation}
    \widehat{\widehat{\mX}} \!=\!  \frac{1}{L}\sum_{l=1}^{L}\big((1\!-\!\alpha){\bf S}^l\widehat{\mX}\!+\!\alpha\widehat{\mX}\big)\!\in\!\mbr{J\times d},
\end{equation}
where ${\bf S}$ is the adjacency matrix capturing connectivity of body joints, whereas $0\leq\alpha\leq 1$ controls the self-importance of each body joint. Appendix \ref{sec:gnn} describes several GNN  variants we experimented with: GCN~\cite{kipf2017semi}, SGC~\cite{felix2019icml}, APPNP \cite{johannes2019iclr} and S$^2$GC~\cite{hao2021iclr}.

Optionally, a transformer\footnote{Our transformer is similar to ViT~\cite{dosovitskiy2020image} but instead of using image patches, we feed each body joint encoded by GNN into the transformer.} (described below in ``Transformer Encoder'') may be used. Finally, an FC layer returns $\mPsi\!\in\!\mbr{d'\times K\times K'\times\tau}$ query feature maps and $\mPsi'\!\in\!\mbr{d'\times\tau'}\!$ support feature maps.
Feature maps are  passed to JEANIE whose output is passed into the similarity classifier. The whole Encoding Network is summarized as follows. 
%
Let support  maps  $\mPsi'\!\equiv\![f(\boldsymbol{X}'_1;\mathcal{F}),\cdots,f(\boldsymbol{X}'_{\tau'};\mathcal{F})]\!\in\!\mbr{d'\times\tau'}$
and query maps $\mPsi\!\equiv\![f(\boldsymbol{X}_{1,1,1};\mathcal{F}),\cdots,f(\boldsymbol{X}_{K,K',\tau};\mathcal{F})]\!\in\!\mbr{d'\times K\times K'\times\tau}$. For $M$ 
query and $M$ support frames per block, $\mX\!\in\!\mbr{3\times J\times M}$ and $\mX'\!\in\!\mbr{3\times J\times M}$. We also define:
%
%
\begin{align}
\label{eq:backbone}
&f(\mX; \mathcal{F})\!=\\
&\text{FC}(\text{Transf}(\text{GNN}(\text{MLP}(\mX; \mathcal{F}_{MLP});  \mathcal{F}_{GNN}); \mathcal{F}_{Tr}); \mathcal{F}_{FC}), \nonumber
\end{align}
where $\mathcal{F}\!\equiv\![\mathcal{F}_{MLP},\mathcal{F}_{GNN},\mathcal{F}_{Tr},\mathcal{F}_{FC}]$ is the  set of  parameters of EN (including an optional transformer).

\vspace{0.1cm}
\noindent\textbf{Transformer Encoder.} Vision transformer  \cite{dosovitskiy2020image} consists of alternating layers of Multi-Head Self-Attention (MHSA) and a feed-forward MLP (2 FC layers with a GELU non-linearity intertwined). LayerNorm (LN) is applied before every block, and residual connections after every block. If transformer is used, each  feature matrix $\widehat{\mX} \in \mathbb{R}^{J \times {d}}$ per temporal block is encoded by a GNN into $\widehat{\widehat{\mX}} \in \mathbb{R}^{J \times {d}}$ and then passed to the transformer. Similarly to the standard transformer, we prepend a learnable vector  
${\bf y}_\text{token}\!\in\!\mbr{1\times {d}}$
to the sequence of block features $\widehat{\mX}$ obtained from GNN,
and we also add the positional embeddings ${\bf E}_\text{pos} \in \mathbb{R}^{(1+J) \times {d}}$ based on the standard sine and cosine functions  so that token ${\bf y}_\text{token}$ and each body joint enjoy their own unique positional encoding.
One can think of our GNN block as replacing the tokenizer linear projection layer of a standard transformer. Compared to the use of FC layer as linear projection layer, our GNN tokenizer in Eq.~\eqref{eq:proj} enjoys (i) better embeddings of human body joints based on the graph structure (ii) no learnable parameters.
From the tokenizer, we obtain  $\mZ_0\!\in\!\mbr{(1+J)\times {d}}$:
\begin{align}
&{\bf Z}_0 = [{\bf y}_\text{token}; \text{GNN}(\widehat{\mX})]+{\bf E}_\text{pos}, \label{eq:proj}
\end{align}
 and feed it into in the following transformer backbone:
 \begin{align}
&{\bf Z}^\prime_k = \text{MHSA}(\text{LN}({\bf Z}_{k-1})) + {\bf Z}_{k-1}, \;k = 1, \cdots, L_\text{tr}\label{eq:mhsa}\\
& {\bf Z}_k = \text{MLP}(\text{LN}({\bf Z}^\prime_k)) + {\bf Z}^\prime_k, \qquad\quad \,k = 1, \cdots, L_\text{tr}\label{eq:mlp} \\
& \vy' = \text{LN}\big(\mZ^{(0)}_{L_\text{tr}}\big) \qquad\qquad\text{ where }\quad\;\; \vy'\in \mathbb{R}^{1 \times {d}} \label{eq:blockfeat}\\
&f(\mX; \mathcal{F})=\text{FC}(\vy'^T; \mathcal{F}_{FC})\qquad\qquad\quad\!\in \mathbb{R}^{d'},\label{eq:final_fc_bl}
\end{align}
where $\mZ^{(0)}_{L_\text{tr}}$ is the first ${d}$-dimensional row vector extracted from the output matrix $\mZ_{L_\text{tr}}\!\in\!\mbr{(J\!+\!1)\times{d}}$, and  $L_\text{tr}$  controls the depth of the transformer (the number of layers), whereas $\mathcal{F}\!\equiv\![\mathcal{F}_{MLP},\mathcal{F}_{GNN},\mathcal{F}_{Tr},\mathcal{F}_{FC}]$ is the  set of  parameters of EN. Finally, $f(\mX; \mathcal{F})$ from Eq. \eqref{eq:final_fc_bl} becomes equivalent of  Eq. \eqref{eq:backbone} with the transformer.

\end{sloppypar}

\subsection{JEANIE}
\label{sec:jeanie}

\begin{sloppypar}
    
Prior to explaining the details of the JEANIE measure, we briefly explain details of soft-DTW.

\vspace{0.1cm}
\noindent{\bf Soft-DTW} \cite{marco2011icml,marco2017icml}. Dynamic Time Warping can be seen as a specialized  ``metric'' with a matching transportation plan\footnote{In analogy to terminology used in Optimal Transport (\eg, the Wasserstein distance), we call it a transportation plan. Also, notice that Soft-DTW may violate some of the metric axioms.} acting on the temporal mode of sequences. Soft-DTW is defined as:
\begin{align}
& d_{\text{DTW}}(\mPsi,\mPsi')\!=\!\softming\limits_{\mA\in\tAnb_{\tau,\tau'}}\big\langle\mA,\mD(\mPsi,\mPsi')\big\rangle,\\
& \text{ where } \softming(\valpha)\!=\!-\gamma\!\log\sum_i\exp(-\alpha_i/\gamma).
\end{align}
The binary  $\mA\!\in\!\tAnb_{\tau,\tau'}$ encodes a path within the transportation plan $\tAnb_{\tau,\tau'}$ which depends on lengths $\tau$ and $\tau'$ of sequences $\mPsi\!\equiv\![\vpsi_1,\cdots,\vpsi_\tau]\!\in\!\mbr{d'\times\tau}$, $\mPsi'\!\equiv\![{\vpsi'}_1,\cdots,{\vpsi'}_{\tau'}]\!\in\!\mbr{d'\times\tau'}$.  $\mD\!\in\!\mbrp{\tau\times\tau'}\!\!\equiv\![d_{\text{base}}(\vpsi_m,\vpsi'_n)]_{(m,n)\in\idx{\tau}\times\idx{\tau'}}$ is the matrix of distances, evaluated for $\tau\!\times\!\tau'$ frames (or temporal blocks) according to some base distance $d_{\text{base}}(\cdot,\cdot)$, \ie, the Euclidean  distance. 

In what follows, we make use of principles of soft-DTW, \ie, the property of time-warping. However, we design a joint alignment between temporal skeleton sequences and simulated skeleton viewpoints, which means we achieve joint time-viewpoint warping (a novel idea never done before).

\vspace{0.1cm}
\noindent{\bf JEANIE.}
Matching query-support pairs requires temporal alignment due to potential offset in locations of discriminative parts of actions, and due to potentially different dynamics/speed of actions taking place. The same concerns the direction of actor's pose, \ie, consider the pose trajectory \wrt ~the camera. Thus, the JEANIE measure is equipped with an extended  transportation plan  $\tAnb'\!\equiv\!\tAnb_{\tau,\tau', K, K'}$, where  apart from temporal block counts $\tau$ and $\tau'$, for query sequences, we have possible $\eta_{az}$ left and $\eta_{az}$ right steps from the initial camera azimuth, and $\eta_{alt}$ up and $\eta_{alt}$ down  steps from the initial camera altitude. Thus, $K\!=\!2\eta_{az}\!+\!1$, $K'\!=\!2\eta_{alt}\!+\!1$.  For the variant with Euler angles, we simply have $\tAnb''\!\equiv\!\tAnb_{\tau,\tau', K, K'}$ where $K\!=\!2\eta_{x}\!+\!1$, $K'\!=\!2\eta_{y}\!+\!1$ instead. The JEANIE formulation is given as:
%
\begin{equation}
d_{\text{JEANIE}}(\mPsi,\mPsi')\!=\!\softming\limits_{\mA\in\tAnb'}\big\langle\mA,\tD(\mPsi,\mPsi')\big\rangle,
\label{eq:d_jeanie}
\end{equation}
\noindent
where  $\tD\!\in\!\mbrp{K\times\!K'\!\times\tau\times\tau'}\!\!\!\equiv\![d_{\text{base}}(\vpsi_{m,k,k'},\vpsi'_n)]_{\substack{ (m,n)\in\idx{\tau}\!\times\!\idx{\tau'}\\(k,k')\in\idx{K}\!\times\!\idx{K'}}},\!\!\!\!\!\!$
and tensor $\tD$ contains distances evaluated between all possible temporal blocks.
\end{sloppypar}

\algblock{while}{endwhile}
\algblock[TryCatchFinally]{try}{endtry}
\algcblockdefx[TryCatchFinally]{TryCatchFinally}{catch}{endtry}
	[1]{\textbf{except}#1}{}
\algcblockdefx[TryCatchFinally]{TryCatchFinally}{elsee}{endtry}
	[1]{\textbf{else}#1}{}
\algtext*{endwhile}
\algtext*{endtry}

\algblock{for}{endfor}
\algtext*{endfor}

\algblockdefx{ifff}{endifff}
	[1]{\textbf{if}#1}{}
\algtext*{endifff}

\algblockdefx{elseee}{endelseee}
	[1]{\textbf{else}#1}{}
\algtext*{endelseee}

\begin{algorithm}[tbp!]
\caption{Joint tEmporal and cAmera viewpoiNt alIgnmEnt (JEANIE).}
\label{code:JEANIE}
{\bf Input} (forward pass): $\mPsi, \mPsi'$, $\gamma\!>\!0$, $d_{\text{base}}(\cdot,\cdot)$, $\iota$-max shift.
\begin{algorithmic}[1]
\State{$r_{:,:,:}\!=\!\infty$, $r_{n,1,1}\!=\!d_{\text{base}}(\vpsi_{n,1}, \vpsi'_{1}),\;\forall n\!\in\!\{-\eta, \cdots, \eta\}$}
\State{$\Pi\equiv\{-\iota,\cdots,0,\cdots,\iota\}\times\{(0,1),(1,0),(1,1)\}$}
\for{ $t\!\in\!\idx{\tau}$:}
\for{ $t'\!\in\!\idx{\tau'}$:}
\ifff{ $t\!\neq\!1$ or $t'\!\!\neq\!1$:}
\for{ $n\!\in\!\{-\eta,\cdots,\eta\}$:}
	\State{$r_{n,t,t'}=d_{\text{base}}(\vpsi_{n,t}, \vpsi'_{t'})$}
    \State{$+\softming\Big([r_{n\!-\!i,t\!-\!j,t'\!-\!k}]_{(i,j,k)\in\Pi}\Big)$}
\endfor
	\endifff
\endfor
\endfor
\end{algorithmic}
{\bf Output:} $\softming\Big([r_{n,\tau,\tau'}]_{n\in\{-\eta,\cdots,\eta\}}\Big)$
\end{algorithm}

\begin{sloppypar}
Figure \ref{fig:jeanie_temp-shift} illustrates one step of JEANIE. Suppose the given viewing angle set is $\{-40^{\circ}, -20^{\circ}, 0^{\circ}, 20^{\circ}, 40^{\circ}\}$. 
For the  current node at $(t,t'\!,n)$ we evaluate, we have to aggregate its base distance with the smallest aggregated distance of its predecessor nodes. The ``1-max shift'' means that the predecessor node must be a direct neighbor of the current node (imagine that dots on a 3D grid are nodes connected by links).  
Thus, for 1-max shift, at location $(t,t'\!,n)$, we extract the node's base distance and add it together with the minimum of aggregated distances at the shown 9 predecessor nodes. We store that aggregated distance at $(t,t'\!,n)$, and we move to the next node. Note that for viewpoint index $n$, we look up $(n\!-\!1,n,n\!+\!1)$ neighbors. 
Extension to the $\iota$-max shift is straightforward. The importance of low value of $\iota$-max shift, \eg, $\iota=1$ is that low value of $\iota$ promotes the so-called smoothness of alignment. That is, while time or viewpoint may be warped, they are not warped abruptly (\eg, the subject's pose is not allowed to suddenly rotate by $90^{\circ}$ in one step then rotate back by $-90^{\circ}$. This smoothness is the key preventing greedy matching that would result in an overoptimistic distance between two sequences.

Algorithm \ref{code:JEANIE} illustrates JEANIE. For brevity, let us tackle the camera viewpoint alignment along the azimuth, \eg, for some shifting steps $-\eta,\cdots,\eta$, each with  size  $\Delta\theta_{az}$. The maximum viewpoint change from block to block is $\iota$-max shift (smoothness). As we have no way to know the initial optimal camera shift, we initialize all possible origins of shifts in accumulator $r_{n,1,1}\!=\!d_{\text{base}}(\vpsi_{n,1}, \vpsi'_{1})$ for all $n\!\in\!\{-\eta, \cdots, \eta\}$. Subsequently, steps related to soft-DTW (temporal-viewpoint matching) take place. Finally, we choose the path with the smallest distance  over all possible 
viewpoint ends by selecting a soft-minimum over  $[r_{n,\tau,\tau'}]_{n\in\{-\eta, \cdots, \eta\}}$. Notice that elements of the accumulator tensor $\boldsymbol{\mathcal{R}}\in\mbr{(2\iota+1)\times\tau\times\tau'}$ are accessed by writing $r_{n,t,t'}$. Moreover, whenever either index $n\!-\!i$, $t\!-\!j$ or $t'\!-\!k$ in $r_{n\!-\!i,t\!-\!j,t'\!-\!k}$ (see algorithm) is out of bounds, we define $r_{n\!-\!i,t\!-\!j,t'\!-\!k}=\infty$.

\end{sloppypar}

\begin{sloppypar}
\vspace{0.1cm}
\noindent{\bf Free Viewpoint Matching (FVM).}  To ascertain whether JEANIE is better than performing separately the temporal and simulated viewpoint alignments, we introduce an important and plausible baseline called Free Viewpoint Matching. FVM, for every step of DTW, seeks the best local viewpoint alignment, thus realizing a non-smooth temporal-viewpoint path in contrast to JEANIE. To this end, we apply soft-DTW in Eq.~\eqref{eq:d_jeanie} with the base distance replaced by:
%
\begin{align}
& d_{\text{FVM}(\vpsi_{t},\vpsi'_{t'})}\!=\!\softming\limits_{m,n\in\{-\eta,\cdots,\eta\}} d_{\text{base}}\big(\vpsi_{m,n,t},\vpsi'_{m',n',t'}\big),
\label{eq:suppl1}
\end{align}
where $\mPsi\!\in\!\mbr{d'\times K\times K'\times\tau}$  and $\mPsi'\!\in\!\mbr{d'\times K\times K'\times\tau'}\!$ are query and support feature maps. We abuse slightly the notation by writing $d_{\text{FVM}(\vpsi_{t},\vpsi'_{t'})}$ as we minimize over viewpoint indexes inside of Eq. \eqref{eq:suppl1}. Thus, we calculate the distance matrix $\mD\!\in\!\mbrp{\tau\times\tau'}\!\!\equiv\![d_{\text{FVM}}(\vpsi_t,\vpsi'_{t'})]_{(t,t')\in\idx{\tau}\times\idx{\tau'}}$ for soft-DTW.

Fig.~\ref{fig:jeanie_fvm_plots} shows the comparison between soft-DTW (view-wise), FVM and our JEANIE. FVM is a greedy matching method which leads to complex zigzag path in 3D space (we illustrate the camera viewpoint in a single mode, \eg, the azimuth for $\vpsi_{n,t}$, and no viewpoint mode for $\vpsi'_{t'}$). Although FVM is able to produce the path with a smaller aggregated distance compared to soft-DTW and JEANIE, it suffers from obvious limitations: (i) It is unreasonable for poses in a given sequence to match under extreme sudden changes of viewpoints.  (ii) Even if two sequences are from two different classes, FVM still yields the smallest distance (decreased inter-class variance). 
\end{sloppypar}

\subsection{Loss Function for Supervised FSAR}
\begin{sloppypar}
For the $N$-way $Z$-shot problem, 
we have   one query feature map  and $N\!\times\!Z$ support feature maps per episode. We form a mini-batch containing $B$ episodes. 
Thus, we have query feature maps $\{\mPsi_b\}_{b\in\idx{B}}$ and support feature maps $\{\mPsi'_{b,n,z}\}_{\substack{b\in\idx{B}\\n\in\idx{N}\\z\in\idx{Z}}}$. Moreover,  $\mPsi_b$ and $\mPsi'_{b,1,:}$ share the same class, one of $N$ classes drawn per episode, forming the  subset $C^{\ddagger} \equiv \{c_1,\cdots,c_N \} \subset \mathcal{I}_C \equiv \mathcal{C}$. 

Specifically, labels  $y(\mPsi_b)\!=\!y(\mPsi'_{b,1,z}), \forall b\!\in\!\idx{B}, z\!\in\!\idx{Z}$ while $y(\mPsi_b)\!\neq\!y(\mPsi'_{b,n,z}), \forall b\!\in\!\idx{B},n\!\in\!\idx{N}\!\setminus\!\{1\},  z\!\in\!\idx{Z}$. In most cases, $y(\mPsi_b)\!\neq\!y(\mPsi_{b'})$ if $b\!\neq\!b'$ and $b,b'\!\in\!\idx{B}$. Selection of $C^{\ddagger}$ per episode is  random. For the $N$-way $Z$-shot protocol, we minimize:
\begin{align}
 & l(\vd^{+}\!,\vd^{-})\!=\!\left(\mu(\vd^{+})\!-\!\{\mu(\topminb(\vd^{+}))\}\right)^2\label{eq:pos}\\
 & \qquad\qquad\!+\!\left(\mu(\vd^{-})\!-\!\{\mu(\topmaxbb(\vd^{-}))\}\right)^2\!,\label{eq:neg}\\
 & 
 \quad\text{where}\quad
 \begin{cases}
 \vd^{+}\!=\![d_\text{JEANIE}(\mPsi_{b},\mPsi'_{b,1,z})]_{\substack{b\in\idx{B}\\z\in\idx{Z}}}&\\
 \vd^{-}\!=\![d_\text{JEANIE}(\mPsi_{b},\mPsi'_{b,n,z})]_{\!\!\!\!\!\!\!\!\!\!\!\substack{b\in\idx{B}\\\;\;\quad n\in\idx{N}\!\setminus\!\{1\}\\z\in\idx{Z}}},
 \end{cases}
 \nonumber
\end{align}
and $\vd^+$ is a set of within-class distances for the mini-batch of size $B$ given $N$-way $Z$-shot learning protocol. By analogy,  $\vd^-$ is a set of between-class distances. Function $\mu(\cdot)$ is simply the mean over coefficients of the input vector, $\{\cdot\}$ detaches the graph during the backpropagation step, whereas $\topminb(\cdot)$ and $\topmaxbb(\cdot)$ return $\beta$ smallest and $NZ\beta$ largest coefficients from the input vectors, respectively. Thus, Eq. \eqref{eq:pos} promotes the within-class similarity while Eq. \eqref{eq:neg} reduces the between-class similarity. Integer $\beta\!\geq\!0$ controls the focus on difficult examples, \eg, $\beta\!=\!1$  encourages all within-class distances in Eq.  \eqref{eq:pos} to be close to the positive target $\mu(\topminb(\cdot))$, the smallest observed within-class distance in the mini-batch. If $\beta\!>\!1$, this means we relax our positive target.  By analogy, if $\beta\!=\!1$, we encourage all between-class distances in Eq.  \eqref{eq:neg} to approach the negative target $\mu(\topmaxbb(\cdot))$, the average over the largest $NZ$ between-class distances. If $\beta\!>\!1$, the negative target is relaxed.

\end{sloppypar}

\subsection{Feature Coding and Dictionary Learning for Unsupervised FSAR}
\label{sec:unsupervised}

\begin{algorithm}[tbp!]
\caption{Unsupervised FSAR (one training iteration by alternating over variables).}
\label{code:unsup}
{\bf Input}: $ \Upsilon\equiv\{\mPsi_{b}\}_{b\in\idx{B}}\cup\{\mPsi'_{b,n,z}\}_{\substack{b\in\idx{B}\\n\in\idx{N}\\z\in\idx{Z}}}$: query/support seq. in  batch; $\mathcal{F}$: EN parameters; $\mM$ and $\malpha$; \texttt{alpha\_iter} and \texttt{dic\_iter}: numbers of iterations for updating $\malpha$ and $\mM$; $\omega, \omega_\text{DL}$ and $\omega_\text{EN}$: the learning rate for $\malpha, \mM$ and $\mathcal{F}$ respectively; $B$: size of the mini-batch. 
\begin{algorithmic}[1]
	\for { i = 1, \cdots, \texttt{alpha\_iter}: \gray{$\;\qquad\quad\qquad$(fix $\mM$ and update $\malpha$)}}
	   \State{$\malpha := \malpha - \omega\nabla_{\malpha} \mathcal{L}_\text{unsup}(\Upsilon; \malpha,\mM,\mathcal{F})$}
	\endfor
	\for { i = 1, \cdots, \texttt{dic\_iter}: \gray{$\;\;\qquad\qquad\qquad$(fix $\malpha$ and update $\mM$)}}
	   \State{$\mM\!:=\!\mM\!-\!\omega_\text{DL}\nabla_{\mM} \mathcal{L}_\text{unsup}(\Upsilon; \malpha,\mM,\mathcal{F})$ }
	\endfor
    \State{ $\mathcal{F} := \mathcal{F}\!-\!\omega_\text{EN}\nabla_{\mathcal{F}}\mathcal{L}_\text{unsup}(\Upsilon; \malpha,\mM,\mathcal{F})$ \gray{$\!\!\!\quad$(fix $\mM$ \& $\malpha$, update $\mathcal{F}$)}}
\end{algorithmic}
{\bf Output:} $\mathcal{F}$ and $\mM$
\end{algorithm}

Recall from Section \ref{sec:intro} that unsupervised FSAR  forms a dictionary from the training data without the use of labels. Assigning labeled test support samples and test query into cells of a dictionary lets infer the query label by associating query with the support sample (to paraphrase, if they share the same dictionary cell, they share the class label).

\begin{figure}[t]
\centering

\includegraphics[trim=0 0 0 0, clip=true,width=0.9\linewidth]{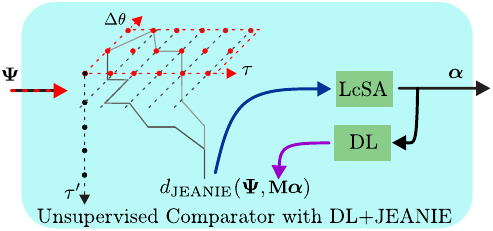}%
\caption{Unsupervised FSAR uses the JEANIE measure as a distance between feature map $\mPhi$ of a sequence and its dictionary-based reconstruction $\mM\valpha$. LcSA performs feature coding to obtain dictionary-coded $\valpha$. DL learns the dictionary $\mM$.
}
\label{fig:unsup}
\end{figure}

In this setting, we also use a mini-batch with $B$ episodes. Thus, $B$ query samples and $BNZ$ support samples give the total of $N'\!=\!B(NZ\!+\!1)$ samples per batch  
for feature coding and dictionary learning. 
Let dictionary $\mM\!\in\!\mbr{d'\cdot\tau^* \times k}$ and  dictionary-coded matrix $\malpha\equiv[\valpha_1,\cdots,\valpha_{N'}] \!\in\! \mbr{k \times N'}$. Let $\tau^*$ be set as the average number of temporal blocks over training sequences. For dictionary $\mM$ and some codes $\malpha$, the reconstructed feature map is given as $\mM\malpha\in\mbr{d' \cdot \tau^* \times N'}$. In what follows we reshape the reconstructed feature map so that $\mM\malpha\in\mbr{d' \times \tau^* \times N'}$. The  feature map per sequence is given as $\mPsi \!\in\! \mbr{d'\times K\times K'\times\tau \times N'}$. All query and support sequences per batch form a set $ \Upsilon\equiv\{\mPsi_{b}\}_{b\in\idx{B}}\cup\{\mPsi'_{b,n,z}\}_{\substack{b\in\idx{B}\\n\in\idx{N}\\z\in\idx{Z}}}$ with $N'$ feature maps which we select by writing $\mPsi_{i}\in \Upsilon$ where $i=1,\cdots,N'$. They are obtained from Encoding Network the same way as for supervised FSAR except that both query and support sequences now are equipped with $K\times K'$ viewpoints. 
Algorithm~\ref{code:unsup} and Figure \ref{fig:unsup} illustrate unsupervised FSAR learning with JEANIE. In short, we minimize the following loss \wrt ~$\mathcal{F}$, $\mM$ and $\malpha$ by alternating over these variables:
\begin{align}
    & \mathcal{L}_\text{unsup}(\Upsilon; \malpha,\mM,\mathcal{F}) \!=\!\nonumber\\
    & \quad\sum_{i=1}^{N'}d^2_{\text{JEANIE}}(\mPsi_i(\mathcal{F}), \mM\valpha_i) + \kappa \mOmega(\valpha_i(\mathcal{F}), \mM, \mPsi_i), \label{eq:unsup}
\end{align}
where $\mathcal{F}\!\equiv\![\mathcal{F}_{MLP},\mathcal{F}_{GNN},\mathcal{F}_{Tr},\mathcal{F}_{FC}]$ is the  set of  parameters of EN associated with $\mPsi$, that is, feature maps depend on these parameters, \ie, we work with a function $\mPsi(\mathcal{F})$ not a constant.

\begin{sloppypar}
Similarly to the Euclidean distance, $d_{\text{JEANIE}}(\cdot,\cdot)$ in Eq. \eqref{eq:unsup} pursues the reconstruction of the feature map $\mPsi_i$ by the linear combination of dictionary codewords, given as $\mM\valpha_i$. The  reconstruction error $d^2_{\text{JEANIE}}(\mPsi_i, \mM\valpha_i)$ is encouraged to be small. However, unlike the Euclidean distance, JEANIE ensures temporal and viewpoint alignment of sequences $\mPsi_i$ with the dictionary-based reconstruction $\mM\valpha_i$. Constraint $\mOmega(\valpha_i, \mM, \mPsi_i)$ is a regularization term depending on the selection of feature coding method. Such a regularization encourages discriminative description, \ie, similar and different feature vectors obtain similar and different dictionary-coded representations, respectively. Appendix \ref{sec:feat_code} provides details of several feature coding and dictionary learning strategies which determine $\mOmega$. In our work, the default choice is Soft Assignment and Dictionary learning from Appendices \ref{sec:feat_coding} and  \ref{sec:dic} due to their simplicity and good performance. 
As the Soft Assignment code \cite{Ni_2022_CVPR} was adapted to use JEANIE, we kept their number of iterations $\texttt{alpha\_iter}\!=\!50$, $\texttt{dic\_iter}\!=\!5$. Dictionary size $k\!=\!4096$ was optimal, whereas $\tau^*$ ranged between 30 and 60 for smaller and larger datasets, respectively.
\end{sloppypar}

During  testing, we  use the trained model $\mathcal{F}$ and the learnt dictionary $\mM$, pass test support and query sequences via Eq. \eqref{eq:unsup} but solve only \wrt ~$\malpha$ by  till $\malpha$ converges. 
Subsequently, we compare the dictionary-coded vectors of query sequences with the corresponding dictionary-coded vectors of support sequences 
by using some distance measure, \eg, the $\ell_1$ or  $\ell_2$ norm. We also explore the use of kernel-based distances, \eg, Histogram Intersection Kernel (HIK) distance and Chi-Square Kernel (CSK) distance, as they are designed for comparing  vectors constrained on the $\ell_1$ simplex (Soft Assignment produces  the $\ell_1$ normalised codes $\boldsymbol{\alpha}$). 
The construction of the kernel distance involves a transformation from similarities to distances. 

Let $\valpha$ and $\valpha'$ be some dictionary-coded vectors
obtained by the use of JEANIE in Eq. \eqref{eq:unsup}. 
Then for a kernel function $k(\valpha, \valpha')$, the induced distance between $\valpha$ and $\valpha'$ is given by 
   $ d(\valpha, \valpha') = k(\valpha, \valpha) + k(\valpha', \valpha') - 2k(\valpha, \valpha')$.
%
Let $\lVert\valpha\rVert_2\!=\!\lVert\valpha'\rVert_2\!=\!1$. 
The HIK distance for $k_\text{HIK}(\valpha, \valpha')\!=\!\sum_{i=1}^{d'}\text{min}(\alpha_i, \alpha'_i)$ is given as 
    $d_\text{HIK}(\valpha, \valpha')\! = \!2-2k_\text{HIK}(\valpha, \valpha')$.
The CSK distance for kernel $k_\text{CSK}(\valpha, \valpha')\!=\!\sum_{i=1}^{d'}\!\frac{2\alpha_i \alpha'_i}{\alpha_i + \alpha'_i}$ is 
    $d_\text{CSK}(\valpha, \valpha')\! = \!2\!-\!2k_\text{CSK}(\valpha, \valpha')$.

The closest nearest neighbor match of test query to elements of the test support set determines the test label of the query sequence.


\subsection{Fusion of Supervised and Unsupervised FSAR}
\label{sec:fused}

\begin{sloppypar}
Our final contribution is to introduce four simple strategies for fusing our supervised and unsupervised FSAR approaches to boost the performance. As supervised learning is label-driven and unsupervised learning is reconstruction-driven, we expect  both such strategies produce complementary feature spaces amenable to fusion. 

In what follows, we make use of both support and query feature maps defined over multiple viewpoints ($\mPsi,\mPsi'\!\in\!\mbr{d'\times K\times K'\times\tau}$):
\begin{align}
&\mPsi'\!\equiv\!f^*(\tX';\mathcal{F})\!\equiv\![f(\boldsymbol{X}'_{1,1,1};\mathcal{F}),\cdots,f(\boldsymbol{X}'_{K,K',\tau'};\mathcal{F})],\nonumber\\
&\mPsi\!\equiv\!f^*(\tX;\mathcal{F})\!\equiv\![f(\boldsymbol{X}_{1,1,1};\mathcal{F}),\cdots,f(\boldsymbol{X}_{K,K',\tau};\mathcal{F})].\nonumber
\end{align}

\end{sloppypar}

\vspace{0.1cm}
\noindent{\bf A weighted fusion of supervised and unsupervised FSAR scores.} The simplest strategy is to train supervised and unsupervised FSAR models separately, and combine their predictions during testing.  We call such a baseline as ``weighted fusion''. During the testing stage, we combine the distances of supervised and unsupervised models as follows:
\begin{equation}
    d_\text{fused}\!=\!\rho\, d_{\text{JEANIE}}(\mPsi_{q},\mPsi'_{n,z}) + (1-\rho) d_\alpha(\valpha_{q},\valpha'_{n,z}),
    \label{eq:fusion_simple}
\end{equation}
where $d_{\alpha}(\cdot, \cdot)$ is the distance measure for dictionary-encoded vectors, \eg, the $\ell_1$ norm, HIK distance or CSK distance, $0\leq\rho\leq 1$ balances the impact of supervised and unsupervised models, respectively. 

\vspace{0.1cm}
\noindent{\bf Finetuning unsupervised model by supervised FSAR}. For this baseline strategy, we firstly train the model using unsupervised FSAR, and then we finetune the learnt unsupervised model by using supervised FSAR. During testing stage, we evaluate on supervised learning, unsupervised learning and a fusion of both based on Eq. \eqref{eq:fusion_simple}. In this case, one EN is trained which results in two sets of parameters--the first set is based on unsupervised training and the second set is based on supervised finetuning. We call it ``finetuning unsup.''

\begin{algorithm}[tbp!]
\caption{Fusion of Supervised and Unsupervised FSAR by MAML-inspired Setting (one training iteration).}
\label{code:maml-fusion}
{\bf Input}: $ \Gamma\equiv\{\tX_{b}\}_{b\in\idx{B}}\cup\{\tX'_{b,n,z}\}_{\substack{b\in\idx{B}\\n\in\idx{N}\\z\in\idx{Z}}}$: query/support blocks in  batch; $\mathcal{F}$: EN parameters; $\mM$ and $\malpha$; \texttt{alpha\_iter} and \texttt{dic\_iter}: numbers of iterations for updating $\malpha$ and $\mM$; $\omega, \omega_\text{DL}$ and $\omega_\text{EN}$: the learning rate for $\malpha, \mM$ and $\mathcal{F}$ respectively; $B$: size of the mini-batch. 
\begin{algorithmic}[1]
\State{$\Upsilon\equiv\{\mPsi_{b}\}_{b\in\idx{B}}\cup\{\mPsi'_{b,n,z}\}_{\substack{b\in\idx{B}\\n\in\idx{N}\\z\in\idx{Z}}}\!$ where
$\begin{cases}
\mPsi_{b}\!=\!f^*(\tX_b;\mathcal{F})&\\
\mPsi'_{b,n,z}\!=\!f^*(\tX'_{b,n,z};\mathcal{F})\!\!\!\!\!\!\!\!\!&
\end{cases}$
\gray{(obtain feature maps for global parameters $\mathcal{F}$)}
}
\State{$(\widehat{\mathcal{F}},\mM)=\text{Algorithm\ref{code:unsup}}(\Upsilon, {\mathcal{F}},\mM, \malpha,$ \gray{$\quad\;\,$(unsupervised FSAR)}}
\Statex{$\qquad\qquad\qquad\texttt{alpha\_iter}, \texttt{dic\_iter}, \omega, \omega_\text{DL}, \omega_\text{EN})$}
\State{$\widehat{\Upsilon}\equiv\{\widehat{\mPsi}_{b}\}_{b\in\idx{B}}\cup\{\widehat{\mPsi}'_{b,n,z}\}_{\substack{b\in\idx{B}\\n\in\idx{N}\\z\in\idx{Z}}}\!$ where
$\begin{cases}
\widehat{\mPsi}_{b}\!=\!f^*(\tX_b;\widehat{\mathcal{F}})&\\
\widehat{\mPsi}'_{b,n,z}\!=\!f^*(\tX'_{b,n,z};\widehat{\mathcal{F}})\!\!\!\!\!\!\!\!\!&
\end{cases}$
\gray{(obtain feature maps for parameters $\widehat{\mathcal{F}}$ from the unsupervised step)}
}
%
%
%
%
\State{$\widehat{\vd}^{+}\!=\![d_{\text{JEANIE}}(\widehat{\mPsi}_{b},\widehat{\mPsi}'_{b,1,z})]_{\substack{b\in\idx{B}\\z\in\idx{Z}}}$ \gray{$\qquad\quad$(within-class distance)}}
\State{$\widehat{\vd}^{-}\!=\![d_\text{JEANIE}(\widehat{\mPsi}_{b},\widehat{\mPsi}'_{b,n,z})]_{\!\!\!\!\!\!\!\!\!\!\!\substack{b\in\idx{B}\\\;\;\quad n\in\idx{N}\!\setminus\!\{1\}\\z\in\idx{Z}}}$ \gray{$\quad\;\,$(between-class distance)}}
\State{${\mathcal{F}}\!:=\!{\mathcal{F}}\!- \!\omega_\text{EN}\nabla_{{\mathcal{F}}} l(\widehat{\vd}^{+},\widehat{\vd}^{-})$}
\end{algorithmic}
{\bf Output:} $\mathcal{F}$ and $\mM$
\end{algorithm}

\begin{sloppypar}
\vspace{0.1cm}
\noindent{\bf MAML-inspired fusion of supervised and unsupervised FSAR}. Inspired by the success of MAML~\cite{f5Model-Agnostic} and categorical learner \cite{zhibin_categorical}, we introduce a fusion strategy where the inner loop uses the unsupervised FSAR (Eq.~\eqref{eq:unsup}) and the outer loop uses the supervised learning loss (Eq.\eqref{eq:pos} and \eqref{eq:neg}) for the model update.  Algorithm~\ref{code:maml-fusion} details our MAML-inspired fusion strategy, called ``MAML-inspired fusion''. 

 Specifically, we start by generating  representations with several viewpoints.  For each mini-batch of size $B$ we form a set with $N'$ feature maps which are passed to Algorithm~\ref{code:unsup} which updates EN parameters $\mathcal{F}$ towards  $\widehat{\mathcal{F}}$ that help accommodate unsupervised reconstruction-driven learning (so-called task-specific gradient where the task is unsupervised learning). We then recompute $N'$ feature maps based on parameters $\widehat{\mathcal{F}}$. Finally, we apply supervised loss on such  feature maps but we update now  parameters ${\mathcal{F}}$ which means that  parameters ${\mathcal{F}}$ are tuned for the global label-driven task with help of unsupervised task. 

Intuitively, it is a second-order gradient model. Specifically, one takes the gradient step in the direction pointed by the unsupervised loss to obtain task-specific EN parameters. Subsequently, given these task-specific parameters, task-specific feature maps are extracted and passed into the supervised loss to perform the gradient descent step  in the direction pointed by the unsupervised loss to obtain update of global EN parameters. 
\end{sloppypar}

\begin{sloppypar}
\vspace{0.1cm}
\noindent{\bf Fusion by alignment of supervised and unsupervised feature maps.} Inspired by domain adaptation~\cite{domain_mixture,domain_openmic,tas2018cnnbased}, Algorithm \ref{code:fusion} in Appendix \ref{sec:fus_al} is an easy-to-interpret simplification (called ``adaptation-based'') of the above MAML-inspired fusion. Instead of complex gradient interplay between unsupervised and supervised loss functions, we explicitly align ``supervised'' feature maps towards ``unsupervised'' feature maps. 

\end{sloppypar}

\section{Experiments}

\subsection{Datasets and Protocols}

Below, we describe the datasets and evaluation protocols on which we  validate our FSAR with JEANIE.

\renewcommand{\labelenumi}{\roman{enumi}.}
\begin{enumerate}[leftmargin=0.6cm]
\item{{\em UWA3D Multiview Activity II}}~\cite{Rahmani2016} contains 30 actions performed by 9 people in a cluttered environment. The Kinect camera was used in 4 distinct views: front view ($V_1$), left view ($V_2$), right view ($V_3$), and top view ($V_4$). 

\item{{\em NTU RGB+D (NTU-60)}}~\cite{Shahroudy_2016_NTURGBD} contains 56,880 video sequences and over 4 million frames. 
This dataset has variable sequence lengths  and high intra-class variations.

\item{{\em NTU RGB+D 120 (NTU-120)}}~\cite{Liu_2019_NTURGBD120} 
contains 120 action classes (daily/health-related), and 114,480 RGB+D video samples  captured with 106 distinct human subjects from 155 different camera viewpoints. 

\item{{\em Kinetics}}~\cite{kay2017kinetics} is a large-scale collection of 
650,000 video clips that cover 400/600/700 human action classes. 
It includes human-object interactions such as {\it playing instruments}, as well as human-human interactions such as {\it shaking hands} and {\it hugging}. As the Kinetics-400 dataset provides only the raw videos, we follow approach~\cite{stgcn2018aaai} and use the estimated joint locations in the pixel coordinate system as the input to our pipeline. To obtain the joint locations, we first resize all videos to the resolution of 340 $\times$ 256, and convert the frame rate to 30 FPS. Then we use the publicly available {\it OpenPose}~\cite{Cao_2017_CVPR} toolbox to estimate the location of 18 joints on every frame of the clips. As OpenPose  produces the 2D body joint coordinates  
and Kinetics-400 does not offer multi-view or depth data, we use a network of Martinez \etal   \cite{martinez_2d23d} pre-trained on  
Human3.6M~\cite{Catalin2014Human3}, combined with the 2D OpenPose output to  estimate 3D coordinates from 2D coordinates. The 2D OpenPose  and the latter network give us $(x,y)$ and $z$ coordinates, respectively.
\end{enumerate}

\noindent{\bf Evaluation protocols.} 
For the UWA3D Multiview Activity II, we use standard multi-view classification protocol~\cite{Rahmani2016,lei_tip_2019}, but we apply it to one-shot learning as the view combinations for training and testing sets are disjoint. For NTU-120, we follow the standard one-shot protocol~\cite{Liu_2019_NTURGBD120}. Based on this protocol, we create a similar one-shot protocol for NTU-60, with  50/10 action classes used for training/testing respectively. To evaluate the effectiveness of the proposed method on viewpoint alignment, we also create two new protocols on NTU-120, for which we group the whole dataset based on (i) horizontal camera views into left, center and right views, (ii) vertical camera views into top, center and bottom views. We conduct two sets of experiments on such disjoint view-wise splits (i) ({\it 100/same 100}): using 100 action classes for training, and testing on the same 100 action classes (ii) ({\it 100/novel 20}): training on 100 action classes but testing on the rest unseen 20 classes. Appendix \ref{supp:protocols} provides more details of training/evaluation protocols (subject splits, \etc) for small-scale datasets as well as the large scale Kinetics-400 dataset.

\vspace{0.1cm}
\noindent{\bf Stereo projections.} For simulating different camera viewpoints, we estimate the fundamental matrix ${\bf F}$ (Eq.~\eqref{eq:f_matrix}),
which
relies on camera parameters. Thus, 
we use the Camera Calibrator from MATLAB to estimate  intrinsic, extrinsic and lens distortion parameters. For a given skeleton dataset, we compute the range of spatial coordinates $x$ and $y$, respectively. We then split them into 3 equally-sized groups to form roughly left, center, right views and other 3 groups for bottom, center, top views. We choose $\sim$15 frame images from each corresponding group, upload them to the Camera Calibrator, and  export  camera parameters.  We then compute the average distance/depth and height per group to estimate the camera position. On NTU-60 and NTU-120, we simply group the whole dataset into 3 cameras, which are left, center and right views, as provided in~\cite{Liu_2019_NTURGBD120}, and then we compute the average distance per camera view based on the height and distance settings given in the table in~\cite{Liu_2019_NTURGBD120}.

\begin{figure}[t]
\centering
\begin{subfigure}[b]{0.49\linewidth}
\includegraphics[trim=0 0 0 0, clip=true,width=0.99\linewidth]{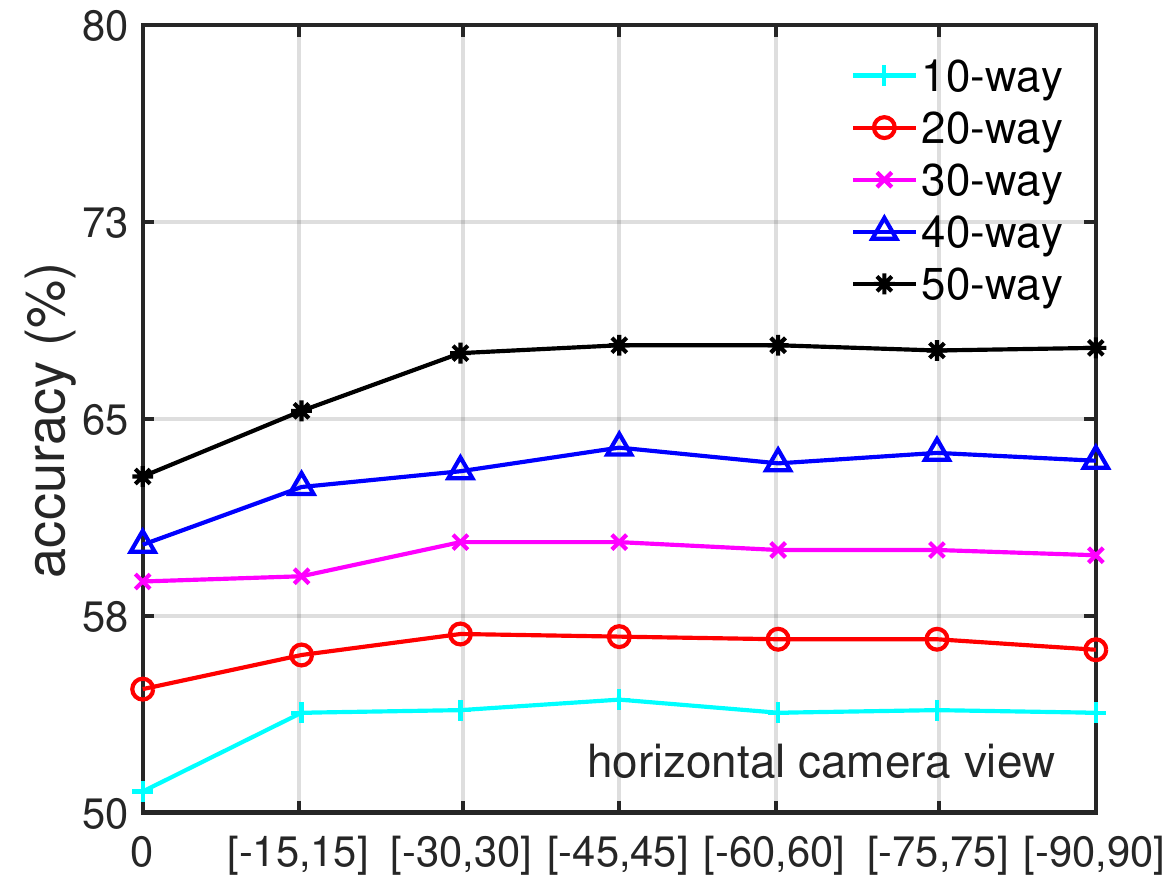}
\caption{\label{fig:sgc_h_angles}}
\end{subfigure}
\begin{subfigure}[b]{0.49\linewidth}
\includegraphics[trim=0 0 0 0, clip=true,width=0.99\linewidth]{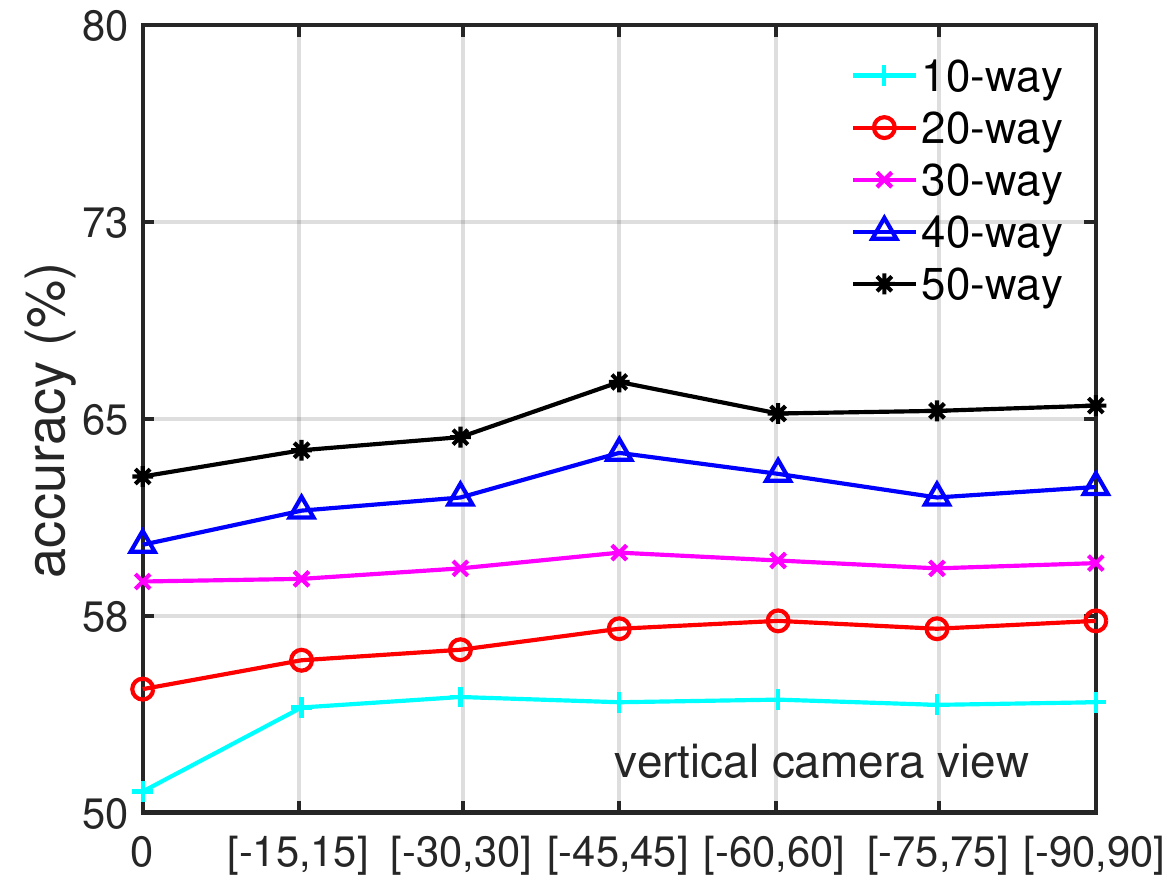}
\caption{\label{fig:sgc_v_angles}}
\end{subfigure}
\caption{The impact of viewing angles in (a) horizontal  and (b) vertical camera views  on NTU-60.
}
\label{fig:h_v_angles}
\end{figure}

\definecolor{LightCyan}{rgb}{0.88,1,1}

\begin{table}[t]
\setlength{\tabcolsep}{0.25em}
\renewcommand{\arraystretch}{0.50}
\caption{Experimental results on NTU-60 (left) and NTU-120 (right) for different camera viewpoint simulations.}
\vspace{-0.6cm}
\begin{center}
\resizebox{\linewidth}{!}{\begin{tabular}{ l c  c c c c  c  c  c c c c}
\toprule
& \multicolumn{5}{c}{NTU-60} & & \multicolumn{5}{c}{NTU-120}\\
\cline{2-6}
\cline{8-12}
\# Training Classes & 10 & 20 & 30 & 40 & 50 & & 20 & 40 & 60 & 80 & 100\\ 
\midrule
Euler simple ($K\!+\!K'$)&  54.3 & 56.2 & 60.4 &  64.0 & 68.1 & &  30.7 & 36.8 & 39.5 &  44.3 & 46.9\\ 
Euler ($K\!\times\!K'$)&  {\bf 60.8} & 67.4 & 67.5 &  70.3 & {\bf 75.0} & &  32.9 & 39.2 & 43.5 & 48.4 & 50.2 \\  
\rowcolor{LightCyan}CamVPC ($K\!\times\!K'$)& 59.7 & {\bf 68.7} & {\bf 68.4} & {\bf 70.4} & 73.2 & & {\bf 33.1} & {\bf 40.8} & {\bf 43.7} & {\bf 48.4} & {\bf 51.4}\\
\bottomrule
\end{tabular}}
\label{ntu60_euler_camvpc}
\end{center}
\vspace{-0.3cm}
\end{table}

\subsection{Ablation Studies}

We start our experiments by investigating various architectural choices and key hyperparameters of our model. 

\begin{sloppypar}

\vspace{0.1cm}
\noindent{\bf Camera viewpoint simulations}. We choose 15 degrees as the step size for the viewpoints simulation. The ranges of camera azimuth and altitude are in [$-90^\circ$, $90^\circ$]. Where stated, we perform a grid search on camera azimuth and altitude with Hyperopt~\cite{bergstra2015hyperopt}. 
Below, we explore the choice of the angle ranges for both horizontal and vertical views.  Fig.~\ref{fig:sgc_h_angles} and~\ref{fig:sgc_v_angles} (evaluations on the NTU-60 dataset) show that the angle range $[-45^\circ, 45^\circ]$ performs the best, and widening the range in both views does not increase the performance any further. 
Table~\ref{ntu60_euler_camvpc} shows results for the chosen range $[-45^\circ,45^\circ]$ of camera viewpoint simulations.  
({\em Euler simple ($K\!+\!K'$)}) denotes a simple concatenation of features from both horizontal and vertical views, whereas ({\em Euler ($K\!\times\!K'$)}) and ({\em CamVPC($K\!\times\!K'$)}) represent the grid search of all possible views.
The table shows that Euler angles for the viewpoint augmentation outperform ({\em Euler simple}), and ({\em CamVPC}) (viewpoints of query sequences are generated by the stereo projection geometry)  outperforms Euler angles in almost all the experiments on NTU-60 and NTU-120. This proves the effectiveness of using the stereo projection geometry for the viewpoint augmentation. 
\end{sloppypar}

\begin{table}[t]
\begin{center}
\setlength{\tabcolsep}{0.25em}
\renewcommand{\arraystretch}{0.50}
\caption{
The impact of the number of frames $M$ in temporal block  under stride step $S$  on results  (NTU-60). $S\!=\!pM$, where $1\!-\!p$ describes the temporal block overlap percentage. Higher $p$ means fewer overlap frames between  temporal blocks.
}
\vspace{-0.35cm}
\setlength{\tabcolsep}{2pt}
\resizebox{\linewidth}{!}{
\renewcommand{\arraystretch}{0.60}
\begin{tabular}{ l c  c  c  c  c  c c  c c  c c c c c}
\toprule
 & \multicolumn{2}{c}{$S = M$} & & \multicolumn{2}{c}{$S = 0.8M$} && \multicolumn{2}{c}{$S = 0.6M$} && \multicolumn{2}{c}{$S = 0.4M$} && \multicolumn{2}{c}{$S = 0.2M$} \\
\cline{2-3}
\cline{5-6}
\cline{8-9}
\cline{11-12}
\cline{14-15}
 $M$ & 50-cls & 20-cls & & 50-cls & 20-cls && 50-cls & 20-cls && 50-cls & 20-cls && 50-cls & 20-cls \\
\midrule
5 & 69.0 & 55.7 && 71.8 & 57.2 && 69.2 & 59.6 && 73.0 & 60.8 && 71.2 & 61.2\\
6 & 69.4 & 54.0 && 65.4 & 54.1 && 67.8 & 58.0 && 72.0 & 57.8 && {\bf 73.0} & {\bf 63.0} \\
8 & 67.0 & 52.7 && 67.0 & 52.5 && \cellcolor{LightCyan}{\bf 73.8}& \cellcolor{LightCyan}{\bf 61.8} && 67.8 & 60.3 && 68.4 & 59.4 \\
10 & 62.2 & 44.5 && 63.6 & 50.9 && 65.2 & 48.4 && 62.4 & 57.0 && 70.4 & 56.7\\
15 & 62.0 & 43.5 && 62.6 & 48.9 && 64.7 & 47.9 && 62.4 & 57.2 && 68.3 & 56.7\\
30 & 55.6 & 42.8 && 57.2 & 44.8 && 59.2 & 43.9 && 58.8 & 55.3 && 60.2 & 53.8\\
45 & 50.0 & 39.8 && 50.5 & 40.6 && 52.3 & 39.9 && 53.0 & 42.1 && 54.0 & 45.2\\
\bottomrule
\end{tabular}}
\label{blockframe_overlap}
\vspace{-0.5cm}
\end{center}
\end{table}

\begin{sloppypar}
\vspace{0.1cm}
\noindent{\bf Block size $M$ and stride size $S$}. Recall from Figure \ref{fig:pipe_basic_A}, that each skeleton sequence is divided into short-term temporal blocks which may also partially overlap.

Table~\ref{blockframe_overlap} shows  evaluations \wrt ~block size $M$ and stride $S$, and indicates %
that the best performance (both 50-class and 20-class settings) is achieved  for smaller block size (frame count in the block) and smaller stride. %
Longer temporal blocks decrease the performance due to the temporal information not reaching the temporal alignment step of JEANIE. Our block encoder encodes each temporal block for learning the local temporal motions, and aggregate these block features finally to form the global temporal motion cues. Smaller stride helps capture more local motion patterns. %
Considering the accuracy-runtime trade-off, we choose $M\!=\!8$ and $S\!=\!0.6M$ for the remaining experiments.
\end{sloppypar}

\vspace{0.1cm}
\noindent{\bf GNN as a block of Encoding Network}. Recall from Section \ref{sec:en} and Appendix \ref{sec:gnn} that our Encoding Network uses a GNN block. For that reason, we investigate several models with the goal of justifying our default choice.

We conduct experiments on 4 GNNs listed in Table~\ref{backboneresults}. S$^2$GC performs the best on large-scale NTU-60 and NTU-120, APPNP outperforms SGC, and SGC outperforms GCN. We also notice that using GNN as a projection layer performs better than single FC layer used in standard transformer by $\sim$5\%. We note that using the RBF-induced distance for $d_{base}(\cdot,\cdot)$ of JEANIE outperforms the Euclidean distance. We choose S$^2$GC 
as a block of our Encoding Network and we use the RBF-induced base distance for JEANIE and other DTW-based models. 

\vspace{0.1cm}
\noindent{\bf $\iota$-max shift}. Recall from Section \ref{sec:jeanie} that the $\iota$-max controls the  smoothness of alignment. 

Table~\ref{ntu60_maxshift} shows the evaluations of $\iota$ for the maximum shift. We notice that $\iota\!=\!2$ yields the best results for all the experimental settings on both NTU-60 and NTU-120. Increasing $\iota$ does not help improve the performance. We think $\iota$ relies on (i) the speeds of action execution (ii) the temporal block size $M$ and the stride $S$. 

\begin{table}[t]
\setlength{\tabcolsep}{0.12em}
\renewcommand{\arraystretch}{0.70}
\begin{center}
\caption{Evaluations of GNN (block of Encoding Network).}
\vspace{-0.3cm}
\resizebox{0.85\linewidth}{!}{
\begin{tabular}{l c c c c c c c}
\toprule
& \multirow{2}{*}{FC layer}& \multirow{2}{*}{GCN}& \multirow{2}{*}{SGC}& \multirow{2}{*}{APPNP}& S$^2$GC & S$^2$GC \\
& & & & & (Eucl.) & (RBF)\\
\midrule
NTU-60 (50-class)& 51.2& 56.0& 68.1& 68.5& 75.6& \cellcolor{LightCyan}{\bf 78.1}\\
NTU-120 (20-class)& 23.3& 27.9& 30.7& 30.8& 34.5& \cellcolor{LightCyan}{\bf 37.2}\\
\bottomrule
\end{tabular}}
\label{backboneresults}
\end{center}
\end{table}

\begin{table}[t]
\setlength{\tabcolsep}{0.25em}
\renewcommand{\arraystretch}{0.50}
\caption{Experimental results on NTU-60 (left) and NTU-120 (right) for $\iota$-max shift.}
\vspace{-0.6cm}
\begin{center}
\resizebox{0.85\linewidth}{!}{\begin{tabular}{ l  c  c  c  c  c  c c  c  c  c  c}
\toprule
& \multicolumn{5}{c}{NTU-60} & & \multicolumn{5}{c}{NTU-120}\\
\cline{2-6}
\cline{8-12}
 & 10 & 20 & 30 & 40 & 50 & & 20 & 40 & 60 & 80 & 100\\ 
\midrule
$\iota\!=\!1$& 60.8  & 70.7 & 72.5 & 72.9  & 75.2 & &  36.3 & 42.5 & 48.7 &  {\bf 50.0} & 54.8\\ 
\rowcolor{LightCyan}$\iota\!=\!2$& {\bf 63.8}  & {\bf 72.9} & {\bf 74.0} & {\bf73.4}  & {\bf 78.1} & & {\bf37.2}  & {\bf43.0} & {\bf 49.2} & {\bf50.0}  & {\bf 55.2}\\  
$\iota\!=\!3$&  55.2 & 58.9 & 65.7 & 67.1  & 72.5 & & 36.7  & {\bf 43.0} & 48.5 & 49.0  & 54.9\\ 
$\iota\!=\!4$&  54.5 & 57.8 & 63.5 &  65.2 & 70.4 & &  36.5 & 42.9 & 48.3 &  48.9 & 54.3\\ 
\bottomrule

\end{tabular}}
\label{ntu60_maxshift}
\end{center}
\end{table}

\subsection{Implementation Details}
\label{sec:det}

Before we discuss our main experimental results, below we provide network configurations and training details.

\vspace{0.1cm}
\noindent{\bf Network configurations.} Given the temporal block size $M$ (the number of frames in a block) and desired output size $d$, the configuration of the 3-layer MLP unit is: FC ($3M \rightarrow 6M$), LayerNorm (LN) as in \cite{dosovitskiy2020image}, ReLU, FC ($6M \rightarrow 9M$), LN, ReLU, Dropout (for smaller datasets, the dropout rate is 0.5; for large-scale datasets, the dropout rate is 0.1), FC ($9M \rightarrow d$), LN. Note that $M$ is the temporal block size
~and $d$ is the output feature dimension per body joint.

\vspace{0.1cm}
\noindent{\bf Transformer block.} The hidden size of our transformer  (the output size of the first FC layer of the MLP in Eq. \eqref{eq:mlp}) depends on the dataset. For smaller scale datasets, the depth of the transformer is $L_\text{tr}\!=\!6$ with $64$ as the hidden size, and the MLP output size is  ${d}\!=\!32$ (note that the MLP which provides $\widehat{\mX}$ and the MLP in the transformer must both have  the same output size). For NTU-60, the depth of the transformer is $L_\text{tr}\!=\!6$, the hidden size is 128 and the MLP output size is  ${d}\!=\!64$. For NTU-120, the depth of the transformer is   $L_\text{tr}\!=\!6$, the hidden size is 256 and the MLP size is ${d}\!=\!128$. For Kinetics-skeleton, the depth for the transformer is $L_\text{tr}\!=\!12$, hidden size is 512 and the MLP output size is ${d}\!=\!256$. The number of heads for the transformer of UWA3D Multiview Activity II, NTU-60, NTU-120 and Kinetics-skeleton is set as 6, 12, 12 and 12, respectively. 
The output size $d'$ of the final FC layer in Eq. \eqref{eq:final_fc_bl} are 50, 100, 200, and 500 for UWA3D Multiview Activity II, NTU-60, NTU-120 and Kinetics-skeleton, respectively.

\vspace{0.1cm}
\noindent{\bf Training details.} The parameters (weights) of the pipeline are initialized with the normal distribution (zero mean and unit standard deviation). We use 1e-3 as the learning rate, and the weight decay is set to 1e-6. We use the SGD optimizer.
We set the number of training episodes to 100K for NTU-60, 200K for NTU-120, 500K for 3D Kinetics-skeleton, and 10K for UWA3D Multiview Activity II. We use Hyperopt \cite{bergstra2015hyperopt} for hyperparameter search on validation sets for all the datasets.

\subsection{Discussion on Supervised Few-shot Action Recognition}


\begin{table}[t]
\setlength{\tabcolsep}{0.12em}
\renewcommand{\arraystretch}{0.70}
\caption{Results on NTU-60 (all use S$^2$GC). 
All methods enjoy temporal alignment by soft-DTW or JEANIE (joint temporal and viewpoint alignment) except where indicated otherwise. 
We use the $\ell_2$ norm for comparing the codes in unsupervised setting with soft-DTW. For unsupervised JEANIE, the distance for comparing the codes is indicated.}
\vspace{-0.6cm}
\begin{center}
\resizebox{\linewidth}{!}{\begin{tabular}{l l c c c c  c  c  c }
\toprule
&  & viewpoint & \multirow{2}{*}{align.} &  \multirow{2}{*}{10} & \multirow{2}{*}{20} & \multirow{2}{*}{30} & \multirow{2}{*}{40} & \multirow{2}{*}{50}\\ 
&  & simulation &   &   &  &  &  & \\ 
\midrule
\multirow{17}{*}{\bf Sup.}
& Matching Nets~\cite{f4Matching}  & &  & 46.1 & 48.6 & 53.3 & 56.2 & 58.8\\
& Matching Nets~\cite{f4Matching}  & & 2V  & 47.2 & 50.7 & 55.4 & 57.7 & 60.2\\
& ProtoNet~\cite{f1} &  & &  47.2 & 51.1 & 54.3 & 58.9 & 63.0\\
& ProtoNet~\cite{f1} &  & 2V  & 49.8 & 53.1 & 56.7 & 60.9 & 64.3 \\
& TAP~\cite{su2022temporal}  & & 
& 54.2 & 57.3 & 61.7 & 64.7 & 68.3 \\
\cdashline{2-9}
& Each frame to frontal view & - & - &  52.9 & 53.3 & 54.6 & 54.2 & 58.3\\
& Each block  to frontal view  & - & - &  53.9 & 56.1 & 60.1 & 63.8 & 68.0\\

& Traj. aligned (video-level)  & - & - &  36.1 & 40.3 & 44.5 & 48.0 & 50.2\\
& Traj. aligned (block-level) & - & - &  52.9 & 55.8 & 59.4 & 63.6 & 66.7\\
& No soft-DTW (S$^2$GC) & - & - &  50.8 & 54.7 & 58.8 & 60.2 & 62.8\\ 
& soft-DTW  & - & T & 53.7 & 56.2 & 60.0 & 63.9 & 67.8 \\ 
& JEANIE & Euler & T+V & 54.0 & 56.0 & 60.2 & 63.8 & 67.8\\
& JEANIE (simple concat.) & Euler & T+2V & 54.3 & 56.2 & 60.4 & 64.0 & 68.1\\ 
& JEANIE & Euler & T+2V &60.8 & 67.4 & 67.5 & 70.3 & 75.0\\ 
& JEANIE & CamVPC & T+2V & 59.7 & 68.7 & 68.4 & 70.4 & 73.2\\
& JEANIE (+crossval.) & CamVPC & T+2V & 63.4 & 72.4 & 73.5 & 73.2 & 78.1\\ 

& \cellcolor{LightCyan}JEANIE (+crossval. +Transf.)  & \cellcolor{LightCyan}CamVPC & \cellcolor{LightCyan}T+2V  & \cellcolor{LightCyan}{\bf 65.0} & \cellcolor{LightCyan}{\bf 75.2} & \cellcolor{LightCyan}{\bf 76.7} & \cellcolor{LightCyan}{\bf 78.9} & \cellcolor{LightCyan}{\bf 80.0}\\
\midrule
\multirow{15}{*}{\parbox{1.0cm}{{\bf Unsup.} +Transf.}}
& soft-DTW (HA)  & - & T & 16.3 & 23.7 & 28.3 & 31.8 & 33.1\\
& soft-DTW (SC)  & - & T & 18.7 & 26.0 & 31.6 & 34.2 & 38.1\\
& soft-DTW (SC$_+$)  & - & T & 18.5 & 25.7 & 30.0 & 33.9 & 37.9\\
& soft-DTW (LLC)  & - & T & 23.1 & 30.1 & 33.0 & 36.4 & 40.9\\
& soft-DTW (SA)  & - & T & 25.4 & 31.7 & 34.6 & 38.0 & 41.7\\
& soft-DTW (LcSA)  & - & T & 25.9 & 32.3 & 35.1 & 38.5 & 42.3\\
& JEANIE (LLC)--$\ell_1$  & CamVPC & T+2V & 27.5 & 33.6 & 36.0 & 41.6 & 44.5\\
& JEANIE (LLC)--$\ell_2$  & CamVPC & T+2V & 27.8 & 33.9 & 36.5 & 41.7 & 44.7\\
& JEANIE (LLC)--HIK  & CamVPC & T+2V & 28.0 & 33.6 & 36.8 & 42.0 & 45.1\\
& JEANIE (LLC)--CSK  & CamVPC & T+2V & 27.8 & 33.9 & 36.8 & 41.7 & 45.0\\
& JEANIE (LcSA)--$\ell_1$  & CamVPC & T+2V & 29.0 & 35.6 & 39.5 & 44.8 & 47.5\\
& JEANIE (LcSA)--$\ell_2$  & CamVPC & T+2V & {\bf 29.1} & {\bf 35.8} & 39.7 & 45.2 & {\bf 48.0}\\
& JEANIE (LcSA)--HIK  & CamVPC & T+2V & 28.8 & {\bf 35.8} & 39.7 & {\bf 45.0} & 47.7\\
& \cellcolor{LightCyan}JEANIE (LcSA)--CSK  & \cellcolor{LightCyan}CamVPC & \cellcolor{LightCyan}T+2V & \cellcolor{LightCyan}29.0 & \cellcolor{LightCyan}{\bf 35.8} & \cellcolor{LightCyan}{\bf 40.0} & \cellcolor{LightCyan}{\bf 45.0} & \cellcolor{LightCyan}{\bf 48.0}\\
& FVM (LcSA)--CSK & CamVPC & T+2V & 27.0 & 33.4 & 36.5 & 42.0 & 45.1 \\
\midrule
\multirow{4}{*}{\parbox{1.0cm}{{\bf Fusion} +Transf.}}
& Weighted fusion & CamVPC & T+2V & 66.5 & 76.9 & 79.0 & 81.2 & 81.5\\
& Finetuning unsup. & CamVPC & T+2V & 67.0 & 77.2  & 79.9 & 82.0 & 84.5\\
& \cellcolor{LightCyan}MAML-inspired fusion & \cellcolor{LightCyan}CamVPC & \cellcolor{LightCyan}T+2V & \cellcolor{LightCyan}{\bf 70.0} & \cellcolor{LightCyan}{\bf 78.3} & \cellcolor{LightCyan}{\bf 81.0} & \cellcolor{LightCyan}{\bf 82.9} & \cellcolor{LightCyan}{\bf 85.0}\\
& Adaptation-based & CamVPC & T+2V & 69.8 & 78.2 & 80.7 & 82.3 & 84.8\\
\bottomrule
\end{tabular}}
\label{ntu60results}
\end{center}
\vspace{-0.3cm}
\end{table}

\begin{table}[t]
\setlength{\tabcolsep}{0.12em}
\renewcommand{\arraystretch}{0.70}
\caption{Experimental results on NTU-120 (S$^2$GC backbone). All methods enjoy temporal
alignment by soft-DTW or JEANIE (joint temporal and viewpoint
alignment) except VA \cite{Zhang_2017_ICCV,8630687} and other cited works. For VA$^*$, we used soft-DTW on temporal blocks while VA generated temporal blocks. For unsupervised soft-DTW and JEANIE, the best distance for comparing the codes is indicated. For brevity, we list unsupervised variants on LcSA but Table \ref{ntu120results_uns} in Appendix \ref{sec:add_ress} contains all variants. }
\vspace{-0.6cm}
\begin{center}
\resizebox{\linewidth}{!}{\begin{tabular}{ l l cc c c  c c c }
\toprule
&  & viewpoint & \multirow{2}{*}{align.} &  \multirow{2}{*}{20} & \multirow{2}{*}{40} & \multirow{2}{*}{60} & \multirow{2}{*}{80} & \multirow{2}{*}{100}\\ 
&  & simulation &   &   &  &  &  & \\ 
\midrule
\multirow{16}{*}{\bf Sup.}
& APSR~\cite{Liu_2019_NTURGBD120} & &   & 29.1 & 34.8& 39.2 & 42.8 & 45.3 \\
& SL-DML~\cite{2021dml}& & &  36.7 & 42.4 & 49.0 & 46.4& 50.9 \\
& Skeleton-DML~\cite{memmesheimer2021skeletondml} & &  & 28.6 & 37.5 & 48.6 & 48.0 & 54.2 \\

& ProtoNet+VA-RNN(aug.)~\cite{Zhang_2017_ICCV} & & & 25.3 & 28.6 & 32.5 & 35.2 & 38.0 \\
& ProtoNet+VA-CNN(aug.)~\cite{8630687}& & & 29.7 & 33.0 & 39.3 & 41.5 & 42.8 \\
& ProtoNet+VA-fusion(aug.)~\cite{8630687} & & & 29.8 & 33.2 & 39.5 & 41.7 & 43.0 \\
& ProtoNet+VA$^*$-fusion(aug.)~\cite{8630687} & & & 33.3 & 38.7 & 45.2 & 46.3 & 49.8 \\
& TAP~\cite{su2022temporal}
& & & 31.2 & 37.7 & 40.9 & 44.5 &  47.3\\
& ALCA-GCN~\cite{zhu2023adaptive} & & & 38.7 & 46.6 & 51.0 & 53.7 & 57.6\\
\cdashline{2-9}
& No soft-DTW (S$^2$GC)& - & - & 30.0 & 35.9 & 39.2 & 43.6 & 46.4 \\ 
& soft-DTW& - & T & 30.3 & 37.2 & 39.7 & 44.0 & 46.8 \\ 
& JEANIE& Euler & T+V& 30.6 & 36.7 & 39.2 & 44.0 & 47.0\\ 
& JEANIE (simple concat.)& Euler & T+2V& 30.7  & 36.8 & 39.5 & 44.3 &  46.9\\ 
& JEANIE & Euler & T+2V& 32.9 & 39.2 & 43.5 & 48.4 & 50.2 \\ 
& JEANIE & CamVPC & T+2V& 33.1 & 40.8 & 43.7 & 48.4 & 51.4 \\
& JEANIE (+crossval.) & CamVPC & T+2V & 37.2 & 43.0 & 49.2 & 50.0 & 55.2 \\
& {\it FVM} (+crossval. +Transf.) & CamVPC & T+2V& 34.5 & 41.9 & 44.2 & 48.7 & 52.0 \\
& \cellcolor{LightCyan}JEANIE (+crossval. +Transf.) & \cellcolor{LightCyan}CamVPC & \cellcolor{LightCyan}T+2V & \cellcolor{LightCyan}{\bf 38.5} & \cellcolor{LightCyan}{\bf 44.1} & \cellcolor{LightCyan}{\bf 50.3} & \cellcolor{LightCyan}{\bf 51.2} & \cellcolor{LightCyan}{\bf 57.0}\\
\midrule
\multirow{3}{*}{\parbox{1.0cm}{{\bf Unsup.} +Transf.}}
& soft-DTW (LcSA)--$\ell_2$  & - & T  & 15.7 & 21.4 & 25.2 & 32.0 & 40.2\\
& \cellcolor{LightCyan}JEANIE (LcSA)--CSK  & \cellcolor{LightCyan}CamVPC & \cellcolor{LightCyan}T+2V  & \cellcolor{LightCyan}{\bf 18.6} & \cellcolor{LightCyan}{\bf 25.2} & \cellcolor{LightCyan}{\bf 32.0} & \cellcolor{LightCyan}{\bf 39.6} & \cellcolor{LightCyan}{\bf 48.5}\\
& FVM (LcSA)--CSK  & CamVPC & T+2V  & 17.5 & 22.4 & 30.7 & 36.1 & 44.5\\
\midrule
\multirow{4}{*}{\parbox{1.0cm}{{\bf Fusion} +Transf.}}
& Weighted fusion & CamVPC & T+2V & 44.4 & 48.6 & 50.8 & 52.0 & 58.3\\
& Finetuning unsup. & CamVPC & T+2V  & 45.6 & 50.8 & 53.0 & 55.0 & 60.2\\
& \cellcolor{LightCyan}MAML-inspired fusion & \cellcolor{LightCyan}CamVPC & \cellcolor{LightCyan}T+2V  & \cellcolor{LightCyan}{\bf 48.2} & \cellcolor{LightCyan}{\bf 53.3} & \cellcolor{LightCyan}{\bf 57.0} & \cellcolor{LightCyan}{\bf 60.3} & \cellcolor{LightCyan}{\bf 62.1}\\
& Adaptation-based & CamVPC & T+2V & 47.9 & 53.0 & 56.5 & 60.0 & 61.9\\

\bottomrule
\end{tabular}}
\label{ntu120results}
\end{center}
\vspace{-0.3cm}
\end{table}

\noindent{\bf NTU-60}. Table~\ref{ntu60results} ({\em Sup.}) shows that using the viewpoint alignment simultaneously in two dimensions,  $x$ and $y$ for Euler angles, or azimuth and altitude the stereo projection geometry ({\em CamVPC}), improves the performance by 5--8\% compared to ({\em Euler}) with a simple concatenation of viewpoints, a variant where the best viewpoint alignment path was chosen from the best alignment path along $x$ and the best alignment path along $y$. Euler with (simple concat.) is  better than  Euler with $y$ rotations only (({\em V}) includes rotations along $y$ while ({\em 2V}) includes rotations along two axes). 
We indicate where temporal alignment  ({\em T}) is also used. When we use  HyperOpt  \cite{bergstra2015hyperopt} to search for the best angle range in which we perform the viewpoint alignment ({\em CamVPC}), the results improve further. Enabling the viewpoint alignment for support sequences ({\em CamVPC}) yields extra improvement, and our best variant of JEANIE boosts the performance by $\sim$2\%.

We also show that aligning query and support trajectories by the angle of torso 3D joint, denoted  ({\em Traj. aligned}) are not very powerful. We note that aligning piece-wise parts (blocks) is better than trying to align entire trajectories. In fact, aligning individual frames by torso to the frontal view ({\em Each frame to frontal view}) and aligning block average of torso direction to the frontal view ({\em Each block  to frontal view}) were marginally better. We note these baselines use soft-DTW.

\begin{sloppypar}
\vspace{0.1cm}
\noindent{\bf NTU-120}.  Table~\ref{ntu120results} ({\em Sup.}) shows that our proposed method achieves the best results on NTU-120, and outperforms the recent SL-DML and Skeleton-DML by 6.1\% and 2.8\% respectively (100 training classes). Note that Skeleton-DML requires the pre-trained model for the weights initialization whereas our proposed model with JEANIE is fully differentiable. 
For comparisons, we extended the view adaptive neural networks~\cite{8630687} by combining them with ProtoNet~\cite{f1}.
VA-RNN+VA-CNN~\cite{8630687} uses 0.47M+24M parameters with random rotation augmentations while JEANIE uses 0.25--0.5M parameters. Their {\em rotation}+{\em translation} keys are not proven to perform smooth optimal alignment as JEANIE. In contrast, $d_\text{JEANIE}$ performs jointly a smooth viewpoint-temporal alignment with smoothness by design. They also use Euler angles which are a worse option (see Table~\ref{ntu60results} and~\ref{ntu120results}) than the camera projection of JEANIE. We notice that ProtoNet+VA backbones is 12\% worse than our JEANIE. Even if we split skeletons into blocks to let soft-DTW perform temporal alignment of prototypes \& query, 
JEANIE is still 4--6\% better. Notice also that JEANIE with transformer is between 3\% and 6\% better than JEANIE with no transformer, which validates the use of transformer on large datasets.
\end{sloppypar}

\begin{table}[t]
\setlength{\tabcolsep}{0.12em}
\renewcommand{\arraystretch}{0.70}
	\centering
	\caption{Experiments on 2D and 3D Kinetics-skeleton. Note that we have no results on JEANIE or FVM  for 2D coordinates as these require very different viewpoint modeling than 3D coordinates.  For brevity, we list unsupervised variants on LcSA but Table \ref{kinetics_results_uns} in Appendix \ref{sec:add_ress} contains more variants.
 }
	\label{kinetics_results}  
	\vspace{-0.3cm}
	\resizebox{0.9\linewidth}{!}{\begin{tabular}{llcccc}  
		\toprule
		&  & viewpoint & \multirow{2}{*}{alignment} & \multirow{2}{*}{2D skel.} & \multirow{2}{*}{3D skel.}\\
		&  & simulation &   & & \\
\midrule
\multirow{6}{*}{\bf Sup.}
& No soft-DTW(S$^2$GC)& -& -& 32.8& 35.9\\
& soft-DTW & - & T & 34.7 & 39.6\\
& FVM & Euler & T+2V & - & 44.1 \\
& JEANIE & Euler & T+2V & - & 50.3\\
& JEANIE(+Transf.) & Euler & T+2V & - & 52.5\\
& \cellcolor{LightCyan}JEANIE(+Transf.) & \cellcolor{LightCyan}CamVPC & \cellcolor{LightCyan}T+2V & \cellcolor{LightCyan}- & \cellcolor{LightCyan}{\bf 52.8}\\
\midrule
\multirow{3}{*}{\parbox{1.0cm}{{\bf Unsup.} +Transf.}}
& soft-DTW(LcSA)--$\ell_2$ & -& T & 19.3 & 22.2\\
& \cellcolor{LightCyan}JEANIE (LcSA)--CSK & \cellcolor{LightCyan}CamVPC & \cellcolor{LightCyan}T+2V & \cellcolor{LightCyan}- & \cellcolor{LightCyan}{\bf 28.3} \\
& FVM (LcSA)--$\ell_2$ & CamVPC & T+2V & - & 25.1\\
\midrule
\multirow{4}{*}{\parbox{1.0cm}{{\bf Fusion} +Transf.}}
& Weighted fusion & CamVPC & T+2V & - & 53.3\\
& Finetuning unsup.& CamVPC & T+2V & - & 54.2\\
& \cellcolor{LightCyan}MAML-inspired fusion & \cellcolor{LightCyan}CamVPC & \cellcolor{LightCyan}T+2V & \cellcolor{LightCyan}- & \cellcolor{LightCyan}{\bf 57.0}\\
& Adaptation-based & CamVPC & T+2V & - & 56.3\\
		\bottomrule
	\end{tabular}}
	\vspace{-0.3cm}
\end{table}

\begin{sloppypar}
\vspace{0.1cm}
\noindent{\bf Kinetics-skeleton}. We evaluate our proposed model on both 2D and 3D Kinetics-skeleton. We follow the training and evaluation protocol in Appendix \ref{supp:protocols}. 
Table~\ref{kinetics_results} shows that using 3D skeletons outperforms the use of 2D skeletons by 3--4\%. The temporal alignment only (with soft-DTW) outperforms baseline (without alignment) by $\sim$2\% and 3\% on 2D and 3D skeletons respectively, and JEANIE outperforms the temporal alignment only by around 5\%. Our best variant with JEANIE further boosts results by  2\%. We notice that the improvements for the use of camera viewpoint simulation ({\em CamVPC}) compared to the use of Euler angles are limited, around 0.3\% and 0.6\% for JEANIE and FVM respectively. The main reason is that the Kinetics-skeleton is a large-scale dataset collected from YouTube videos, and the camera viewpoint simulation becomes unreliable especially when videos are captured by multiple different devices, \eg, camera and mobile phone.
\end{sloppypar}

\begin{table*}[t]
\begin{center}
\caption{Experiments  on the UWA3D Multiview Activity II. All with S$^2$GC layer unless specified.
}
\vspace{-0.3cm}
\resizebox{0.95\linewidth}{!}{\begin{tabular}{ l l c c  c  c  c  c  c c  c  c  c  c c  c  c  }
\toprule

& & \multirow{2}{*}{align.}  & \!\!\!Train\!\!\! & \multicolumn{2}{c}{$V_1$ \& $V_2$} & \multicolumn{2}{c}{$V_1$ \& $V_3$} & \multicolumn{2}{c}{$V_1$ \& $V_4$} & \multicolumn{2}{c}{$V_2$ \& $V_3$} & \multicolumn{2}{c}{$V_2$ \& $V_4$} & \multicolumn{2}{c}{$V_3$ \& $V_4$} & \multirow{2}{*}{Mean}\\
\cline{4-16}
& & & \!\!\!Test & $V_3$ & $V_4$ & $V_2$ & $V_4$ & $V_2$ & $V_3$ & $V_1$ & $V_4$ & $V_1$ & $V_3$ & $V_1$ & $V_2$ & \\
\midrule
\multirow{11}{*}{\bf Sup.}
& GCN & - & &36.4 & 26.2 &20.6 & 30.2 & 33.7 & 22.4 & 43.1 & 26.6 & 16.9 & 12.8 & 26.3 & 36.5 & 27.6 \\
\cdashline{2-17}
& SGC & - & &40.9&60.3&44.1&52.6&48.5&38.7&50.6&52.8&52.8&37.2&57.8&49.6&48.8\\
& \!\!\!+soft-DTW & T & & 43.9&60.8&48.1&54.6&52.6&45.7&54.0&58.2&56.7&40.2&60.2&51.1&52.2\\
& \!\!\!+JEANIE & T+2V& &47.0&62.8&50.4&57.8&53.6&47.0&57.9&62.3&57.0&44.8&61.7&52.3&54.6\\
\cdashline{2-17}
& APPNP&-& & 42.9&61.9&47.8&58.7&53.8&44.0&52.3&60.3&55.1&38.2&58.3&47.9&51.8\\
& \!\!\!+soft-DTW &T& & 44.3&63.2&50.7&62.3&53.9&45.0&56.9&62.8&56.4&39.3&60.1&51.9&53.9\\
& \!\!\!+JEANIE & T+2V & &46.8&64.6&51.3&65.1&54.7&46.4&58.2&65.1&58.8&43.9&60.3&52.5&55.6\\
\cdashline{2-17}
& S$^2$GC&-&&  45.5&64.4&46.8&61.6&49.5&43.2&57.3&61.2&51.0&42.9&57.0&49.2&52.5\\
& \!\!\!+soft-DTW & T & &48.2&67.2&51.2&67.0&53.2&46.8&62.4&66.2&57.8&45.0&62.2&53.0&56.7\\
& \!\!\!+FVM & T+2V & &50.7&68.8&56.3&69.2&55.8&47.1&63.7&68.8&62.5&51.4&63.8&55.7&59.5\\
& \cellcolor{LightCyan}\!\!\!+JEANIE & \cellcolor{LightCyan}T+2V & \cellcolor{LightCyan}&\cellcolor{LightCyan}{\bf 55.3}&\cellcolor{LightCyan}{\bf 70.2}&\cellcolor{LightCyan}{\bf 61.4}&\cellcolor{LightCyan}{\bf 72.5}&\cellcolor{LightCyan}{\bf 60.9}&\cellcolor{LightCyan}{\bf 50.8}&\cellcolor{LightCyan}{\bf 66.4}&\cellcolor{LightCyan}{\bf 73.9}&\cellcolor{LightCyan}{\bf 68.8}&\cellcolor{LightCyan}{\bf 57.2}&\cellcolor{LightCyan}{\bf 66.7}&\cellcolor{LightCyan}{\bf 60.2}&\cellcolor{LightCyan}{\bf 63.7}\\
\hline
\multirow{3}{*}{\bf Unsup.}
& soft-DTW(LcSA)--$\ell_2$& T & & 40.5 & 41.4 & 40.2 & 43.6 & 38.2 & 39.9 & 38.2 & 40.2 & 41.5 & 39.7& 40.9 & 38.8 & 40.3\\
& \cellcolor{LightCyan}JEANIE(LcSA)--CSK& \cellcolor{LightCyan}T+2V &\cellcolor{LightCyan} & \cellcolor{LightCyan}{\bf 53.0} & \cellcolor{LightCyan}{\bf 52.5} & \cellcolor{LightCyan}{\bf 50.1}& \cellcolor{LightCyan}{\bf 51.0}& \cellcolor{LightCyan}{\bf 47.6} & \cellcolor{LightCyan}{\bf 49.2} & \cellcolor{LightCyan}{\bf 49.5} & \cellcolor{LightCyan}{\bf 52.3} & \cellcolor{LightCyan}{\bf 51.3} & \cellcolor{LightCyan}{\bf 49.0}& \cellcolor{LightCyan}{\bf 49.2}& \cellcolor{LightCyan}{\bf 47.1}& \cellcolor{LightCyan}{\bf 50.2}\\
& FVM(LcSA)--CSK& T+2V & & 46.2& 44.0& 45.1& 48.0& 43.5& 44.1& 43.8& 46.0& 47.2& 43.5& 45.8& 43.1 & 45.0\\
\hline
\multirow{4}{*}{\bf Fusion}
& Weighted fusion & T+2V & & 64.9 & 70.4 & 63.9 & 73.4& 62.1& 57.3& 67.8& 74.1& 69.7& 61.3 & 68.9& 63.2&   66.4\\
& Finetuning unsup. & T+2V & & 73.3 & 70.8 & 68.8 & 74.0 & 62.7 & 61.7 & 69.4 & 74.3 & 71.1& 67.9& 72.1& 65.8&  69.3\\
& \cellcolor{LightCyan}MAML-inspired fusion & \cellcolor{LightCyan}T+2V &\cellcolor{LightCyan} & \cellcolor{LightCyan}{\bf 78.7} & \cellcolor{LightCyan}{\bf 73.9} & \cellcolor{LightCyan}{\bf 72.7} & \cellcolor{LightCyan}{\bf 75.9} & \cellcolor{LightCyan}{\bf 65.8} & \cellcolor{LightCyan}{\bf 70.9} & \cellcolor{LightCyan}{\bf 74.3} & \cellcolor{LightCyan}{\bf 76.2} & \cellcolor{LightCyan}{\bf 77.9} & \cellcolor{LightCyan}{\bf 77.3} & \cellcolor{LightCyan}{\bf 80.2} & \cellcolor{LightCyan}{\bf 73.0} &  \cellcolor{LightCyan}{\bf 74.7}\\
& Adaptation-based & T+2V & & 76.3 & 71.5 & 72.0& 75.0 & {\bf 65.8} & 69.2 & 72.8 & 75.5 & 75.9 & 76.5 & 78.3 & 71.7 &  73.4\\
\bottomrule
\end{tabular}}
\label{uwa3dmresults}
\end{center}
\vspace{-0.3cm}
\end{table*}

\begin{table}[t]
\setlength{\tabcolsep}{0.10em}
\renewcommand{\arraystretch}{0.70}
\caption{Results on NTU-120 (multi-view classification). 
We use S$^2$GC.
}
\vspace{-0.6cm}
\begin{center}
\resizebox{\linewidth}{!}{\begin{tabular}{ l l l c  c c c c  c }
\toprule
& \multirow{2}{*}{Eval. Protocol}& Train & bott. & bott. & bott.\& cent. & left & left & left \& cent.\\
\cline{3-9}
& & Test & cent. & top & top & cent. & right & right \\
\midrule
\multirow{6}{*}{\bf Sup.}
& \multirow{3}{*}{100/same 100} & soft-DTW & 74.2 & 73.8 & 75.0 & 58.3 & 57.2 & 68.9\\
& & {\it FVM} & 79.9 & 78.2 & 80.0 & 65.9 & 63.9 & 75.0\\
& &\cellcolor{LightCyan}{\it JEANIE}&  \cellcolor{LightCyan}{\bf 81.5} & \cellcolor{LightCyan}{\bf 79.2} & \cellcolor{LightCyan}{\bf 83.9} & \cellcolor{LightCyan}{\bf 67.7} & \cellcolor{LightCyan}{\bf 66.9} & \cellcolor{LightCyan}{\bf 79.2}\\
\cdashline{2-9}
& \multirow{3}{*}{100/novel 20}&soft-DTW&  58.2 & 58.2 & 61.3 & 51.3 & 47.2 & 53.7 \\
& &{\it FVM}&  66.0 & 65.3 & 68.2 & 58.8 & 53.9 & 60.1 \\
& &\cellcolor{LightCyan}{\it JEANIE}& \cellcolor{LightCyan} {\bf 67.8} & \cellcolor{LightCyan}{\bf 65.8} &\cellcolor{LightCyan} {\bf 70.8} & \cellcolor{LightCyan}{\bf 59.5} & \cellcolor{LightCyan}{\bf 55.0} & \cellcolor{LightCyan}{\bf 62.7} \\
\hline
\multirow{6}{*}{\bf Unsup.}
& \multirow{3}{*}{100/same 100} & soft-DTW & 55.6 & 53.9 & 56.1 & 40.9 & 39.7 & 47.3\\
& & {\it FVM} & 57.8 & 58.0 & 59.7 & 47.9 & 43.1 & 48.8\\
& &\cellcolor{LightCyan}{\it JEANIE}& \cellcolor{LightCyan}{\bf 60.3} & \cellcolor{LightCyan}{\bf 61.7} & \cellcolor{LightCyan}{\bf 63.2} & \cellcolor{LightCyan}{\bf 51.7} & \cellcolor{LightCyan}{\bf 46.9} & \cellcolor{LightCyan}{\bf 52.5}\\
\cdashline{2-9}
& \multirow{3}{*}{100/novel 20}&soft-DTW& 40.2 & 39.7 & 40.8 & 33.7 & 32.9 & 45.5\\
& &{\it FVM}  & 46.2 & 44.5 & 47.0 & 38.1 & 34.0 & 47.1\\
& &\cellcolor{LightCyan}{\it JEANIE}& \cellcolor{LightCyan}{\bf 48.8} & \cellcolor{LightCyan}{\bf 47.2} & \cellcolor{LightCyan}{\bf 50.0} & \cellcolor{LightCyan}{\bf 41.0} & \cellcolor{LightCyan}{\bf 39.7} & \cellcolor{LightCyan}{\bf 51.8}\\
\hline
\multirow{8}{*}{\bf Fusion}
& \multirow{4}{*}{100/same 100} & Weighted fusion & 82.8 & 80.2 & 84.6 & 68.3 & 67.4 & 79.7\\
& & Finetuning unsup. & 83.2 & 81.0 & 86.0 & 69.7 & 68.9 & 80.5\\
& &\cellcolor{LightCyan}MAML-inspired fusion& \cellcolor{LightCyan}{\bf 85.3} & \cellcolor{LightCyan}{\bf 83.2} & \cellcolor{LightCyan}{\bf 87.1} & \cellcolor{LightCyan}{\bf 72.2} & \cellcolor{LightCyan}{\bf 71.7} & \cellcolor{LightCyan}{\bf 82.3}\\
& &Adaptation-based& 85.0 & 82.4 & 86.8 & 71.3 & 69.8 & 81.0\\
\cdashline{2-9}
& \multirow{4}{*}{100/novel 20}&Weighted fusion& 68.7 & 66.3 & 71.2 & 60.4 & 55.9 & 63.3\\
& &Finetuning unsup.& 69.2 & 67.3 & 72.8 & 61.1 & 56.8 & 64.6\\
& &\cellcolor{LightCyan}MAML-inspired fusion& \cellcolor{LightCyan}{\bf 72.3} & \cellcolor{LightCyan}{\bf 69.0} & \cellcolor{LightCyan}{\bf 74.9} & \cellcolor{LightCyan}{\bf 63.0} & \cellcolor{LightCyan}{\bf 58.7} & \cellcolor{LightCyan}{\bf 67.1}\\
& &Adaptation-based& 71.9 & 68.1 & 73.3 & 62.7 & 56.9 & 66.0\\
\bottomrule
\end{tabular}}
\label{ntu120results_view}
\end{center}
\end{table}

\begin{table}[t]
\setlength{\tabcolsep}{0.12em}
\renewcommand{\arraystretch}{0.70}
\caption{Evaluation of different testing strategies, \eg, with supervised learning, unsupervised learning and a combination of both on Kinetics-skeleton when the model is trained with the fusion of both supervised and unsupervised FSAR.
}
\vspace{-0.6cm}
\begin{center}
\resizebox{0.85\linewidth}{!}{\begin{tabular}{ l c c c c c c c}
\toprule
\multirow{2}{*}{Train with fusion} &\multirow{2}{*}{\# ENs} & \multicolumn{6}{c}{Different test cases} \\
\cline{3-8}
& & \multicolumn{2}{c}{sup. only} & \multicolumn{2}{c}{unsup. only} & \multicolumn{2}{c}{sup.+unsup.}\\
\midrule
Weighted fusion & 2 & 52.8 &  & 28.3 &  & 53.3 & \\
Finetuning unsup. & 1 & 53.1 & (\textcolor{red}{$\uparrow$0.6}) & 40.7 & (\textcolor{blue}{$\uparrow$12.4})& 54.2 & (\textcolor{purple}{$\uparrow$0.9})\\
Adaptation-based & 1 & 53.7 & (\textcolor{red}{$\uparrow$1.2}) & 49.6 & (\textcolor{blue}{$\uparrow$21.3}) & 56.3 & (\textcolor{purple}{$\uparrow$3.0})\\
\cellcolor{LightCyan}MAML-inspired fusion & \cellcolor{LightCyan}1 & \cellcolor{LightCyan}54.0 & \cellcolor{LightCyan}(\textcolor{red}{$\uparrow$1.5}) & \cellcolor{LightCyan}50.3 & \cellcolor{LightCyan}(\textcolor{blue}{$\uparrow$22.0}) & \cellcolor{LightCyan}57.0 & \cellcolor{LightCyan}(\textcolor{purple}{$\uparrow$3.7})\\

\bottomrule
\end{tabular}}
\label{tab:kinetics-skeleton2}
\end{center}
\vspace{-0.3cm}
\end{table}

\subsection{Discussion on Unsupervised Few-shot Action Recognition}
\label{sec:unsup-dis}

\begin{sloppypar}
Recall from Section \ref{sec:unsupervised} that JEANIE can help train unsupervised FSAR by forming a dictionary that relies on temporal-viewpoint alignment of JEANIE which factors out nuisance temporal and pose variations in sequences. 

However, the  choice of feature coding and dictionary learning method can  affect the performance of unsupervised learning. Thus, we investigate several variants from Appendix \ref{sec:feat_code}.

Table \ref{ntu60results} ({\em Unsup.}) and Table \ref{ntu120results_uns} in Appendix \ref{sec:add_ress} (extension of Table \ref{ntu120results} ({\em Unsup.})) show on NTU-60 and NTU-120 that the LcSA coder performs better than SA by $\sim$0.6\% and 1.5\%, whereas SA outperforms LLC by $\sim$1.5\% and 2\%. As LcSA and SA are based on the non-linear sigmoid-like reconstruction functions, we suspect they are more robust than linear reconstruction function of LLC. Since the LcSA is the best performer in our experiments followed by SA and LLC or SC, we choose LcSA 
for further analysis. 

Table \ref{ntu60results} ({\em Unsup.}), and Tables \ref{ntu120results_uns} and \ref{kinetics_results_uns} in Appendix \ref{sec:add_ress}  (extensions of Tables \ref{ntu120results} ({\em Unsup.}) and \ref{kinetics_results} ({\em Unsup.})) also show that the choose of different distance measures for comparing the dictionary-coded vectors of sequences during the test stage does not affect the performance by much. The kernel-induced distances, \eg, HIK distance and CSK distance and $\ell_2$-norm outperform the $\ell_1$ norm by $\sim$0.5\% on average. %
We choose the CSK distance for unsupervised JEANIE with LcSA as the default distance for comparing dictionary-coded vectors as it was marginally better performer in the majority of experiments.

Tables \ref{ntu60results} ({\em Unsup.}), \ref{ntu120results} ({\em Unsup.}) and \ref{kinetics_results} ({\em Unsup.}) show that unsupervised JEANIE (temporal-viewpoint alignment) outperforms  soft-DTW (temporal alignment only) by up to 5\%, 9\% and 6\% on NTU-60, NTU-120 and Kinetics-skeleton, respectively. Table \ref{uwa3dmresults} ({\em Unsup.}) shows that the biggest improvement is obtained when using  unsupervised JEANIE on UWA3D Multiview Activity II dataset, with 10\% performance gain. This outlines the importance of the joint temporal-viewpoint alignment under heavy camera pose variations.

Interestingly, FVM in unsupervised learning performs worse compared to our JEANIE, \eg, JEANIE suppresses FVM by $\sim$3\%, 4\% and 3\% respectively on NTU-60, NTU-120 and Kinetics-skeleton in Tables \ref{ntu60results} ({\em Unsup.}), \ref{ntu120results} ({\em Unsup.}) and \ref{kinetics_results} ({\em Unsup.}). On UWA3D Multiview Activity II in Table~\ref{uwa3dmresults} ({\em Unsup.}), JEANIE outperforms FVM by more than 5\%. This is because FVM always seeks the best local viewpoint alignment for every step of soft-DTW which realizes a non-smooth temporal-viewpoint path in contrast to JEANIE. Without the guidance of label information, FVM fails to capture the corresponding relationships between each temporal and viewpoint alignment. Thus, FVM produces a worse dictionary than JEANIE which validates the need for factoring out jointly temporal and viewpoint nuisance variations from sequences.

Table~\ref{ntu120results_view} ({\em Unsup.}) shows that on our newly introduced multi-view classification protocol on NTU-120, for the unsupervised learning experiments, JEANIE outperforms the baseline (temporal alignment only with soft-DTW) by 7\% and 8\% on average on ({\em 100/same 100}) and ({\em 100/novel 20}) respectively . Moreover, JEANIE outperforms the FVM by around 4\% and 3\% on ({\em 100/same 100}) and ({\em 100/novel 20}) respectively.
\end{sloppypar}

\subsection{Discussion on JEANIE and FVM}
\label{dis:jeanie_vs_fvm}

For supervised learning, JEANIE outperforms FVM by 2-4\% on NTU-120, and outperforms FVM by around 6\% on Kinetics-skeleton. For unsupervised learning, JEANIE improves the performance by around 3\% on average on NTU-60, NTU-120 and Kinetics-skeleton. On UWA3D Multiview Activity II, JEANIE suppresses FVM by 4\% and 5\% respectively for supervised and unsupervised experiments. This shows that seeking jointly the best temporal-viewpoint alignment is more valuable than considering viewpoint alignment as a separate local alignment task (free range alignment per each step of soft-DTW). By and large, FVM often performs better than soft-DTW (temporal alignment only) by 3--5\% on average.

\begin{figure}[t]
\centering
\begin{subfigure}[b]{0.49\linewidth}
\includegraphics[trim=1.5cm 2.4cm 1.5cm 3.5cm, clip=true,width=0.99\linewidth]{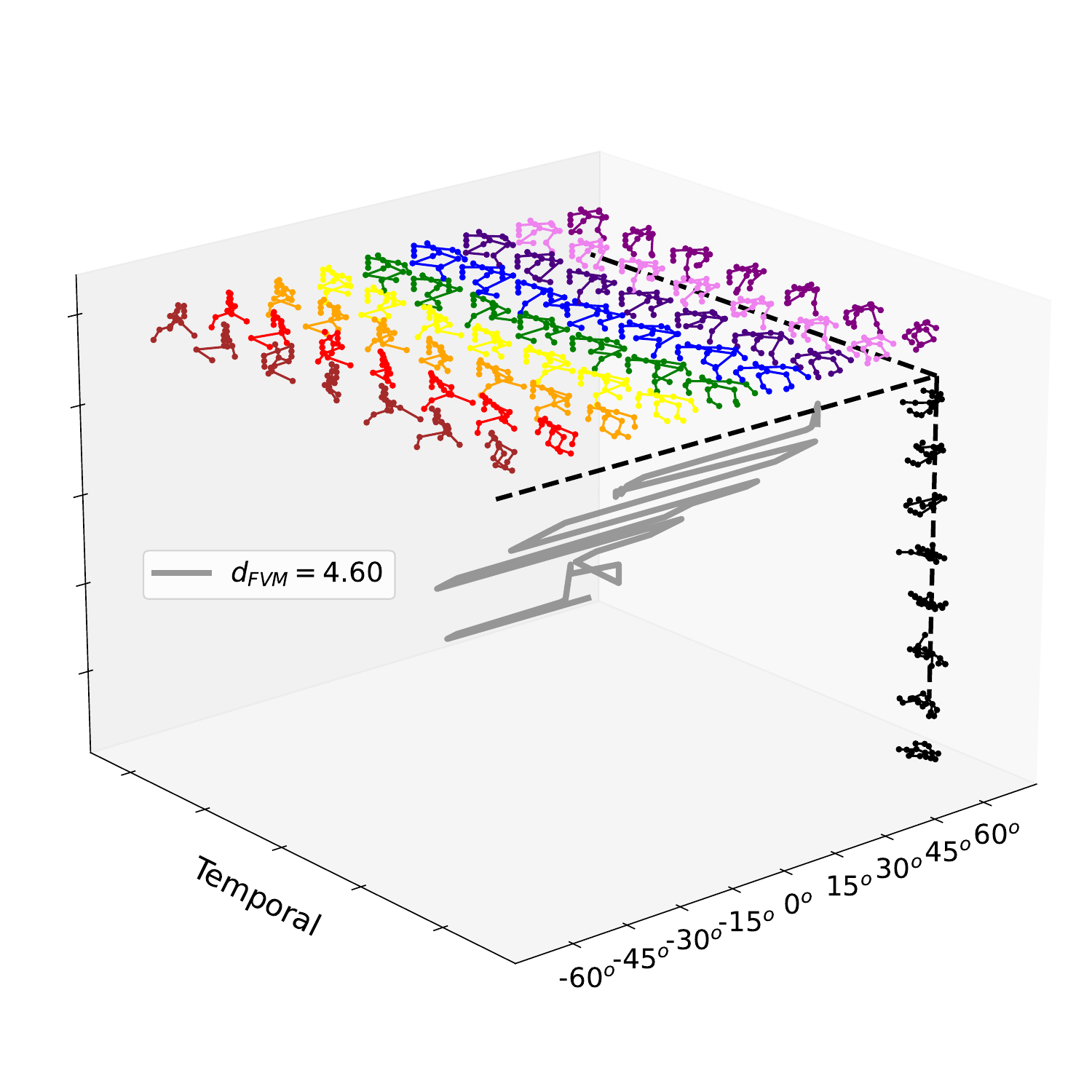}
\caption{{\it walking} \vs~{\it walking} ($d_\text{FVM}\!=\!4.60$)}\label{fig:fvm_walk_walk}
\end{subfigure}
\begin{subfigure}[b]{0.49\linewidth}
\includegraphics[trim=1.5cm 2.4cm 1.5cm 3.5cm, clip=true,width=0.99\linewidth]{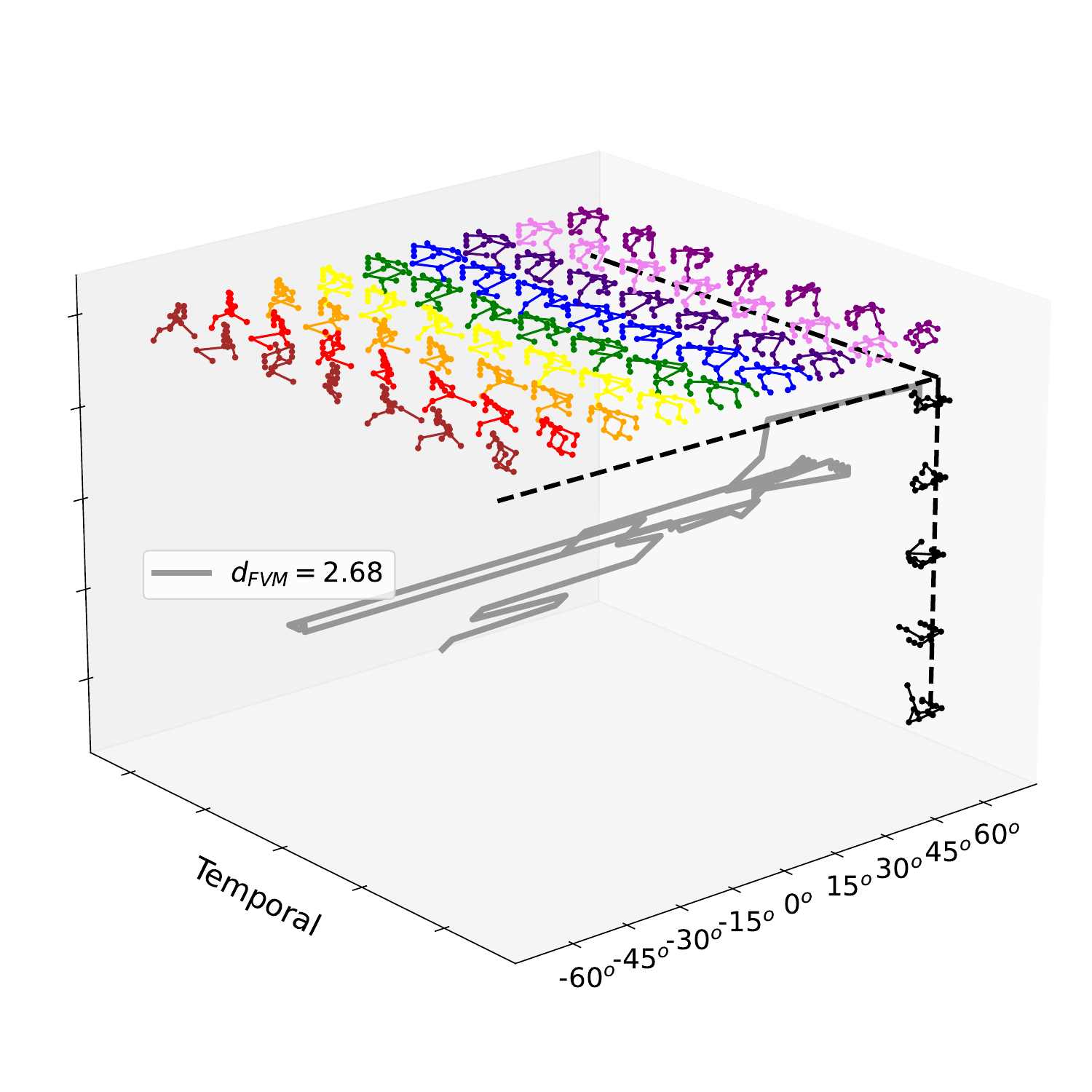}
\caption{{\it walking} \vs~{\it running} ($d_\text{FVM}\!=\!2.68$)}\label{fig:fvm_walk_run}
\end{subfigure}
\begin{subfigure}[b]{0.49\linewidth}
\includegraphics[trim=1.5cm 2.4cm 1.5cm 3.5cm, clip=true,width=0.99\linewidth]{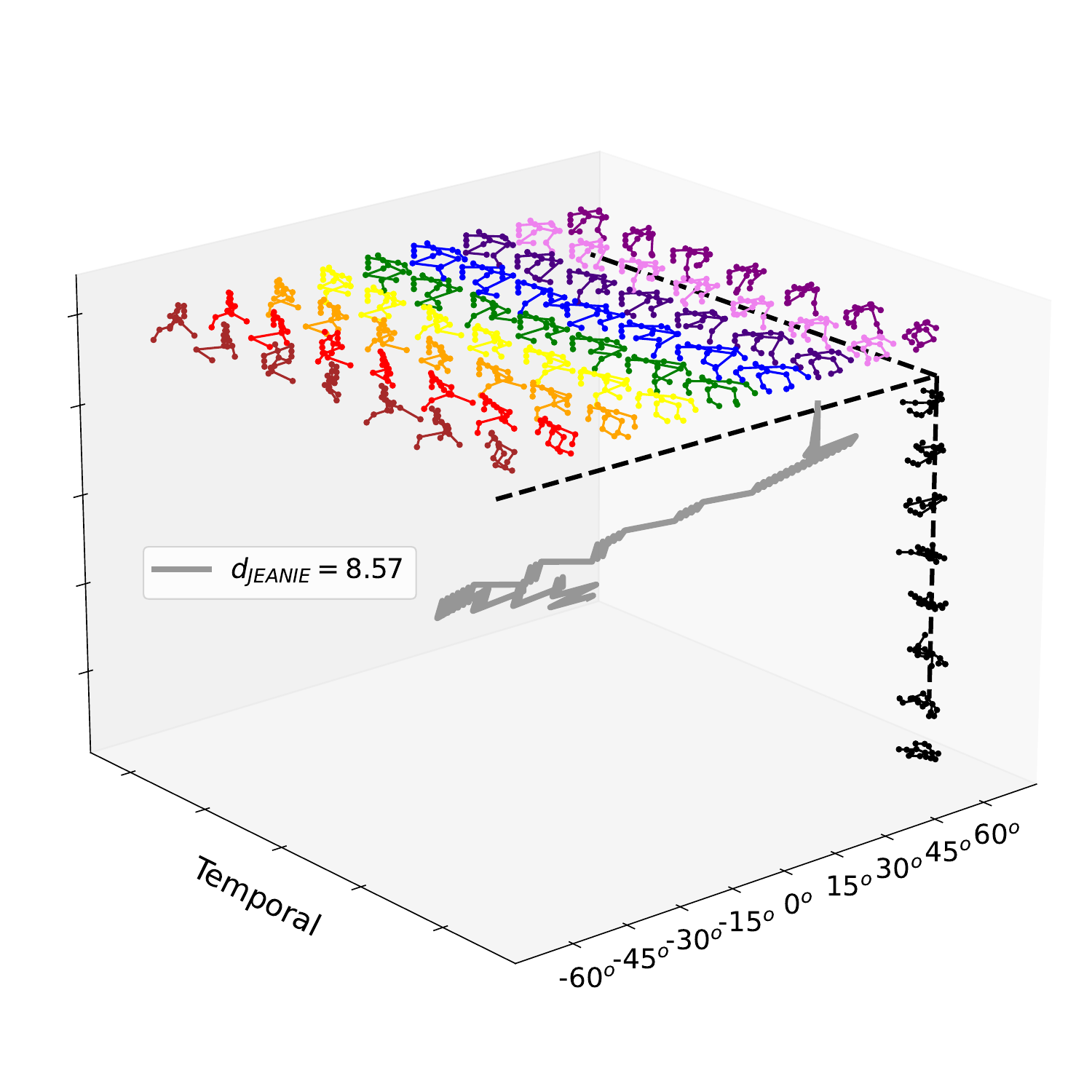}
\caption{{\it walking} \vs~{\it walking} ($d_\text{JEANIE}\!=\!8.57$)}\label{fig:jeanie_walk_walk}
\end{subfigure}
\begin{subfigure}[b]{0.49\linewidth}
\includegraphics[trim=1.5cm 2.4cm 1.5cm 3.5cm, clip=true,width=0.99\linewidth]{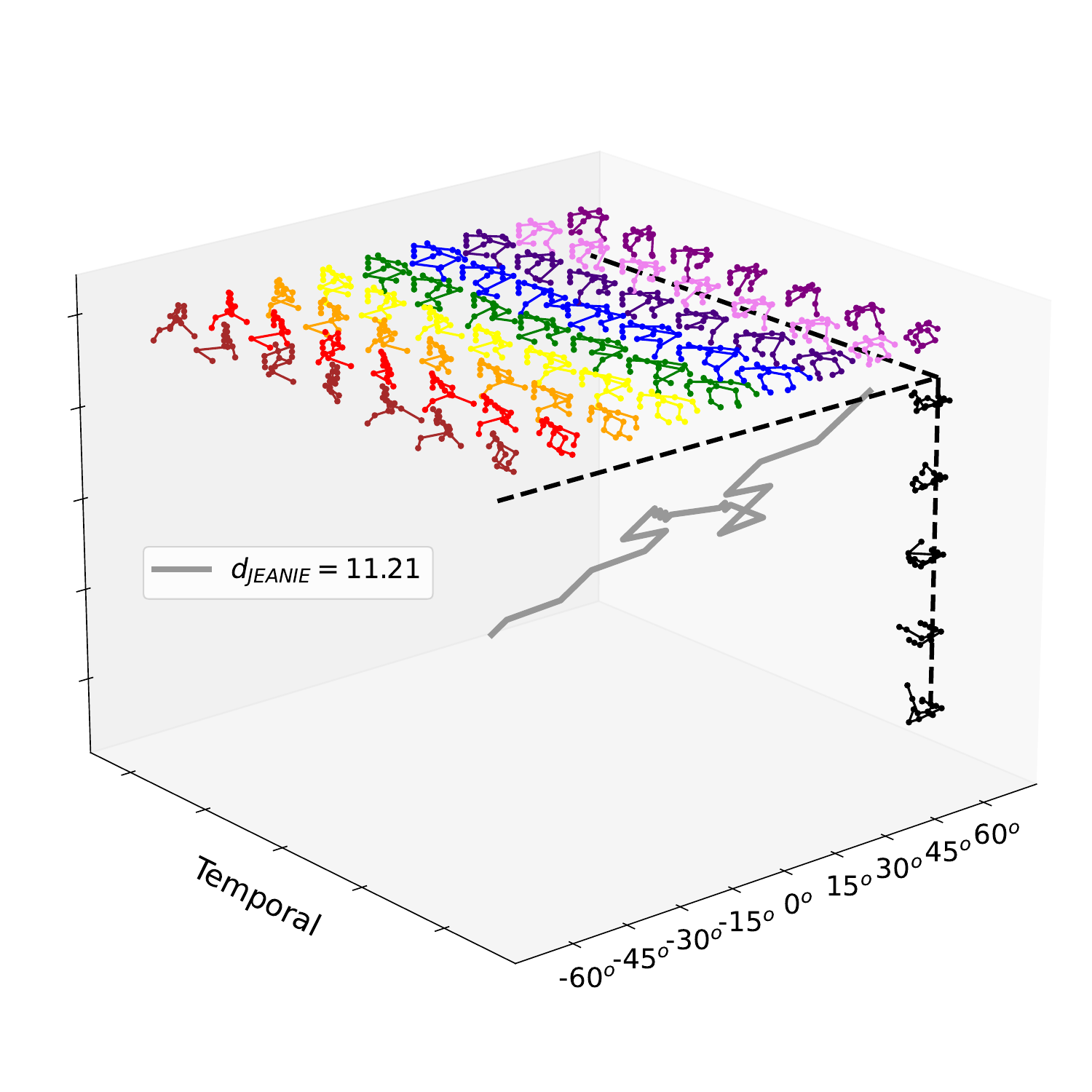}
\caption{{\it walking} \vs~{\it running} ($d_\text{JEANIE}\!=\!11.21$)}\label{fig:jeanie_walk_run}
\end{subfigure}
\caption{Visualization of FVM and JEANIE for {\it walking} \vs~{\it walking} (two different sequences) and {\it walking} \vs~{\it running}. 
We notice that for two different action sequences in (b), the greedy FVM finds the path with a very small distance $d_\text{FVM}\!=\!2.68$ but for sequences of the same action class, FVM gives $d_\text{FVM}\!=\!4.60$. This is clearly suboptimal as the within-class distance is higher then the between-class distance (to counteract this issue, we propose JEANIE). In contrast, our JEANIE is able to produce a smaller distance for within-class sequences and a larger distance for between-class sequences, which is a very important property when comparing pairs of sequences.}
\label{fig:fvm_vs_jeanie_supp}
\end{figure}

\begin{figure}[t]
\centering
\begin{subfigure}[b]{0.49\linewidth}
\includegraphics[trim=1.5cm 2.4cm 1.5cm 3.5cm, clip=true,width=0.99\linewidth]{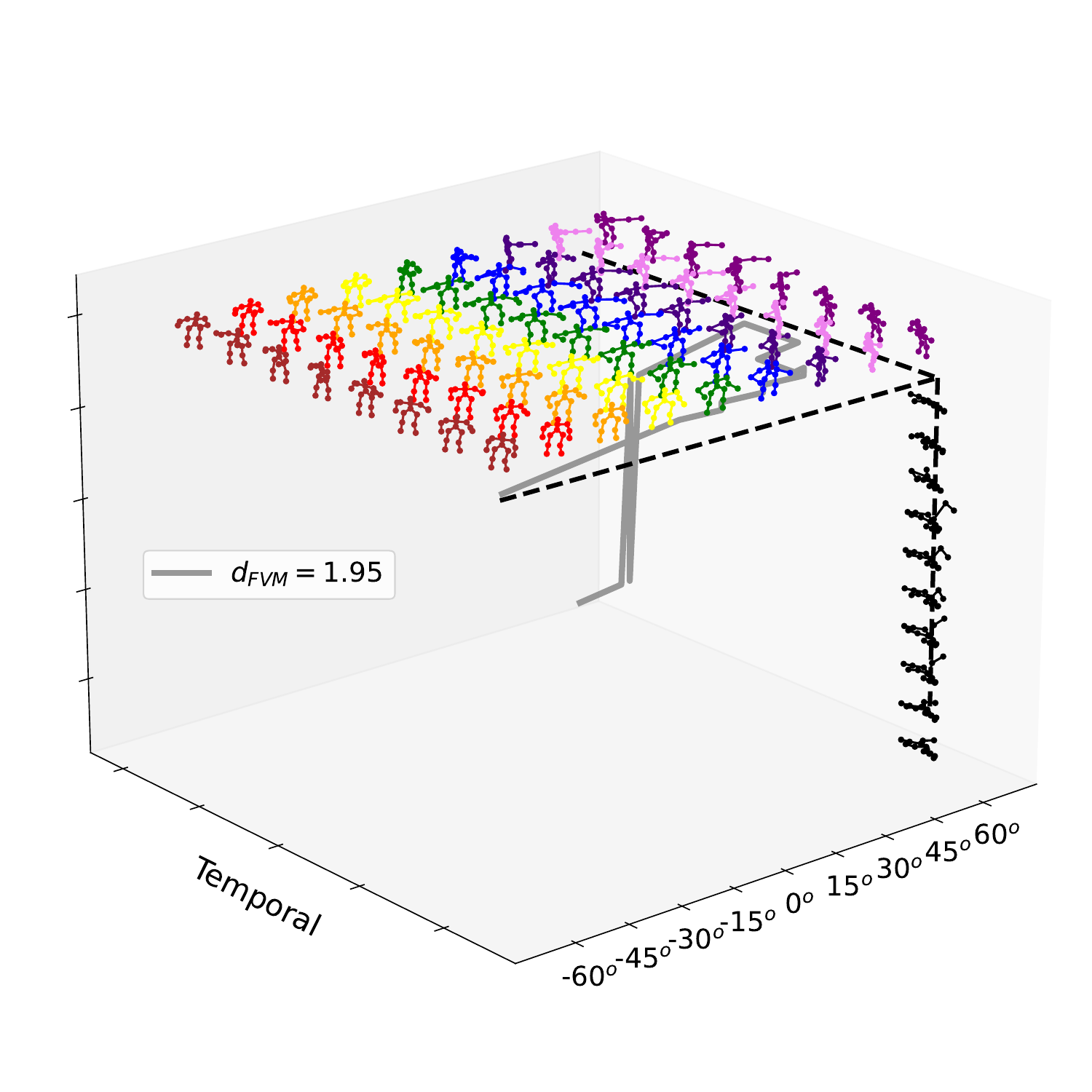}
\caption{{\it two hand punching} \vs~{\it two hand punching} ($d_\text{FVM}\!=\!1.95$)}\label{fig:fvm_punch_punch}
\end{subfigure}
\begin{subfigure}[b]{0.49\linewidth}
\includegraphics[trim=1.5cm 2.4cm 1.5cm 3.5cm, clip=true,width=0.99\linewidth]{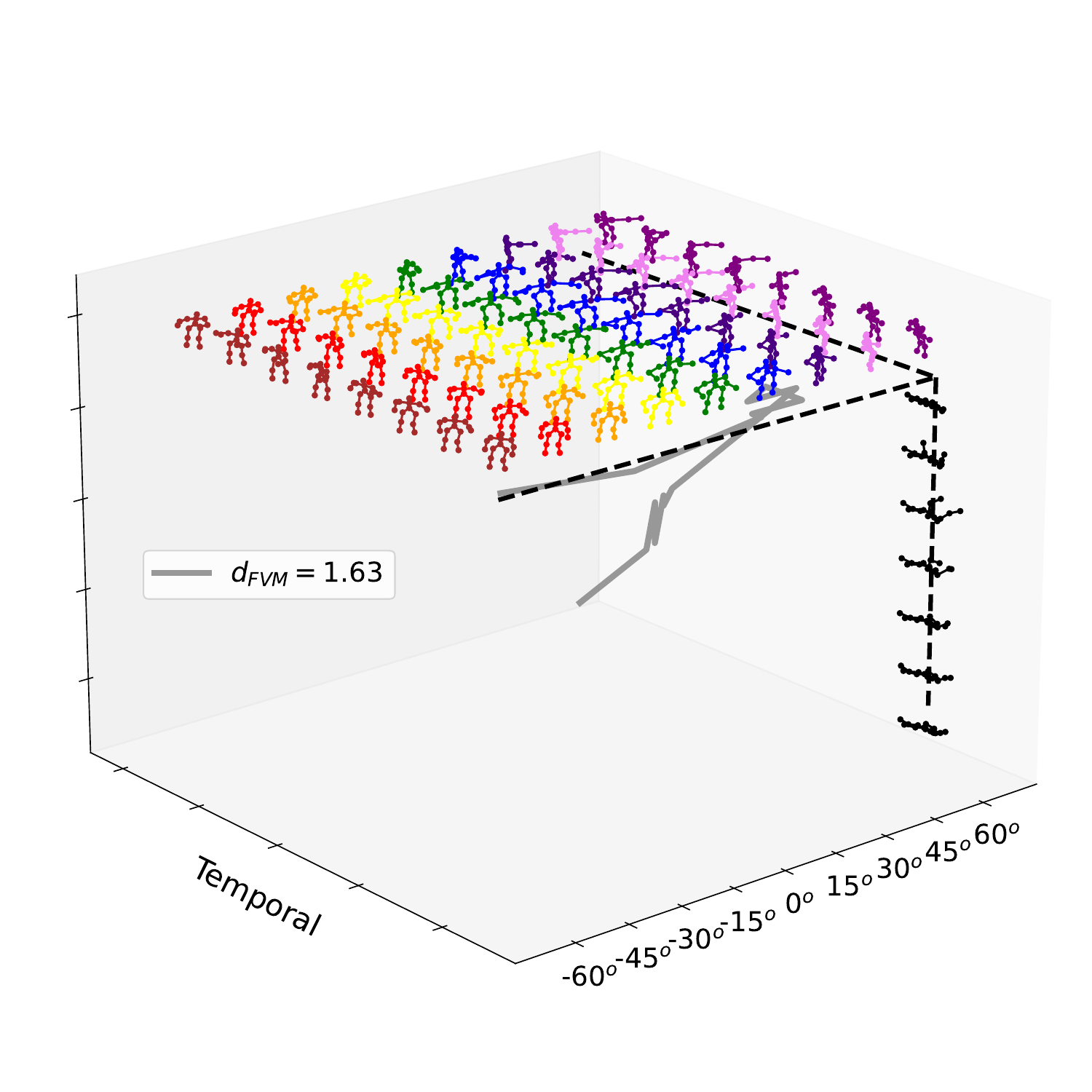}
\caption{{\it two hand punching} \vs~{\it holding head} ($d_\text{FVM}\!=\!1.63$)}\label{fig:fvm_punch_hold}
\end{subfigure}
\begin{subfigure}[b]{0.49\linewidth}
\includegraphics[trim=1.5cm 2.4cm 1.5cm 3.5cm, clip=true,width=0.99\linewidth]{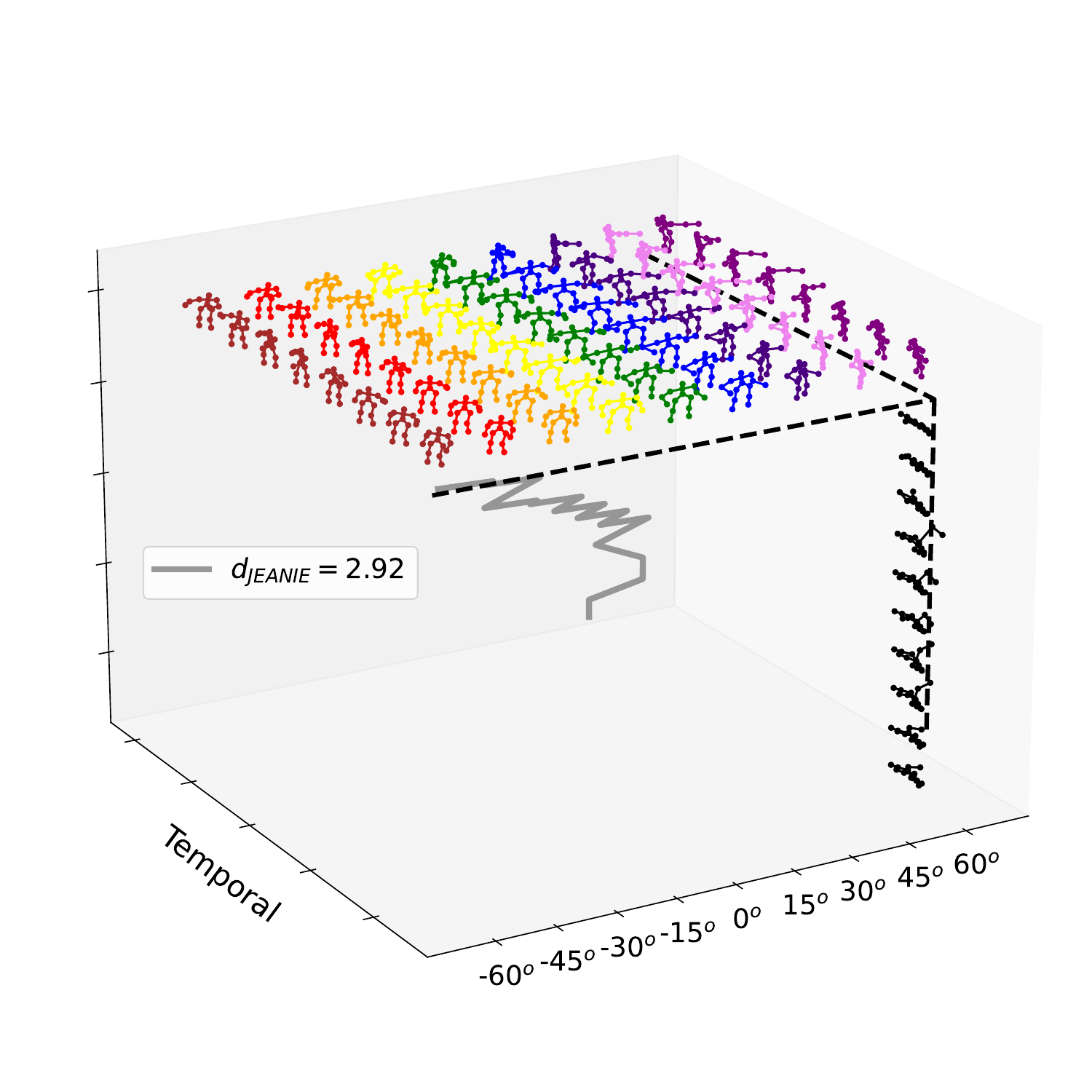}
\caption{{\it two hand punching} \vs~{\it two hand punching} ($d_\text{JEANIE}\!=\!2.92$)}\label{fig:jeanie_punch_punch}
\end{subfigure}
\begin{subfigure}[b]{0.49\linewidth}
\includegraphics[trim=1.5cm 2.4cm 1.5cm 3.5cm, clip=true,width=0.99\linewidth]{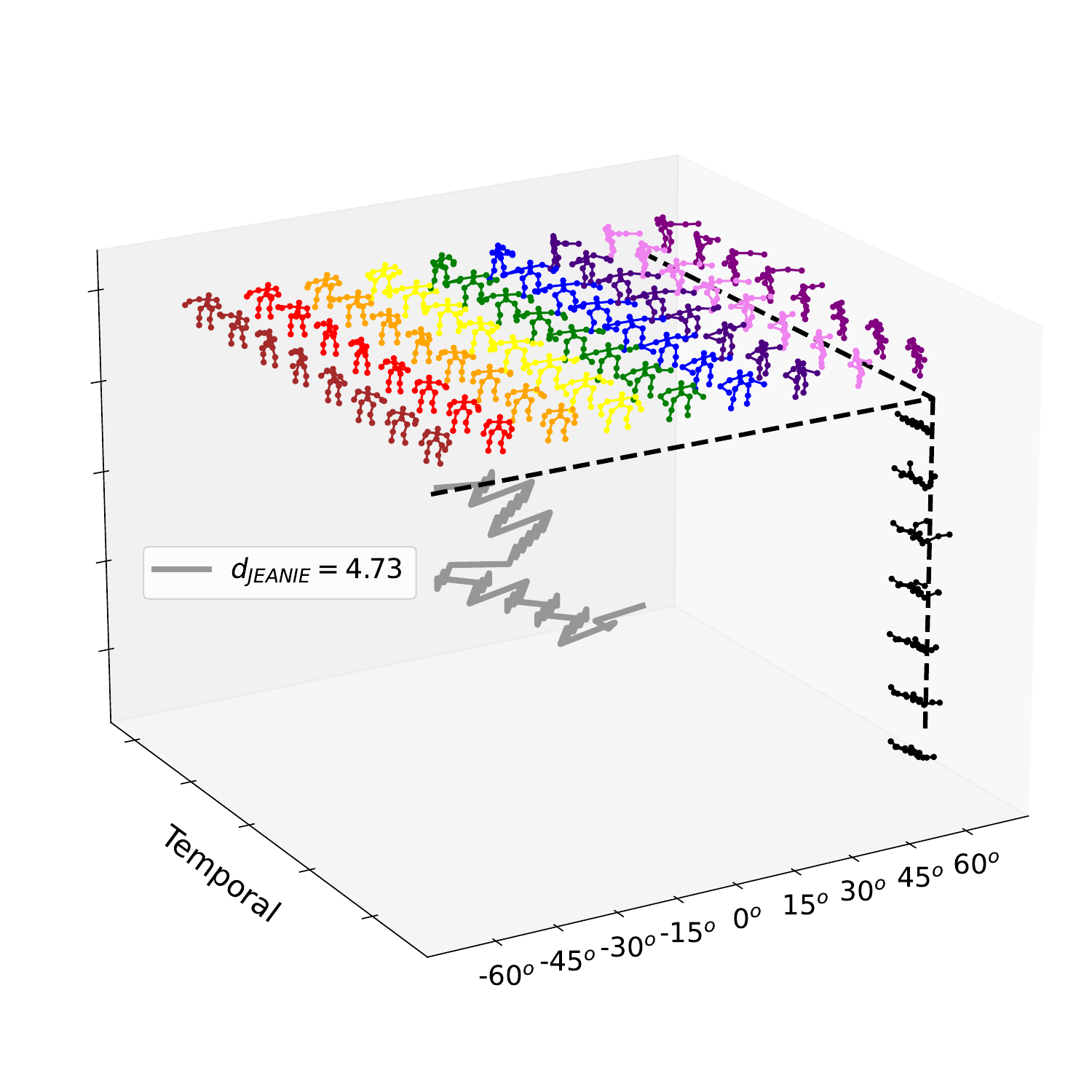}
\caption{{\it two hand punching} \vs~{\it holding head} ($d_\text{JEANIE}\!=\!4.73$)}\label{fig:jeanie_punch_hold}
\end{subfigure}
\caption{Visualization of FVM and JEANIE for {\it two hand punching} \vs~{\it two hand punching} (two different sequences) and {\it two hand punching} \vs~{\it holding head}. 
We notice that for two different action sequences in (b), the greedy FVM finds the path which results in $d_\text{FVM}\!=\!1.63$ for sequences of different action classes, yet FVM gives $d_\text{FVM}\!=\!1.95$ for two sequences of the same class. The within-class distance should be smaller than the between-class distance but greedy approaches such as FVM cannot handle this requirement well. JEANIE gives smaller distance when comparing within-class sequences compared to between-class sequences. This is very important  for comparing  sequences.}
\label{fig:fvm_vs_jeanie_supp3}
\end{figure}

To explain what makes JEANIE perform well on the task of comparing pairs of sequences, we perform some visualisations. To this end, we choose skeleton sequences from UWA3D Multiview Activity II for experiments and visualizations of FVM and JEANIE. UWA3D Multiview Activity II  contains rich viewpoint configurations and so is perfect for our investigations. We verify that our JEANIE is able to find the better matching distances compared to FVM on two following scenarios.

\vspace{0.1cm}
\noindent{\bf Matching similar actions.} We choose a {\it walking} skeleton sequence (`{\tt a12\_s01\_e01\_v01}') as the query sample with more viewing angles for the camera viewpoint simulation, and we select another {\it walking} skeleton sequence of a different view (`{\tt a12\_s01\_e01\_v03}') and a {\it running} skeleton sequence (`{\tt a20\_s01\_e01\_v02}') as support samples respectively.

\begin{sloppypar}
\vspace{0.1cm}
\noindent{\bf Matching actions with similar motion trajectories.} We choose a {\it two hand punching} skeleton sequence (`{\tt a04\_s01\_e01\_v01}') as the query sample with more viewing angles for the camera viewpoint simulation, and we select another {\it two hand punching} skeleton sequence of a different view (`{\tt a04\_s05\_e01\_v02}') and a {\it holding head} skeleton sequence (`{\tt a10\_s05\_e01\_v02}') as support samples respectively.
\end{sloppypar}

\begin{sloppypar}
Figures \ref{fig:fvm_vs_jeanie_supp} and~\ref{fig:fvm_vs_jeanie_supp3} show the visualizations. Comparing Figures \ref{fig:fvm_walk_walk} and \ref{fig:fvm_walk_run} of FVM, we notice that for skeleton sequences from different action classes ({\it walking} \vs~ {\it running}), FVM finds the path with a very small distance $d_\text{FVM}\!=\!2.68$. In contrast, for sequences from the same action class ({\it walking} \vs~{\it walking}), FVM gives $d_\text{FVM}\!=\!4.60$ which is higher than in case of within-class sequences. This is an undesired effect which may result in wrong comparison decision. In contrast, in Figures \ref{fig:jeanie_walk_walk} and \ref{fig:jeanie_walk_run}, our JEANIE gives $d_\text{JEANIE}\!=\!8.57$ for sequences of the same action class and $d_\text{JEANIE}\!=\!11.21$ for sequences from different action classes, which means that the within-class distances are smaller than between-class distances. This is a very important property when comparing pairs of sequences.

Figure \ref{fig:fvm_vs_jeanie_supp3} provides similar observations that  JEANIE produces more reasonable matching distances than FVM.
\end{sloppypar}

\subsection{Discussion on Multi-view Action Recognition}

As mentioned in Section \ref{sec:unsup-dis}, JEANIE yields good results especially in unsupervised learning, with the performance gain over 5\% on UWA3D Multiview Activity II and 4\% on NTU-120 multi-view classification protocols. Below we discuss the multi-view supervised FSAR.

Table~\ref{uwa3dmresults} ({\em Sup.}) shows that adding temporal alignment (with soft-DTW) to SGC, APPNP and S$^2$GC improves results on UWA3D Multiview Activity II, and the big performance gain is obtained via further adding the viewpoint alignment by JEANIE. Despite the dataset is challenging due to novel viewpoints, JEANIE performs consistently well on all different combinations of training/testing viewpoint settings. This is expected as our method aligns both temporal and camera viewpoint which allows a robust classification. JEANIE outperforms  FVM by 4.2\% and the baseline ({\em temporal alignment only with soft-DTW}) by 7\% on average.

Influence of camera views has been explored in~\cite{lei_thesis_2017, lei_tip_2019} on UWA3D Multiview Activity II, and they show that when the left view $V_2$ and right view $V_3$ were used for training and front view $V_1$ for testing, the recognition accuracy is high since the viewing angle of the front view $V_1$ is between $V_2$ and $V_3$; when the left view $V_2$ and top view $V_4$ are used for training and right view $V_3$ is used for testing (or the front view $V_1$ and right view $V_3$ are used for training and top view $V_4$ is used for testing), the recognition accuracies are slightly lower. However, as shown in Table~\ref{uwa3dmresults} ({\em Sup.}), our JEANIE is able to handle the influence of viewpoints and performs almost equally well on all 12 different view combinations which highlights the importance of jointly aligning both temporal and viewpoint modes of sequences.

Table~\ref{ntu120results_view} ({\em Sup.}) shows the experimental results on the NTU-120. We notice that adding more camera viewpoints to the training process helps the multi-view classification, \eg, using bottom and center views for training and top view for testing, and using left and center views for training and the right view for testing, and the performance gain is more than 4\% on ({\em 100/same 100}). Notice that even though we test on 20 novel classes ({\em 100/novel 20}) which are never used in the training set, we still achieve 62.7\% and 70.8\% for multi-view classification in horizontal/vertical camera viewpoints.

\subsection{Fusion of Supervised and Unsupervised FSAR}

\begin{sloppypar}
Recall that Section \ref{sec:fused} defines two baseline and two advanced fusion strategies for 
supervised and unsupervised learning due to their complementary nature.

Tables \ref{ntu60results} ({\em Fusion}), \ref{ntu120results} ({\em Fusion}), \ref{kinetics_results} ({\em Fusion}), and \ref{uwa3dmresults} ({\em Fusion}) show that fusion improves the performance. The MAML-inspired fusion yields 5\%, 5.1\%, 4.2\% and 9\% improvements  compared to the supervised FSAR only on NTU-60, NTU-120, Kinetics-skeleton and UWA3D Multiview Activity II, respectively. This validates our assertion that JEANIE helps design robust feature space for comparing sequences both in supervised and unsupervised scenarios.

The adaptation-based fusion ({\em Adaptation-based}) performs almost as well as the MAML-inspired fusion, within 1\% difference across datasets. This is expected as MAML algorithms are designed to learn across multiple tasks, \ie, in our case the unsupervised reconstruction-driven loss and the supervised loss interact together via gradient updates in such a way that the unsupervised information (a form of clustering) is transferred to guide the supervised loss. The domain adaptation inspired feature alignment achieves a similar effect but the transfer between unsupervised and supervised losses occurs at the feature representation level due to feature alignment.

 Training one EN with the fusion of both supervised and unsupervised FSAR outperforms a naive fusion of scores ({\em Weighted fusion}) from two Encoding Networks trained separately. Finetuning an unsupervised model with supervised loss ({\em Finetuning unsup.}) outperforms the weighted fusion. 

Table \ref{tab:kinetics-skeleton2} compares different testing strategies on fusion models.
The MAML-inspired fusion achieves the best results, 
with 1.5\%, 22.0\% and 3.7\% improvements when tested on supervised learning, unsupervised learning and a fusion of both. 
For both adaptation-based and MAML-inspired fusions, testing on unsupervised FSAR only (nearest neighbor on dictionary-encoded vectors) performs close to the results obtained from supervised FSAR only (nearest neighbor on feature maps), \ie, within 5\% difference. The reduced  performance gap between supervised and unsupervised FSAR suggests that the feature space of EN is adapted to both unsupervised and supervised FSAR.

\end{sloppypar}

\section{Conclusions}
\begin{sloppypar}
We have proposed Joint tEmporal and cAmera viewpoiNt alIgnmEnt (JEANIE) for sequence pairs  and evaluated it on 3D skeleton sequences whose poses/camera views are easy to manipulate in 3D. We have shown that the smooth property of alignment jointly in temporal and viewpoint modes is advantageous compared to the temporal alignment alone (soft-DTW) or models that freely align viewpoint per each temporal block without imposing the smoothness on variations of the matching path.

  JEANIE can match  correctly support and query sequence pairs as it factors out nuisance variations, which is essential under limited
samples of novel classes. Especially, unsupervised FSAR benefits in such a scenario, \ie, when  nuisance variations are factored out, sequences of the same class are more likely to occupy similar/same set of atoms in the dictionary. As supervised FSAR forms the feature space driven by the similarity learning loss and the unsupervised FSAR by the dictionary reconstruction-driven loss, fusing both learning strategies has helped achieve further gains.

Our experiments have shown that using the stereo camera geometry is more efficient than simply generating multiple views by Euler angles. Finally, we have contributed unsupervised, supervised and fused FSAR approaches to the small family of FSAR  for  articulated 3D body joints.
\end{sloppypar}


\vspace{0.9cm}
\noindent{\fontsize{14}{14}\selectfont Appendices}
\vspace{-0.4cm}

\begin{appendix}





\section{Euler Rotations and Simulated Camera Views}
\label{supp:prereq_euler}

\vspace{0.1cm}
\noindent\textbf{Euler angles} \cite{eulera} are defined as successive planar rotation angles around $x$, $y$, and $z$ axes.  For 3D coordinates, we have  the following rotation matrices ${\bf R}_x$, ${\bf R}_y$ and ${\bf R}_z$:
%
%
\begin{align}
\left[\arraycolsep=1.4pt\def\arraystretch{0.5}\begin{array}{ccc}
1 & 0 & 0\\
0 & \text{cos}\theta_x & \text{sin}\theta_x\\
0 & -\text{sin}\theta_x & \text{cos}\theta_x
\end{array} 
\right ],
\left[\arraycolsep=1.4pt\def\arraystretch{0.5}\begin{array}{ccc}
\text{cos}\theta_y & 0 & -\text{sin}\theta_y\\
0 & 1 & 0\\
\text{sin}\theta_y & 0 & \text{cos}\theta_y
\end{array} 
\right],
\left[\arraycolsep=1.4pt\def\arraystretch{0.5} \begin{array}{ccc}
\text{cos}\theta_z & \text{sin}\theta_z &  0\\
-\text{sin}\theta_z & \text{cos}\theta_z & 0\\
0 & 0 & 1
\end{array} 
\right]
\end{align}

\noindent As the resulting composite rotation matrix  depends on the order of  rotation axes, \ie, $\mathbf{R}_x\mathbf{R}_y\mathbf{R}_z\!\neq\!\mathbf{R}_z\mathbf{R}_y\mathbf{R}_x$, we also investigate    the algebra of stereo projection.

\vspace{0.1cm}
\noindent\textbf{Stereo projections} \cite{sterproj}. 
Suppose we have a rotation matrix ${\bf R}$ and a translation vector ${\bf t}\!=\![t_x, t_y, t_z]^T$ between left/right cameras (imagine some non-existent stereo camera). Let ${\bf M}_l$ and ${\bf M}_r$ be  the intrinsic matrices of the left/right cameras.  
Let ${\bf p}_l$ and ${\bf p}_r$ be coordinates of the left/right camera. As the origin of the right camera in the left camera coordinates is ${\bf t}$,  we have: ${\bf p}_r\!=\!{\bf R}({\bf p}_l\!-\!{\bf t})$ and  
$({\bf p}_l\!-\!{\bf t})^T\!=\!({\bf R}^{T}{\bf p}_r)^T$. 
The plane  (polar surface) formed by all points passing through ${\bf t}$ can be expressed by $({\bf p}_l\!-\!{\bf t})^T({\bf p}_l\!\times\!{\bf t})\!=\!0$. Then, ${\bf p}_l\!\times \!{\bf t}\!=\!{\bf S}{\bf p}_l$ where ${\bf S}\!=\! \left[\arraycolsep=1.4pt\def\arraystretch{0.5}\begin{array}{ccc}
0 & -t_z & t_y \\
t_z & 0 & -t_x\\
-t_y & t_x & 0
\end{array} 
\right]$. 

Based on the above equations, we obtain ${{\bf p}_r}^T{\bf R}{\bf S}{\bf p}_l\!=\!0$, and note that ${\bf R}{\bf S}\!=\!{\bf E}$ is the Essential Matrix, and ${{\bf p}^T_r}{\bf E} {\bf p}_l\!=\!0$ describes the relationship for the same physical point under the left and right camera coordinate system. 
As ${\bf E}$ contains no internal information about the camera, and ${\bf E}$ is based on the camera coordinates, we  use a fundamental matrix {\bf F} that describes the relationship for the same physical point under the camera pixel coordinate system. 
The relationship between the pixel  and camera coordinates is: ${\bf p}^*\!=\!{\bf M}{\bf p}'$ and ${{\bf p}'_r}^T{\bf E} {\bf p}'_l\!=\!0$.

Suppose the pixel coordinates of ${\bf p}'_l$ and ${\bf p}'_r$ in the pixel coordinate system are ${\bf p}^*_{l}$ and ${\bf p}^*_{r}$, then we can write ${{\bf p}^*_{r}}^T({\bf M}_r^{-1})^T{\bf E}{\bf M}_l^{-1}{\bf p}^*_{l}\!=\!0$, where  ${\bf F}\!=\!({\bf M}_r^{-1})^T{\bf E}{\bf M}_l^{-1}$ is the fundamental matrix. Thus, the relationship for the same point in the pixel coordinate system of the left/right camera is:
%
\begin{equation}
{{\bf p}^*_{r}}^{T}{\bf F}{\bf p}^*_{l}\!=\!0.
\label{eq:f_matrix}
\end{equation}

\noindent We treat 3D body joint coordinates as ${\bf p}^*_{l}$. Given ${\bf F}$, we obtain their coordinates ${\bf p}^*_{r}$ in the new view.

\section{Graph Neural Network as a Block of Encoding Network}
\label{sec:gnn}

\vspace{0.1cm}
\noindent\textbf{GNN notations.} 
Firstly, let $G\!=\!(\bf{V}, {\bf E})$ be a graph with the vertex set $\bf{V}$  with nodes $\{v_1, \cdots, v_n\}$, and ${\bf E}$ are edges of the graph. Let ${\bf A}$ and ${\bf D}$ be the adjacency and diagonal degree matrix, respectively. Let $\tilde{\bf A}\!=\!{\bf A}\!+\!{\bf I}$ be the adjacency matrix with self-loops (identity matrix) with the corresponding diagonal degree matrix $\tilde{\bf D}$ such that $\tilde{D}_{ii}\!=\!\sum_j ({\bf A}^{ij}\!+\! {\bf I}^{ij})$. 
Let ${\bf S}\!=\!\tilde{\bf D}^{-\frac{1}{2}} \tilde{\bf A}\tilde{\bf D}^{-\frac{1}{2}}$ be the normalized adjacency matrix with added self-loops. For the $l$-th layer, we use ${\bf \Theta}^{(l)}$ to denote the learnt weight matrix, and ${\bf \Phi}$ to denote the outputs from the graph networks. Below, we list backbones used by us.

\vspace{0.1cm}
\noindent\textbf{GCN}~\cite{kipf2017semi}. GCNs learn the feature representations for the features $\mathbf{x}_i$ of each node over multiple layers. For the $l$-th layer, we denote the input  by  ${\bf H}^{(l-1)}$ and the output by ${\bf H}^{(l)}$. Let the input (initial) node representations be ${\bf H}^{(0)}\!=\! \mX$. By $\mX$ we mean some node features for generality of explanation. For our particular case, following the notation in Eq. \eqref{eq:mlp_input}, we would be setting ${\bf H}^{(0)}\!=\! \widehat{\mX}$ for each temporal block.
For an $L$-layer GCN, the output representations are given by:
\begin{equation}
    {\bf \Phi_\text{GCN}}\!=\!{\bf S}{\bf H}^{(L-1)}{\bf \Theta}^{(L)} \text{ where } {\bf H}^{(l)}\!\!=\!\text{ReLU}({\bf S}{\bf H}^{(l-1)}{\bf \Theta}^{(l)}).
\end{equation}

\vspace{0.1cm}
\noindent\textbf{APPNP}~\cite{johannes2019iclr}.  Personalized Propagation of Neural Predictions (PPNP) and its fast approximation, APPNP, are based on the  personalized PageRank. Let  ${\bf H}^{(0)}\!=\!f_{\bf \Theta}(\mX)$ be the input to APPNP, where $f_{\bf \Theta}(\cdot)$ can be an MLP with parameters  ${\bf \Theta}$. Let the output of the $l$-th layer be ${\bf H}^{(l)}\!=\! (1-\alpha){\bf S}{\bf H}^{(l-1)}\!+\!\alpha{\bf H}^{(0)}$, where $\alpha$ is the teleport (or restart) probability  in range $(0, 1]$. For an $L$-layer APPNP, we have:
\begin{equation}
    {\bf \Phi_\text{APPNP}}\!=\!{(1\!-\!\alpha)\bf S}{\bf H}^L\!+\!\alpha{\bf H^{(0)}}.
\end{equation}

\vspace{0.1cm}
\noindent\textbf{SGC}~\cite{felix2019icml} { \&} {{\bf S$^2$GC}}~\cite{hao2021iclr}. 
SGC captures the  $L$-hops neighborhood in the graph by the $L$-th power of the transition matrix used as a spectral filter. For an $L$-layer SGC,  
we obtain:
\begin{equation}
    {\bf \Phi_\text{SGC}}\!=\!{\bf S}^L{\bf X}{\bf \Theta}.
\end{equation}

Based on a modified Markov Diffusion Kernel, Simple Spectral Graph Convolution (S$^2$GC) is the summation over $l$-hops, $l\!=\!1,\cdots,L$.  The output of S$^2$GC is: 
\begin{equation}
    {\bf \Phi_\text{S$^2$GC}} \!=\!  \frac{1}{L}\sum_{l=1}^{L}((1\!-\!\alpha){\bf S}^l{\bf X}\!+\!\alpha{\bf X}){\bf \Theta}.
\end{equation}

In case of APPNP, SGC and S$^2$GC, $|\mathcal{F}_{GNN}|\!=\!0$ because we do not use their learnable parameters $\bf \Theta$ (\ie, think $\bf \Theta$ is set as the identity matrix). The GNN outputs are further passes into a Transformer and an FC layer, which returns $\mPsi\!\in\!\mbr{d'\times K\times K'\times\tau}$ query feature maps and $\mPsi'\!\in\!\mbr{d'\times\tau'}\!$ support feature maps.

\section{Feature Coding and Dictionary Learning}
\label{sec:feat_code}

The core idea of feature coding is to reconstruct a feature vector with codewords by solving a least squares based optimization problem with constraints imposed on the codewords. The full codewords (\aka elements or atoms) compose a dictionary. Atoms in the dictionary are not required to be orthogonal and the dictionary may be an over-complete (the number of atoms is larger than their dimension). For most feature coding algorithms, only a subset of codewords are chosen by the solver to represent a feature vector, and thus the coding vector $\valpha$ may be sparse, \ie, the responses are zeros on those codewords which are not chosen.
In what follows, we however replace the Euclidean distance with the JEANIE measure.

\begin{sloppypar}
    The main difference among various feature coding methods lies in the constraint term.
Alternatively, we obtain $\valpha$ by defining some  specific function $\valpha(\mPsi_i;\mM)$ that implicitly realizes the regularization term. The choice of $\mOmega(\valpha_i, \mM, \mPsi_i)$ realizes some desired constraints via regularization $\kappa > 0$, \eg, $\mOmega(\valpha_i, \mM, \mPsi_i)\!=\!\|\valpha_i\|_1$ encourages sparsity of $\valpha$. %
\end{sloppypar}

\subsection{Feature Coding}
\label{sec:feat_coding}

Below we detail different feature coders we explore in our work, \ie, Hard Assignment (HA)~\cite{Csurka04visualcategorization}, Sparse Coding (SC) \cite{10.5555/2976456.2976557, 5206757}, Non-negative Sparse Coding (SC$_+$)~\cite{1030067}, Locality-constrained Linear Coding (LLC)~\cite{5540018}, Soft Assignment (SA)~\cite{bilmes1998gentle, 10.1007/978-3-540-88690-7_52}, and Locality-constrained Soft Assignment (LcSA)~\cite{6116129,6126534}. LcSA is our default feature coder due to its simplicity and strong performance.

\vspace{0.1cm}
\noindent{\bf Hard Assignment (HA)}. This encoder assigns each $\mPsi$ to its nearest $\vm$ by solving the following optimisation problem:
\begin{align}
    & \valpha(\mPsi)=\argmin_{\alpha' \in \{0, 1\}^{k}} d^2_{\text{JEANIE}}(\mPsi,\mM \valpha'), \\
    & \quad\quad\quad\quad\; \text{s.t.} \quad \|\valpha'\|_1 = 1. \nonumber
\end{align}

\vspace{0.1cm}
\noindent{\bf Sparse Coding (SC)} \& {\bf Non-negative Sparse Coding (SC$_+$)}. SC encodes each $\mPsi$ as a sparse linear combination of atoms $\mM$ by optimising the following objective:
\begin{align}
    & \valpha(\mPsi)=\argmin_{\alpha'} d^2_{\text{JEANIE}}(\mPsi,\mM \valpha') + \kappa\|\valpha'\|_1,
\end{align}
whereas SC$_+$ additionally imposes a constraint that $\valpha' \geq 0$. Both SC and SC$_+$ encode each $\mPsi$ on a subspace of $\mM$ of size controlled by the sparsity term.

\vspace{0.1cm}
\noindent{\bf Locality-constrained Linear Coding (LLC)}. The LLC encoder uses the following criteria for each $\mPsi$:
\begin{align}
    & \valpha(\mPsi)=\argmin_{\alpha'} d^2_{\text{JEANIE}}(\mPsi,\mM \valpha') + \kappa \| \mathbf{d} \odot \valpha\|_2^2, \\
    & \quad\quad\quad\quad\; \text{s.t.} \quad \mathbf{1}^{\text{T}} \valpha' = 1, \nonumber
\end{align}
where $\odot$ denotes the element-wise multiplication and $\mathbf{d}\!\in\!\mbr{k}$ is the non-locality penalty that penalises selection of dictionary atoms that are far from $\mPsi$. Specifically,
\begin{equation}
    \mathbf{d}=\left[\text{exp}^{\frac{d^2_{\text{JEANIE}}(\mPsi, \vm_1)}{\sigma}},\cdots,\text{exp}^{\frac{d^2_{\text{JEANIE}}(\mPsi, \vm_k)}{\sigma}}\right]^T,
\end{equation}
where $\sigma\geq 0$ adjusts the weight decay speed for the non-locality penalty. We further normalize $\mathbf{d}$ to be between 0 and 1. The constraint $\mathbf{1}^{\text{T}} \valpha'\!=\!1$ follows the shift-invariant requirements of the LLC encoder.

\vspace{0.1cm}
\noindent{\bf Soft Assignment (SA)} \& {\bf Locality-constrained Soft Assignment (LcSA)}.
SA expresses each $\mPsi$ as the membership probability of $\mPsi$ belonging to each $\vm$ in $\mM$, a concept known from the MLE of Gaussian Mixture Models (GMM). SA is derived under equal mixing probability and shared variance $\sigma$ of GMM components. SA is a closed-form term:
\begin{align}
    & \valpha'(\mPsi; \mM,\!\sigma)\!=\!\frac{1}{Z(\mPsi;\mM,\! \sigma)}\!\left[\text{e}^{\!-\frac{d^2_{\text{JEANIE}}(\mPsi,\vm_1)}{2\sigma^2}}
    \!, \cdots, \text{e}^{\!-\frac{d^2_{\text{JEANIE}}(\mPsi,\vm_k\|)}{2\sigma^2}}\!\right]\!^{\text{T}}, \nonumber \\
    & \quad\quad\; \text{where}\quad Z(\mPsi;\mM,\! \sigma)\!=\!\sum_{k'=1,\cdots,k}\text{e}^{-\frac{1}{2\sigma^2}d^2_{\text{JEANIE}}(\mPsi, \vm_{k'})}. \label{eq:sa}
\end{align}
The above model usually yields largest values of $\alpha'_i$ for anchor $\vm_i$ in $\mM$ that is a close JEANIE neighbor of $\mPsi$. However, even for $\vm_i$ that is far from $\mPsi$, $\alpha'_i > 0$. For this reason, SA is only approximately locality-constrained. 

LcSA admits the locality-constrained membership probability of the form:
\begin{equation}
    \valpha(\mPsi) = \pi(\valpha'(\mPsi;\mM_{\text{NN}(\mPsi;k')})), \label{eq:lcsa}
\end{equation}
where $\mM_{\text{NN}(\mPsi;k')}$ returns $k'$ nearest neighbors of $\mPsi$ in $\mM$ based on the JEANIE measure, whereas $\pi(\cdot)$ projects back coefficients of $\valpha'$ into $\valpha$ at positions following original indexes of nearest neighbors in dictionary $\mM$. Remaining locations in $\valpha$ are zeroed. LcSA forms subspaces of size $k'$.

\subsection{Dictionary Learning}
\label{sec:dic}

For all the above listed feature coding methods, we employ a simple dictionary learning objective which follows Eq. \eqref{eq:unsup}. We assume some evaluated/fixed dictionary-coded vectors as a coding matrix $\malpha\!\equiv\![\valpha_1, \cdots, \valpha_{N'}]$ ($N'$ is the number of samples per mini-batch), and we compute:
\begin{equation}
    \mM = \argmin_{\mM'}\sum_{i=1}^{N'}d^2_{\text{JEANIE}}(\mPsi_i, \mM'\valpha_i).
\end{equation}
Notice that for fixed $\malpha$ and fixed feature matrices $\mPsi$, the regularization term becomes a constant.
%
%
For the dictionary learning step, we detach $\mPsi$ and $\valpha$, and run 10 iterations of gradient descent per mini-batch \wrt ~$\mM$. %

\section{Fusion by Alignment}
\label{sec:fus_al}
\begin{sloppypar}

\begin{algorithm}[tbp!]
\caption{Fusion of Supervised and Unsupervised FSAR by Feature Maps Alignment (one training iteration).}
\label{code:fusion}
{\bf Input}: $ \Gamma\equiv\{\tX_{b}\}_{b\in\idx{B}}\cup\{\tX'_{b,n,z}\}_{\substack{b\in\idx{B}\\n\in\idx{N}\\z\in\idx{Z}}}$: query/support blocks in  batch; $\mathcal{F}$: EN parameters; $\mM$ and $\malpha$; \texttt{alpha\_iter} and \texttt{dic\_iter}: numbers of iterations for updating $\malpha$ and $\mM$; $\omega, \omega_\text{DL}$ and $\omega_\text{EN}$: the learning rate for $\malpha, \mM$ and $\mathcal{F}$ respectively; $B$: size of the mini-batch; $\lambda$: regularization parameter. 
\begin{algorithmic}[1]
\State{$\Upsilon\equiv\{\mPsi_{b}\}_{b\in\idx{B}}\cup\{\mPsi'_{b,n,z}\}_{\substack{b\in\idx{B}\\n\in\idx{N}\\z\in\idx{Z}}}\!$ where
$\begin{cases}
\mPsi_{b}\!=\!f^*(\tX_b;\mathcal{F})&\\
\mPsi'_{b,n,z}\!=\!f^*(\tX'_{b,n,z};\mathcal{F})\!\!\!\!\!\!\!\!\!&
\end{cases}$
\gray{(obtain feature maps for global parameters $\mathcal{F}$)}
}
\State{$\widehat{\mathcal{F}}:=\mathcal{F}$ \gray{$\quad\qquad\qquad\qquad\qquad\qquad\qquad$(copy parameters of EN)} }
\State{$(\widehat{\mathcal{F}},\mM)=\text{Algorithm\ref{code:unsup}}(\Upsilon, \widehat{\mathcal{F}},\mM, \malpha,$ \gray{$\quad\;\,$(unsupervised FSAR)}}
\Statex{$\qquad\qquad\qquad\texttt{alpha\_iter}, \texttt{dic\_iter}, \omega, \omega_\text{DL}, \omega_\text{EN})$}
\State{$\widehat{\Upsilon}\equiv\{\widehat{\mPsi}_{b}\}_{b\in\idx{B}}\cup\{\widehat{\mPsi}'_{b,n,z}\}_{\substack{b\in\idx{B}\\n\in\idx{N}\\z\in\idx{Z}}}\!$ where
$\begin{cases}
\widehat{\mPsi}_{b}\!=\!f^*(\tX_b;\widehat{\mathcal{F}})&\\
\widehat{\mPsi}'_{b,n,z}\!=\!f^*(\tX'_{b,n,z};\widehat{\mathcal{F}})\!\!\!\!\!\!\!\!\!&
\end{cases}$
\gray{(obtain feature maps for parameters $\widehat{\mathcal{F}}$ from the unsupervised step)}
}
%
%

\State{$\mathcal{L}_\text{align}=\sum_{i=1}^{N'}d^2_{\text{JEANIE}}(\mPsi_i, \widehat{\mPsi}_i)$ \gray{(alignment of sup. \& unsup. maps)}$\!\!$} 
\Statex{$\qquad\qquad\qquad\qquad\qquad\qquad$where $\;N'\!=\!|\Upsilon|$, $\mPsi\!\in\!\Upsilon, \widehat{\mPsi}\!\in\!\widehat{\Upsilon}$}
\State{$\vd^{+}\!=\![d_{\text{JEANIE}}(\mPsi_{b},\mPsi'_{b,1,z})]_{\substack{b\in\idx{B}\\z\in\idx{Z}}}$ \gray{$\qquad\quad$(within-class distance)}}
\State{$\vd^{-}\!=\![d_\text{JEANIE}(\mPsi_{b},\mPsi'_{b,n,z})]_{\!\!\!\!\!\!\!\!\!\!\!\substack{b\in\idx{B}\\\;\;\quad n\in\idx{N}\!\setminus\!\{1\}\\z\in\idx{Z}}}$ \gray{$\quad\;\,$(between-class distance)}}
\State{${\mathcal{F}}\!:=\!{\mathcal{F}}\!- \!\omega_\text{EN}\nabla_{{\mathcal{F}}} \big(l(\vd^{+},\vd^{-})+\lambda\mathcal{L}_\text{align}\big)$}
\end{algorithmic}
{\bf Output:} $\mathcal{F}$ and $\mM$
\end{algorithm}

\begin{table}[!htbp]
\setlength{\tabcolsep}{0.12em}
\renewcommand{\arraystretch}{0.70}
\caption{Experimental results on NTU-120 (S$^2$GC backbone). All methods enjoy temporal
alignment by soft-DTW or JEANIE (joint temporal and viewpoint
alignment). We use the $\ell_2$ norm for comparing the codes in unsupervised setting with soft-DTW. For unsupervised JEANIE, the distance for comparing the codes is indicated.}
\vspace{-0.6cm}
\begin{center}
\resizebox{\linewidth}{!}{\begin{tabular}{ l l cc c c  c c c }
\toprule
&  & viewpoint & \multirow{2}{*}{align.} &  \multirow{2}{*}{20} & \multirow{2}{*}{40} & \multirow{2}{*}{60} & \multirow{2}{*}{80} & \multirow{2}{*}{100}\\ 
&  & simulation &   &   &  &  &  & \\ 
\midrule
\multirow{15}{*}{\parbox{1.0cm}{{\bf Unsup.} +Transf.}}
& soft-DTW (HA)  & - & T  &  11.2 &  16.3 & 19.0 & 25.8 & 30.9\\
& soft-DTW (SC)  & - & T  & 12.1 & 17.4 & 21.4 & 27.0 & 32.7\\
& soft-DTW (SC$_+$)  & - & T & 11.8 & 17.0 & 21.2 & 26.5 & 32.2\\
& soft-DTW (LLC)  & - & T  & 14.0 & 18.7 & 23.1 & 29.3 & 34.1\\
& soft-DTW (SA)  & - & T  & 15.0 & 20.1 & 24.3 & 30.5 & 38.3\\
& soft-DTW (LcSA)  & - & T  & 15.7 & 21.4 & 25.2 & 32.0 & 40.2\\
& JEANIE (LLC)--$\ell_1$  & CamVPC & T+2V  & 18.0 & 23.8 & 30.5 & 36.3 & 43.0\\
& JEANIE (LLC)--$\ell_2$  & CamVPC & T+2V  & 18.3 & 24.2 & 30.8 & 36.0 & 43.3\\
& JEANIE (LLC)--HIK  & CamVPC & T+2V  & 18.3 & 24.0 & 31.0 & 36.3 & 43.0\\
& JEANIE (LLC)--CSK  & CamVPC & T+2V  & 17.8 & 24.0 & 30.8 & 36.3 & 43.0\\
& JEANIE (LcSA)--$\ell_1$  & CamVPC & T+2V  & 18.3 & 24.5 & 32.0 & 39.5 & 48.0\\
& JEANIE (LcSA)--$\ell_2$  & CamVPC & T+2V  & {\bf 18.6} & 25.0 & 32.2 & {\bf 40.0} & {\bf 48.5}\\
& JEANIE (LcSA)--HIK  & CamVPC & T+2V  & 18.3 & 24.8 & {\bf 32.2} & 39.6 & 48.0\\
& \cellcolor{LightCyan}JEANIE (LcSA)--CSK  & \cellcolor{LightCyan}CamVPC & \cellcolor{LightCyan}T+2V  & \cellcolor{LightCyan}{\bf 18.6} & \cellcolor{LightCyan}{\bf 25.2} & \cellcolor{LightCyan}32.0 & \cellcolor{LightCyan}39.6 & \cellcolor{LightCyan}{\bf 48.5}\\
& FVM (LcSA)--CSK  & CamVPC & T+2V  & 17.5 & 22.4 & 30.7 & 36.1 & 44.5\\
\bottomrule
\end{tabular}}
\label{ntu120results_uns}
\end{center}
\vspace{-0.3cm}
\end{table}

\begin{table}[!htbp]
\setlength{\tabcolsep}{0.12em}
\renewcommand{\arraystretch}{0.70}
	\centering
	\caption{Experiments on 2D and 3D Kinetics-skeleton. We
use the $\ell_2$ norm for comparing the codes in unsupervised
setting with soft-DTW. For unsupervised JEANIE, the distance for comparing the codes is indicated.}
	\label{kinetics_results_uns}  
	\vspace{-0.3cm}
	\resizebox{0.9\linewidth}{!}{\begin{tabular}{llcccc}  
		\toprule
		&  & viewpoint & \multirow{2}{*}{alignment} & \multirow{2}{*}{2D skel.} & \multirow{2}{*}{3D skel.}\\
		&  & simulation &   & & \\
  \midrule
\multirow{8}{*}{\parbox{1.0cm}{{\bf Unsup.} +Transf.}}
& soft-DTW(LLC) & -& T & 18.7 & 21.3\\
& soft-DTW(SA) & -& T & 18.7 & 21.8\\
& soft-DTW(LcSA) & -& T & 19.3 & 22.2\\
& JEANIE (LcSA)--$\ell_1$ & CamVPC & T+2V & - & 28.0\\
& JEANIE (LcSA)--$\ell_2$ & CamVPC & T+2V & - & {\bf 28.3}\\
& \cellcolor{LightCyan}JEANIE (LcSA)--HIK & \cellcolor{LightCyan}CamVPC & \cellcolor{LightCyan}T+2V & \cellcolor{LightCyan}- & \cellcolor{LightCyan}{\bf 28.3}\\
& JEANIE (LcSA)--CSK & CamVPC & T+2V & - & {\bf 28.3} \\
& FVM (LcSA)--$\ell_2$ & CamVPC & T+2V & - & 25.1\\
\bottomrule
\end{tabular}}
\end{table}

\begin{table*}[!tbp]
    \centering
    \resizebox{1\linewidth}{!}{
    \begin{tabular}{lccccccccccccc}
        \toprule
         & \multirow{3}{*}{Supervised} & \multicolumn{11}{c}{Unsupervised} & \multirow{3}{*}{MAML-inspired Fusion}\\
         \cline{3-13}
         & & \texttt{alpha\_iter}: & 100 & \multicolumn{4}{c}{50} & 30 & 10 & 10 & 10 & 10 & \\
         & & \texttt{dic\_iter}: & 10 & \multicolumn{4}{c}{10} & 10 & 10 & 30 & 50 & 100 & \\
         \cline{5-8}
         & & & & $k\!=\!1024$ & 2048 & 4096 & 8192 & & & & & & \\
        \midrule
        Time (s)& $1.95$& & $26.99$& $7.94 $& $11.76$ & $24.06$& $42.72$ & $22.48$& $21.10$ & $34.77$& $43.53$& $72.56$ & $38.77$ \\
        Accuracy (\%)  & \cellcolor{LightCyan}64.6 &  & \cellcolor{LightCyan}62.6 & 58.8 & 62.3 & \cellcolor{LightCyan}62.6 & 59.2 & 59.5 & 58.5 & 58.8 & 61.3 & 61.9 & \cellcolor{LightCyan}72.0\\
        \bottomrule
    \end{tabular}}
    \caption{Time cost (seconds) per 10 episodes \vs~ performance (\%) on MSRAction3D. We set stride step $S\!=\!5$ and $M\!=\!10$. Dictionary size $k\!=\!4096$ unless indicated otherwise, and $\tau^*\!=\!30$. See the text for remarks about a relatively larger number of epochs required for the convergence of  supervised FSAR compared to the unsupervised FSAR.}
    \label{tab:scratch}
\end{table*}

\vspace{0.1cm}
\noindent{\bf Fusion by alignment of supervised and unsupervised feature maps.} Inspired by domain adaptation, Algorithm \ref{code:fusion} performs a fusion of supervised and unsupervised FSAR by alignment of feature maps obtained with supervised and unsupervised FSAR. Specifically, we start by generating  representations with several viewpoints.  For each mini-batch of size $B$ we form a set with $N'$ feature maps which are passed to Algorithm~\ref{code:unsup}. Subsequently, from EN parameters $\mathcal{F}$ we obtain parameters $\widehat{\mathcal{F}}$ that help accommodate unsupervised reconstruction-driven learning. We compute ``unsupervised'' feature maps for such parameters and encourage ``supervised'' feature maps to align with them based on the JEANIE measure. Parameter $\lambda\geq 0$ controls the strength of alignment. For the supervised step, we use the supervised loss from Eq. \eqref{eq:pos} and \eqref{eq:neg}. Finally, we update EN parameters $\mathcal{F}$. 
\end{sloppypar}

\section{Additional Results on Unsupervised FSAR}
\label{sec:add_ress}

Tables \ref{ntu120results_uns} and \ref{kinetics_results_uns} below show additional results on the NTU-120, the 2D and 3D Kinetics-skeleton datasets.

\section{Training Speeds}

Table \ref{tab:scratch} investigates supervised, unsupervised and fusion strategies in terms of speed. While supervised training appears to be faster, it also takes more episodes to converge, \eg, 400 \vs 80. Worth noting is that the unsupervised strategy is a non-optimized code whose dictionary learning and code assignment can be parallelized to bring computations times significantly down. 

\section{Inference Time}

\begin{table}[!htbp]
\vspace{-0.5cm}
\setlength{\tabcolsep}{0.12em}
\renewcommand{\arraystretch}{0.70}
\caption{A comparison of training/inference time (per query) on NTU-60 (\#training classes = 10).}
\begin{center}
\vspace{-0.3cm}
\resizebox{0.95\linewidth}{!}{
\begin{tabular}{l c c c c}
\toprule
& Training time (s)$\;\;$ & Inference time (s)$\;\;$ & Total inference $\;\;$ & Acc. (\%)\\
& & & time (s) & \\
\midrule
soft-DTW$_\text{aug}$& 0.098 & 0.019 & 178.5 & 56.8\\
TAP$_\text{aug}$& 0.124 & 0.024 & 225.5 & 57.6\\
\rowcolor{LightCyan} JEANIE &  0.099 &  0.020 & 187.0 &  65.0\\
\bottomrule
\end{tabular}}
\label{tab:inference-time}
\end{center}
\vspace{-0.5cm}
\end{table}

Table~\ref{tab:inference-time} below compares training and inference times per query on Titan RTX 2090. For soft-DTW, each query is augmented by $K\!\times\!K'\!=\!9$ viewpoints. In the test time, we average match distance over $K\!\times\!K'\!=\!9$ viewpoints of each test query (this is a popular standard test augmentation strategy) \wrt~support samples. 
This strategy is denoted as soft-DTW$_\text{aug}$. 
We also apply the above strategy to TAP (denoted as  TAP$_\text{aug}$). 
JEANIE also uses $K\!\times\!K'\!=\!9$ viewpoints per query.
We exclude the time of applying viewpoint generation as skeletons can be pre-processed at once (1.6h with non-optimized CPU code) and stored for the future use.
Among methods which use multiple viewpoints, JEANIE outperforms soft-DTW$_\text{aug}$ and TAP$_\text{aug}$ by 8.2\% and 7.4\% respectively. JEANIE outperforms ordinary soft-DTW and TAP by 11.3\% and 10.8\%. For soft-DTW$_\text{aug}$ and TAP$_\text{aug}$, their total training and testing were  $5\!\times$ and $9\!\times$ slowed compared to counterpart soft-DTW and TAP. This is expected as they had to deal with $K\!\times\!K'\!=\!9$ more samples. We tried also parallel JEANIE. Training JEANIE$_\text{par}$ with 4  Titan RTX 2090 took 44h, the total inference was 48s.

\section{Training and Evaluation Protocols for Skeletal FSAR}
\label{supp:protocols}

As MSRAction3D, 3D Action Pairs, and UWA3D Activity have not been used in FSAR, we created 10 training/testing splits by choosing half of class concepts for training, and half for testing per split per dataset. Training splits were further subdivided for crossvalidation. Below, we explain the selection process.

\vspace{0.1cm}
\noindent\textbf{FSAR (MSRAction3D).} As this dataset contains 20 action classes, we randomly choose 10 action classes for training and the remaining 10 for testing. We repeat this sampling process 10 times to form in total 10 training/test splits. For each split, we have 5-way and 10-way experimental settings. The overall performance on this dataset is computed by averaging the performance over 10 splits.

\vspace{0.1cm}
\noindent\textbf{FSAR (3D Action Pairs).} This dataset has in total 6 action pairs (12 action classes), each pair of action has very similar motion trajectories, \eg, {\it pick up a box} and {\it put down a box}. We randomly choose 3 action pairs to form a training set (6 action classes) and the half action pairs for the test set, and in total there are ${\binom nk}\!=\!{\binom {6}{3}\!=\!20}$ different combinations of train/test splits. As our training/test splits are based on action pairs, we are able to test  whether the algorithm is able to classify unseen action pairs that share similar motion trajectories. We use the 5-way protocol 
and average over all 20 splits.

\vspace{0.1cm}
\noindent\textbf{FSAR (UWA3D Activity).} This dataset has 30 action classes. We randomly choose 15 action classes for training and the rest  action classes for testing. We form in total 10 train/test splits, and we use 5-way and 10-way protocols on this dataset, averaged over all 10 splits.

\vspace{0.1cm}
\begin{sloppypar}
\noindent\textbf{Kinetics-skeleton.} In our experiments, we follow the training and evaluation protocol from work~\cite{ma2022learning,10377291}. We use the first 120 actions out of 400, with 100 samples per class. The numbers of training, validation and test categories are 80, 20 and 20, respectively. Below is the breakdown of 120 categories into training, validation and test categories:
\begin{enumerate}
    \item \textbf{Train}: $[1,2,5,6,8,9,11,12,13,14,15,16,17,18,19,20,21,23,\\24,25, 26,27,28,29,30,31,32,33,34,38,39,40,41,43,47,48,\\49,50,51,52,58, 59,63,64,65,66,67,68,69,70,71,72,73,74,\\75,76,77,78,79,82,83,84, 85,86,87,89,90,91,93,95,97,\\102,105,109,110,111,113,117,118,119]$
    \item \textbf{Validation}: $[4, 7, 10, 36, 37, 42, 44, 45, 55, 56, 61, 88, 94, 98, 103, \\104, 106,108,116,120]$
    \item \textbf{Test}:$[3,22,35,46,53,54,57,60,62,80,81,92,96,99,100,101,\\107,112, 114,115]$
\end{enumerate}
\end{sloppypar}

\begin{figure*}[!htbp]
\centering
\begin{subfigure}[b]{0.485\linewidth}
\includegraphics[trim=0 0 0 0, clip=true,width=0.99\linewidth]{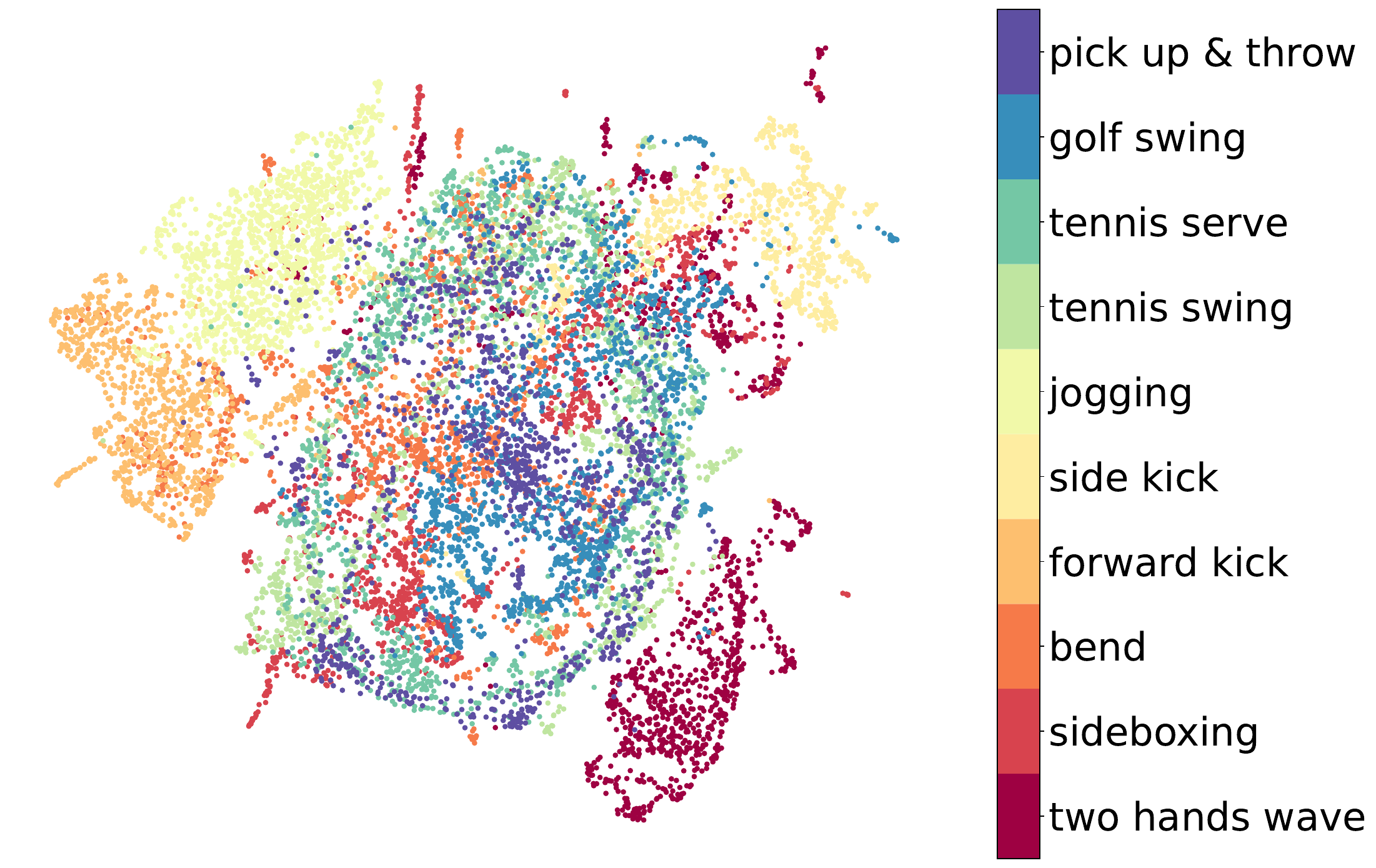}
\caption{\label{fig:jeanie-umap}JEANIE, temporal blocks (test accuracy: 80.28\%)}
\end{subfigure}
\begin{subfigure}[b]{0.485\linewidth}
\includegraphics[trim=0 0 0 0, clip=true,width=0.99\linewidth]{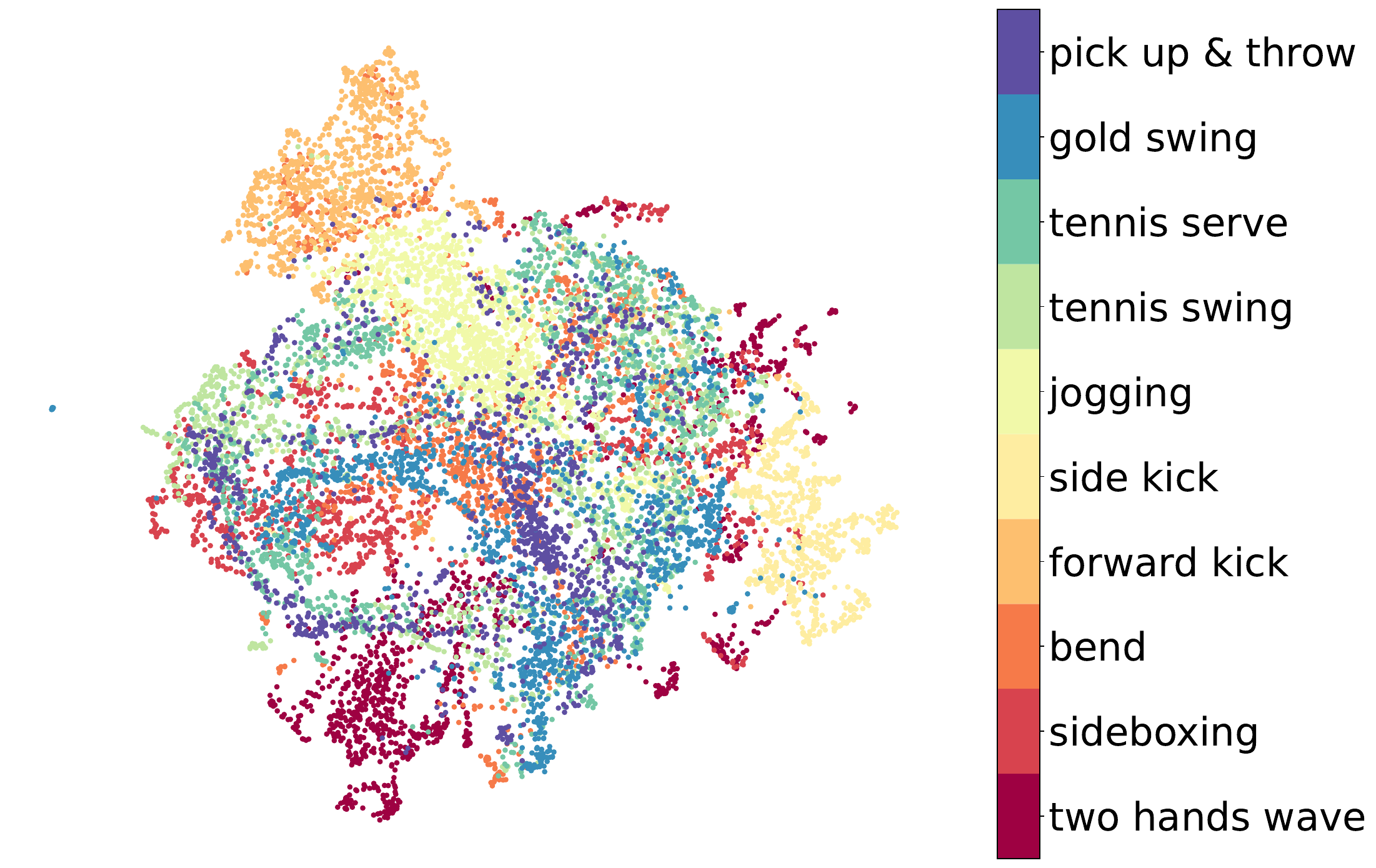}
\caption{\label{fig:softdtw-umap}Soft-DTW, temporal blocks (test accuracy: 77.51\%)}
\end{subfigure}
\begin{subfigure}[b]{0.485\linewidth}
\includegraphics[trim=0 0 0 0, clip=true,width=0.99\linewidth]{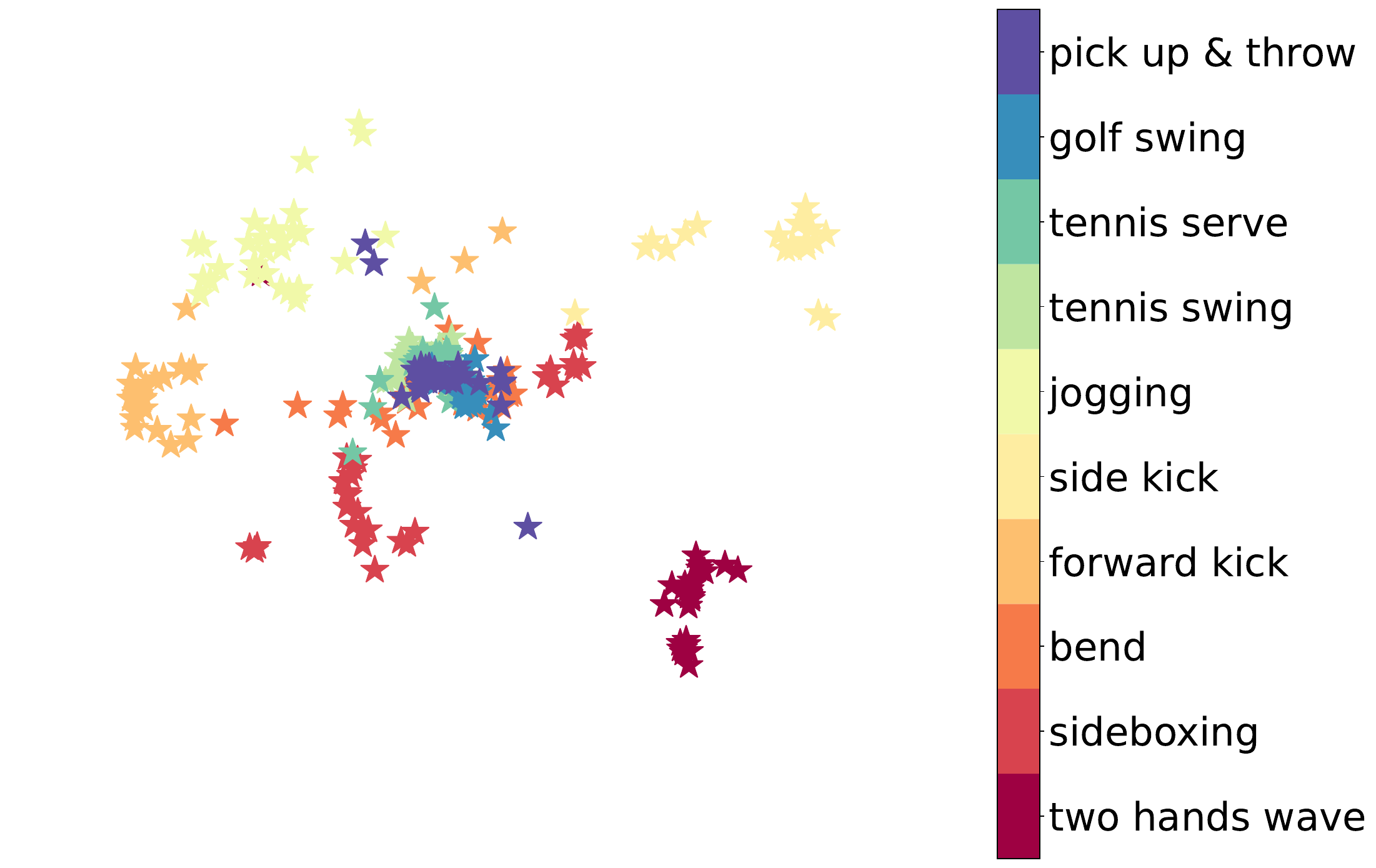}
\caption{\label{fig:jeanie-umap2}JEANIE, average-pooled temp. blocks (test accuracy: 80.28\%)}
\end{subfigure}
\begin{subfigure}[b]{0.485\linewidth}
\includegraphics[trim=0 0 0 0, clip=true,width=0.99\linewidth]{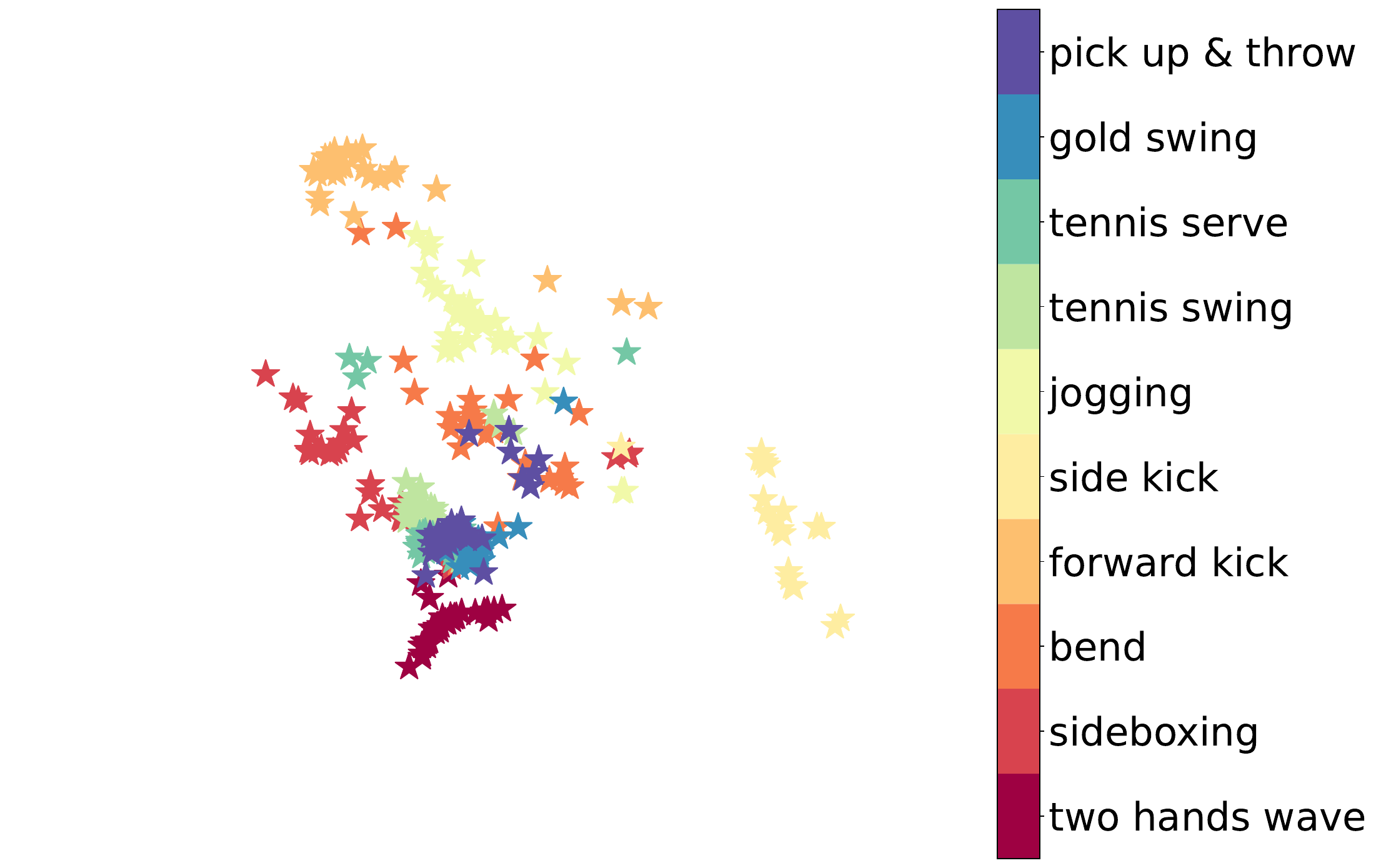}
\caption{\label{fig:softdtw-umap2}Soft-DTW, average-pooled temp. blocks (test accuracy: 77.51\%)}
\end{subfigure}
\begin{subfigure}[b]{0.485\linewidth}
\includegraphics[trim=0 0 0 0, clip=true,width=0.99\linewidth]{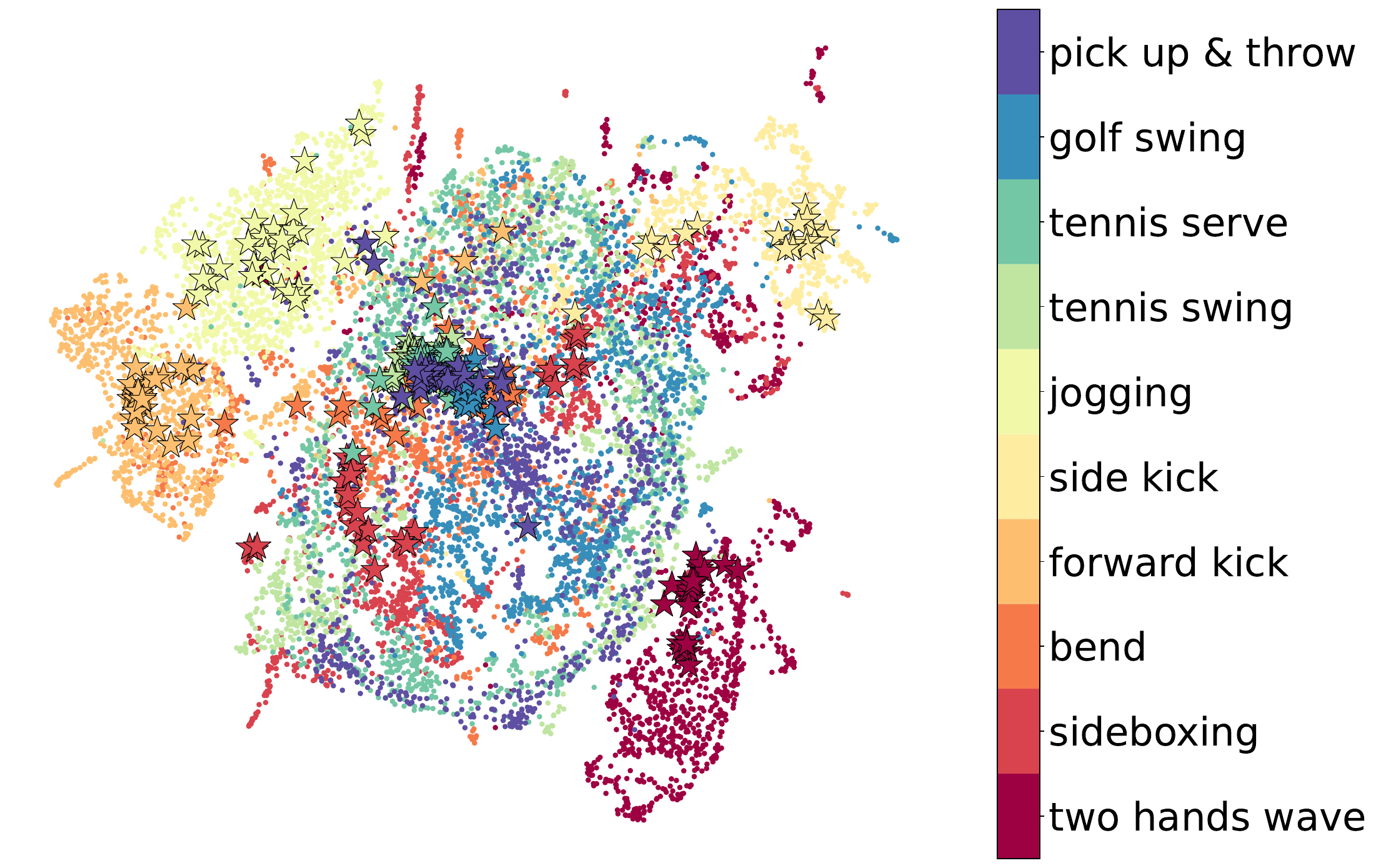}
\caption{\label{fig:jeanie-umap3}JEANIE, temporal blocks (dots) \& average-pooled temp.\\ blocks (stars) overlaid  (test accuracy: 80.28\%)}
\end{subfigure}
\begin{subfigure}[b]{0.485\linewidth}
\includegraphics[trim=0 0 0 0, clip=true,width=0.99\linewidth]{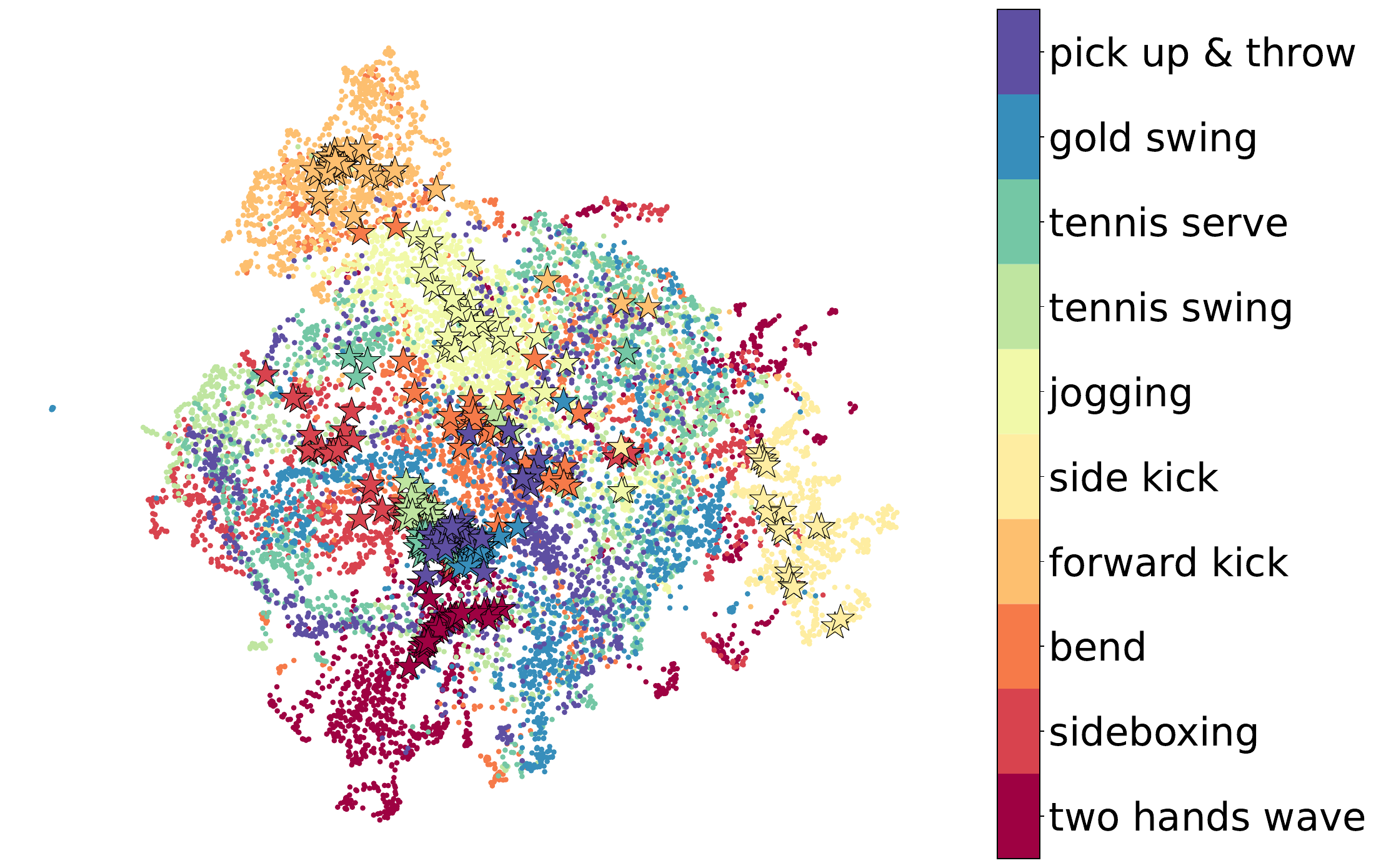}
\caption{\label{fig:softdtw-umap3}Soft-DTW, temporal blocks (dots) \& average-pooled temp.\\ blocks (stars) overlaid (test accuracy: 77.51\%)}
\end{subfigure}
\caption{UMAP-based visualizations of ({\em left}) JEANIE and ({\em right}) soft-DTW are generated using models trained on MSRAction3D. The test set is used for the visualisations. We visualize  temporal block representations (Fig. \ref{fig:jeanie-umap} and \ref{fig:softdtw-umap}), and average-pooled over blocks (along the temporal mode) feature representations (one per video) (Fig. \ref{fig:jeanie-umap2} and \ref{fig:softdtw-umap2}). Figures \ref{fig:jeanie-umap3} and \ref{fig:softdtw-umap3} overlay the figures above for better visualisation. Note that each colored dot represents a temporal block feature representation, whereas each star represents a video feature representation after average pooling  over blocks. JEANIE achieves somewhat more compact and better-separated class-wise clusters compared to soft-DTW.
}
\vspace{-0.4cm}
\label{fig:umap-vis}
\end{figure*}

\section{Visualisation based on UMAP}
We select the test set of MSRAction3D to showcase UMAP visualizations~\cite{mcinnes2018umap} of (i) temporal block features and (ii) average-pooled features along the temporal dimension for both JEANIE and soft-DTW. We use pre-trained models with the test accuracies  80.28\% and 77.51\% for JEANIE and soft-DTW, respectively. Figure \ref{fig:umap-vis} illustrates the comparisons. Each dot represents a temporal block feature, and each star denotes a video feature representation (after average pooling along the temporal mode). The figure shows that JEANIE appears to yield more compact and more separated clusters in comparison to soft-DTW.

\end{appendix}

{\small
\bibliographystyle{spmpsci}
\bibliography{egbib.bib}
}

\end{document}